\documentclass[10pt,journal,compsoc]{IEEEtran}

\usepackage{amsmath,amsfonts}
\usepackage{array}
\usepackage{textcomp}
\usepackage{stfloats}
\usepackage{url}
\usepackage{verbatim}
\usepackage{cite}

\usepackage{subcaption}
\usepackage{graphicx}
\usepackage{amsmath}
\usepackage{amssymb}
\usepackage{booktabs}
\usepackage{amsthm}
\usepackage{soul}
\usepackage{float}
\usepackage{xcolor}
\usepackage{multirow}
\usepackage{multicol}

\usepackage[flushleft]{threeparttable}
\usepackage{algorithm}
\usepackage{algpseudocode}
\usepackage[pagebackref,breaklinks,colorlinks]{hyperref}
\usepackage[capitalize]{cleveref}

\crefname{section}{Sec.}{Secs.}
\Crefname{section}{Section}{Sections}
\Crefname{table}{Table}{Tables}
\crefname{table}{Tab.}{Tabs.}

\newtheorem{thm}{Theorem}
\newtheorem{lem}{Lemma}

\newtheorem{prop}{Proposition}
\newtheorem{rmk}{Remark}

\newcommand{\tabincell}[2]{\begin{tabular}{@{}#1@{}}#2\end{tabular}}

\DeclareMathOperator*{\argmin}{arg\,min}

\newcommand{\calD}{\mathcal{D}}
\newcommand{\calO}{\tilde{\mathcal{O}}}

\newcommand{\calL}{\mathcal{L }}

\newcommand{\calN}{\mathcal{N}}

\newcommand{\bw}{\mathbf{w}}
\newcommand{\bg}{\mathbf{g}}

\newcommand{\RR}{\mathbb{R}}
\newcommand{\bx}{\mathbf{x}}

\newcommand{\by}{\mathbf{y}}
\newcommand{\bO}{\mathbf{O}}

\newcommand{\RomanNumeralCaps}[1]
    {\MakeUppercase{\romannumeral #1}}
\algnewcommand\algorithmicforpara{\textbf{for}}
\algnewcommand\algorithmicdoinparallel{\textbf{do in parallel}}
\algdef{S}[FOR]{ForParallel}[1]{\algorithmicforpara\ #1\ \algorithmicdoinparallel}

\begin{document}

\title{Temporal Gradient Inversion Attacks with Robust Optimization}

\author{Bowen~Li,
        Hanlin~Gu,
        Ruoxin~Chen, 
        Jie~Li,~\IEEEmembership{Senior Member,~IEEE},
        Chentao~Wu, ~\IEEEmembership{Member,~IEEE},
        Na~Ruan, ~\IEEEmembership{Member,~IEEE},
        Xueming Si  
        and~Lixin~Fan,~\IEEEmembership{Member,~IEEE}
\IEEEcompsocitemizethanks{\IEEEcompsocthanksitem Bowen Li, Ruoxin Chen, Jie Li, Chentao Wu, Na Ruan and Xueming Si are with the Department of Computer Science and Engineering, Shanghai Jiao Tong University, Shanghai, 200240, China. 
E-mail: \{li-bowen, chenruoxin, lijiecs, sixueming\}@sjtu.edu.cn, \{wuct, naruan\}@cs.sjtu.edu.cn.
\IEEEcompsocthanksitem Hanlin Gu and Lixin Fan are with WeBank AI Lab, WeBank, China. E-mail: \{ghltsl123, Lixin.Fan01\}@gmail.com, lixinfan@webank.com, Hanlin Gu and Bowen Li make equal contributions to this work.
}
\thanks {Corresponding author: Jie Li.}}

%



\IEEEtitleabstractindextext{%
\begin{abstract}
Federated Learning (FL) has emerged as a promising approach for collaborative model training without sharing private data. However, privacy concerns regarding information exchanged during FL have received significant research attention. \textit{Gradient Inversion Attacks (GIAs)} have been proposed to reconstruct the private data retained by local clients from the exchanged gradients. While recovering private data, the data dimensions and the model complexity increase, which thwart data reconstruction by GIAs. 
Existing methods adopt prior knowledge about private data to overcome those challenges. In this paper, we first observe that  
GIAs with gradients from a single iteration fail to reconstruct private data due to insufficient dimensions of leaked gradients, complex model architectures, and invalid gradient information.  
We investigate a Temporal Gradient Inversion Attack with a Robust Optimization framework, called TGIAs-RO, which recovers private data without any prior knowledge by leveraging multiple temporal gradients. To eliminate the negative impacts of outliers, e.g., invalid gradients for collaborative optimization, robust statistics are proposed. Theoretical guarantees on the recovery performance and robustness of TGIAs-RO against invalid gradients are also provided. Extensive empirical results on MNIST, CIFAR10, ImageNet and Reuters 21578 datasets show that the proposed TGIAs-RO with 10 temporal gradients improves reconstruction performance compared to state-of-the-art methods, even for large batch sizes (up to 128), complex models like ResNet18, and large datasets like ImageNet (224$\times$224 pixels). Furthermore, the proposed attack method  inspires further exploration of privacy-preserving methods in the context of FL. 
\end{abstract}

}

\maketitle

\IEEEdisplaynontitleabstractindextext

\IEEEpeerreviewmaketitle

\section{Introduction}

\IEEEPARstart{F}{ederated} learning (FL) \cite{mcmahan2017communication} is an emerging distributed learning framework where multiple participants collaboratively train a machine learning model without sharing private data directly. FL finds applications in diverse domains, including healthcare, finance, and the Internet of Things (IoT), among others \cite{yang2019federated}. However, there exist potential privacy risks associated with federated learning (particularly in the absence of protection methods) arising from the transmission of model parameters and gradients \cite{zhu2019deep,geiping2020inverting,yin2021see}.

Zhu et al. \cite{zhu2019deep} introduces \textit{Gradient Inversion Attacks (GIAs)}, which enable adversaries to reconstruct private image data with pixel-level accuracy by minimizing the discrepancy between gradients of reconstructed data and the leaked gradients. Subsequent research on GIAs has explored various approaches to leverage prior knowledge \cite{jeon2021gradient, yin2021see, zhu2020r}, such as utilizing statistics from batch normalization layers or training generative models to capture private data distributions \cite{yin2018byzantine, jeon2021gradient}. However, these methods often rely on impractical assumptions, limiting their applicability to idealized scenarios. Furthermore, existing GIAs that rely on gradients from a single iteration \cite{zhu2019deep, geiping2020inverting, wang2020sapag, zhao2020idlg} struggle to reconstruct private images effectively when faced with larger batch sizes, higher-dimensional data, and complex model architectures. Challenges arise from the presence of ReLU and Batch Normalization (BN) layers in the model architectures, as well as the inclusion of invalid gradient information. These factors contribute to optimization difficulties, leading to the convergence of GIAs to suboptimal solutions or complete failure in reconstruction tasks as demonstrated in Sect. \ref{sec:failure}.

Several privacy-preserving methods have been proposed to mitigate the vulnerabilities posed by existing gradient inversion attacks (GIAs) in federated learning. Notable approaches include differential privacy \cite{DLDP_Abadi16}, gradient sparsification \cite{wangni2018gradient}, among others. These methods have demonstrated their effectiveness in mitigating GIAs that rely on single iterations \cite{zhu2019deep, sun2021soteria}. For instance, increasing the level of noise in differential privacy prevents successful data recovery by GIAs \cite{zhu2019deep}. However, differential privacy acknowledges the increased privacy leakage with a growing number of gradient exposures \cite{DLDP_Abadi16}. In practice, model gradients are exchanged multiple times during the training processes. Therefore, our research aims to explore a new gradient inversion attack that leverages the advantages of multiple temporal gradients, which encourages further exploration and development of more robust privacy-preserving methods.

In this paper, we investigate a more powerful gradient inversion attack by leveraging multiple temporal gradients in Sect. \ref{sec:method},
named Temporal Gradient Inversion Attacks with Robust Optimization (TGIAs-RO). The proposed attack has the following two advantages: First, 
sufficient information (multiple temporal gradients) is adopted such that the private data is restored with a higher quality. Second, we formulate TGIAs-RO as one optimization problem that each temporal gradients collaboratively learn the common private data, which helps the restored data escape the bad local minima that single-temporal GIAs suffer. Furthermore, in order to 
eliminate the negative impacts of invalid gradients, robust statistics (such as Krum \cite{blanchard2017machine}, coordinate-wised median and trimmed mean \cite{yin2018byzantine}) are introduced for the collaborative temporal optimization in TGIAs-RO, which filter out the outliers, i.e., collapsed restored data during the optimization process. In short, our contributions are summarized as followings:

\begin{itemize}
    \item We first point out that GIAs adopting single-temporal gradients fail due to the insufficient gradient knowledge, complex model architectures and invalid gradients.
    
    
    \item We investigate a stronger gradient inversion attack without any prior knowledge, named Temporal Gradient Inversion Attacks with Robust Optimization (TGIAs-RO), by leveraging the multiple temporal gradients. We formulate GIAs with multiple temporal gradients as one optimization problem in a collaborative scheme that temporal gradients cooperatively learn the common private data, and further introduce the robust statistics to overcome limitation of the single-temporal GIAs.

    \item we provide theoretical guarantees on the recovery performance of our proposed method using multiple temporal gradients and robust statistics.
    
    \item We conduct extensive experiments on different models (LeNet, AlexNet, and ResNet) and different datasets (MNIST, CIFAR10, and ImageNet). Our empirical experiments present the superior data recovery performance of TGIAs-RO with just 10 temporal gradients. Furthermore, We provide the robustness of TGIAs-RO under different strategies used in federated learning, i.e.,  privacy-preserving mechanisms and client sampling strategy to improve communication efficiency.

\end{itemize}

\section{Related Work}\label{sect:related}
\begin{table*}[t]
  \renewcommand{\arraystretch}{1.25}
     \centering
     \caption{ Comparisons of TGIAs-RO with the  state-of-the-art GIAs methods in FL. }\label{tab: Related}
     \resizebox{0.99\textwidth}{!}{
      \begin{tabular}[l]{cccccc}
        \toprule
         \textbf{Method} &  \textbf{Optimization term} &
         \tabincell{c}{\textbf{Reported }\\\textbf{batch size}}
          & \tabincell{c}{\textbf{Image }\\\textbf{size}}
         &  \tabincell{c}{\textbf{Theoretical }\\\textbf{guarantee}} & 
          \tabincell{c}{\textbf{Reported prior}\\\textbf{information}}
          \\ \midrule
        
        DLG\cite{zhu2019deep}
         & $l_2$ distance & 8 & 32 & No & No \\
        
        \tabincell{c}{Inverting Gradients  \cite{geiping2020inverting}}
        & Cosine similarity & 8 & 224 & No & No \\
        SAPAG\cite{wang2020sapag} & Gaussian kernel based function & 8 & 32 & No & No \\
        R-GAP\cite{zhu2020r} & Recursive gradient loss & 5 & 32 & Yes & The rank of gradient matrix \\
        GradientInversion\cite{yin2021see} & 
        \tabincell{c}{Fidelity regularizers, \\Group regularizers}
        & 48 & 224 & No & 
        \tabincell{c}{Statistics from \\ normalization layers}
          \\
        GIML\cite{jeon2021gradient} & Cosine similarity & 4 & 48 & No & 
        \tabincell{c}{Distribution of 
private data}
          \\
        \textbf{TGIAs-RO (ours)} & \textbf{$l_2$ distance} & \textbf{128} & \textbf{224} & \textbf{Yes} & \textbf{No} 
        \\ \bottomrule
      \end{tabular}
      }
\end{table*}



\subsection{Gradient Inversion Attacks}
Recently, privacy risk of FL from the exchanged gradients gains growing interests. Zhu et al. \cite{zhu2019deep} firstly proposes a \textit{Deep Leakage from Gradients} (DLG) attack by matching the gradients of reconstructed data and the original gradients leaked from federated learning. 

Previous works on GIAs are mainly categorized into two classes, one class of works is developed by leveraging various kinds of prior knowledge. Geiping \textit{et al.} \cite{geiping2020inverting} improves the attack with the smoothness of data (such as image). Zhao \textit{et. al} \cite{zhao2020idlg} develops the prior with the label information of the data, Yin et al. \cite{yin2021see} applies group consistency and Batch Normalization information of estimated data to the training process. Another class of works adopts a pretrained generative model which is encoded with the knowledge of raw data distribution to enhance the prior knowledge of GIAs \cite{jeon2021gradient,li2022auditing}. All of above works introduce new regularization terms by leveraging the prior information.  Moreover, Jin et al. considers GIAs in vertical federated learning scenarios assuming the exposed gradients and batch ID to be known for attackers \cite{jin2021cafe}, whereas this assumption is too unrealistic to work in practice. Other works \cite{zhu2020r,pan2020exploring} provide the theoretical analysis of condition for successful GIAs. In addition, another work focuses on recovering the private data only given the leaked model weights instead of gradients \cite{geng2021towards,xu2022agic,dimitrov2022data}. Noted that Geng et al. and Xu et al. attempt to use a small amount of temporal gradients (up to 4) to recover data \cite{geng2021towards,xu2022agic}. Nevertheless, these works  are limited in both batch size and model complexity, further they lack theoretical justification on the capability of data reconstruction.


\subsection{Robust Statistics Based Aggregation} 
\textit{Robust Statistics} is widely adopted to solve Byzantine problem in distribution learning, which aims to remove the outliers generated by Byzantine attackers. Blanchard et al. \cite{blanchard2017machine} designs attackers whose gradients were close to other clients via Krum function; Other work designs coordinate-wise median method by considering median and its variants\cite{yin2018byzantine,pillutla2019robust,xie2018generalized}; In the process of GIAs, when recovering the private data according to multiple temporal gradients, single optimization process is easy to collapse or stuck in local minima. In temporal collaborative optimization, it is crucial to introduce \textit{Robust Statistics} to find the normal restored data w.r.t. corresponding temporal gradients and filter out those damaged data, which is far away from the normal intermediate restored data.

\section{Problem Statement and Proposed Method}
In this section, we first point out that GIAs adopting single-temporal gradients fail due to the insufficient gradient knowledge, complex model architectures and invalid gradients. Then we introduce a gradient inversion attack framework named Temporal Gradient Inversion Attacks with Robust Optimization (TGIAs-RO) leveraging multi-temporal gradients and robust statistics to overcome the limitation of the single-temporal GIAs. 

\begin{table}[!htbp] 
  \renewcommand{\arraystretch}{1.05}
  \centering
  \setlength{\belowcaptionskip}{15pt}
  \caption{Table of Notations}
  \label{table: notation}
    \begin{tabular}{c|p{5.5cm}}
    \toprule
    Notation & Meaning\cr
    \midrule\
    $K$ & Total number of clients \cr \hline
    $k$ & Index of client $k$ \cr \hline
    $\calD = (\bx, \by)$ & Original data including input images $\bx$ and label $\by$ \cr \hline
    $b$ & Batch size of  $\calD$ \cr \hline
    $n$ & Size (Dimension) of $\calD$ \cr \hline
    $\hat{\calD}, \hat{\bx}, \hat{\by}$ &  Restored data, input images and label\cr \hline
    $\bw, \calL$ & Model weights and loss of main task\cr \hline
     $\nabla_{\bw} \calL(\bw, \calD)$ & Model Gradients for loss $\calL$ on data $\calD$ w.r.t. $\bw$ \cr \hline

 $p$ & Dimension of Model gradients\cr \hline
 $T$ & Number of temporal gradients used \cr \hline
  $()_t$ & Items for $t_{th}$ temporal gradients\cr \hline
  $()^s$ & Items for $s_{th}$ updating\cr \hline
 $f_t$ & Empirical loss for GIAs in $t_{th}$ temporal information as Eq. \eqref{each in multi}\cr \hline
  $\alpha_t$ & The weights of $t_{th}$ local empirical loss function\cr \hline
  $f$ & Total collaborative loss for GIAs for $T$ temporal information as Eq. \eqref{eq: multi_temporal} \cr \hline
  $\nabla f_t(\hat{\bx}_t)$ & Gradients of function $f_t$ w.r.t. $\hat{\bx}_t$\cr \hline
   $R_g$ & Number of steps for global aggregation\cr \hline
 $R_l$ & Number of steps for local optimization\cr \hline
 $m$ & Number of collapsed items\cr \hline
 $\|\cdot \| $ & $\ell_2$ norm \cr 
    \bottomrule
    \end{tabular}
\end{table}
\subsection{Settings} \label{sec:preliminary}
\noindent\textbf{Federated Learning with Shared Gradients.} In federated learning scenarios without any protection methods, $K$ clients share raw gradients to collaboratively train a model as following steps:
\begin{itemize}
    \item Each client $ k$  uses its own dataset $\mathcal D_k$ to update new model  parameters $\mathbf w_{k}$ by optimizing the model main task and send the updates $\mathbf g_{k}$ to a server.
    \item The server aggregates the received local model updates to  average global updates $\mathbf g$  as the following \eqref{FLavg} and distributes them to each client:   
    \begin{equation}
   \mathbf g = \sum\limits_{k=1}\limits^K \frac{\lambda_k}{K}\mathbf g_{k}, 
   \label{FLavg}
\end{equation}

   where $\lambda_k$ is the weight of $\mathbf w_{k}$ and $\sum\limits_{k=1}\limits^K  \lambda_k =K$. 
\end{itemize}
These two steps iterate until the performance of the federated model does not improve. 

\noindent\textbf{Threat Model in FL.} In this work, we assume the server to be an \textit{honest-but-curious} adversary, which obeys the training protocol but attempts to obtain the private data of clients according to model weights and updates. Formally, clients need to exchange the intermediate gradient $\bg$ generated by private data $\bx$ as $\bg=f(\bx)$. In case that $f()$ is an invertible function, it is straightforward for adversaries to infer data $x = f^{-1}(\bg)$ from the exposed information. Even if  $f()$ is NOT invertible, adversaries may still estimate $\hat{\bx}$ by minimizing $||f(\hat{\bx}) -\bg||$. 

\noindent\textbf{Strategies in FL.} In this paper, we also consider the more realistic scenario in FL adopting client sampling and privacy-preserving strategies (see Sect. \ref{subsec:exp-robust}). The sampling strategy represents the server randomly selects $s$ clients' gradients to update the global model. Privacy-preserving mechanisms we consider include Differential Privacy \cite{DLDP_Abadi16} (add Gaussian noise into uploaded gradients) and Gradient Sparsification \cite{wangni2018gradient} (upload partial gradients for each client).\\
\noindent\textbf{Multiple Temporal Gradients.}
This paper investigates the FedSGD setting proposed by McMahan et al. \cite{mcmahan2017communication}, where clients participate in each communication round by uploading gradients corresponding to a single batch of data. We make the assumption that the local batch data is randomly sampled from the local dataset and used for training at each client. In this scenario, semi-honest adversaries, such as the server, have the ability to observe the gradient information $\{\bg_1, \bg_2, \cdots, \bg_{T_0}\}$ pertaining to a specific client multiple times, and their objective is to infer the original data point $\bx$ based on the available set of gradients $\{\bg_1, \bg_2, \cdots, \bg_{T_0}\}$.\\
\noindent\textbf{Loss Function and Gradients.} We assume that the model is a neural network parameterized by $\bw$, the loss function of the main task is $\calL(\bw, \calD)$, where $\calD = (\bx, \by)$ is batch data including $b$ local data $\{(x_i, y_i)\}_{i=1}^b$. The gradients of loss w.r.t. $\bw$ is denoted as
\begin{equation}
    \nabla_{\bw}\calL(\bw, \calD) = \frac{\partial\calL(\bw, \calD)}{\partial \bw} = \frac{1}{b}\sum_i^b\frac{\partial\calL(\bw, x_i, y_i )}{\partial \bw}.
\end{equation}
In addition, during the training of FL, the gradients may be leaked in each training iteration, and we define leaked gradients $\bO_t$ at the $t_{th}$ iteration as $\nabla_{\bw_t}\calL(\bw_t, \calD) $. We refer reviewers to see notion of parameters in Appendix A.

\subsection{Why Gradient Inversion Attacks (GIAs) with Single-temporal Gradients Fail?} \label{sec:failure}
Existing GIAs take use of gradients in a single training iteration to recover private data, which always suffers a failure. We motivate our design by providing intuitions on why GIAs with a single iteration always fail. Assume that adversaries use the objective loss function as Eq. \eqref{eq:gradient-inversion-single} \cite{zhu2019deep}, aimed at recovering the original batch data $\calD$ by comparing the gradients generated by restored batch data and the leaked gradients of original batch data $\calD$ as following:
\begin{equation} \label{eq:gradient-inversion-single}
\begin{split}
    \hat{\calD} = \argmin_{\hat{\calD}}||\nabla_{\bw}\calL(\bw, \hat{\calD}) - \nabla_{\bw}\calL(\bw, \calD)||^2, 
\end{split}
\end{equation}

\noindent where $ \nabla_{\bw}\calL(\bw, \calD) \in \RR^p$ is a $p$ dimension tensor, i.e., gradients known by adversaries and $\calD \in \RR^n$ is the $n$ dimension data to be reconstructed. There are three challenges that make single-temporal GIAs easy to fail.  
\subsubsection{Insufficient Dimensions of Leaked Gradients} Geiping et. al.\cite{geiping2020inverting} pointed out that reconstruction quality could be regarded as the number of parameters $p$ versus $n$, and reconstruction is at least as difficult as image recovery from incomplete data \cite{benning2018modern} when $p<n$. Moreover, \cite{aono2017privacy} attempted to formulate GIAs in fully connected layer as solving linear equation systems where the number of linear equations is closed to $p$. Nevertheless, \cite{zhu2020r} illustrated part of linear equations are invalid especially for convolution neural network, which implies partial leaked gradients is useless. Since the private data dimension $n$ is affected by input image size and batch size $b$, thus it is challenged to recover data as batch size and image size increase. For example, if GIAs 
are conducted on a LeNet-5 ($p=5e^4$) to recover CIFAR10 images in a batch $b$ = 128 with the pixel size $3\times32\times32$, thus $n=3\times32\times32\times128 = 4e^5,$ it is obvious that $p<<n$, which makes GIAs almost impossible, our experimental results in Sect. \ref{sec:exp} support this observation. 

\begin{figure} [h!]
\begin{subfigure}[t]{1\linewidth}
\centering
\begin{minipage}[t]{0.18\textwidth}
\centering
\scriptsize Initialization
\end{minipage}
\begin{minipage}[t]{0.18\textwidth}
\centering
\scriptsize Iteration 1
\end{minipage}
\begin{minipage}[t]{0.18\textwidth}
\centering
\scriptsize Iteration 2
\end{minipage}
\begin{minipage}[t]{0.18\textwidth}
\centering
\scriptsize Iteration 3
\end{minipage}
\begin{minipage}[t]{0.18\textwidth}
\centering
\scriptsize Ground Truth
\end{minipage}
\end{subfigure}

\begin{subfigure}[t]{1\linewidth}
\centering
\begin{minipage}[t]{0.18\textwidth}
\centering
\includegraphics[width=1.3cm]{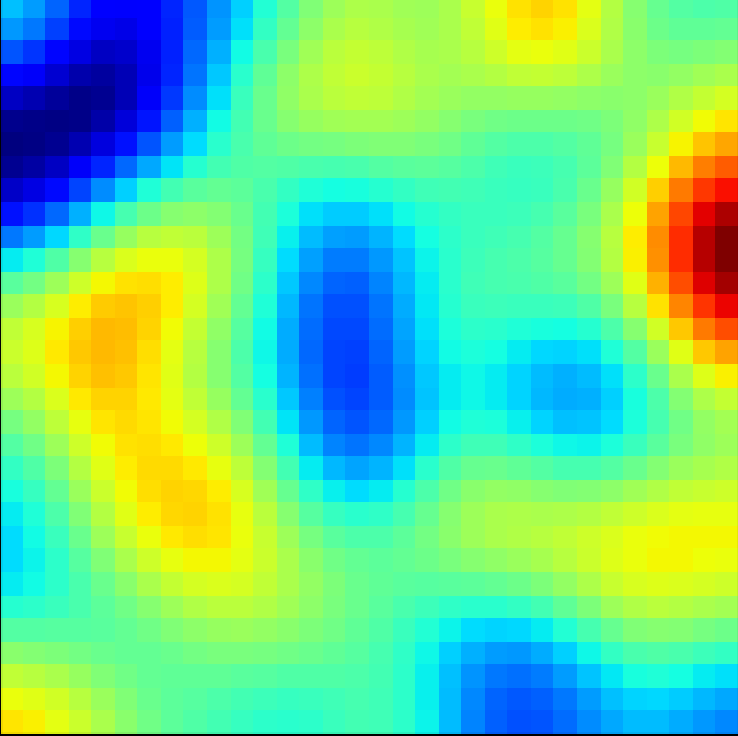}
\end{minipage}
\begin{minipage}[t]{0.18\textwidth}
\centering
\includegraphics[width=1.3cm]{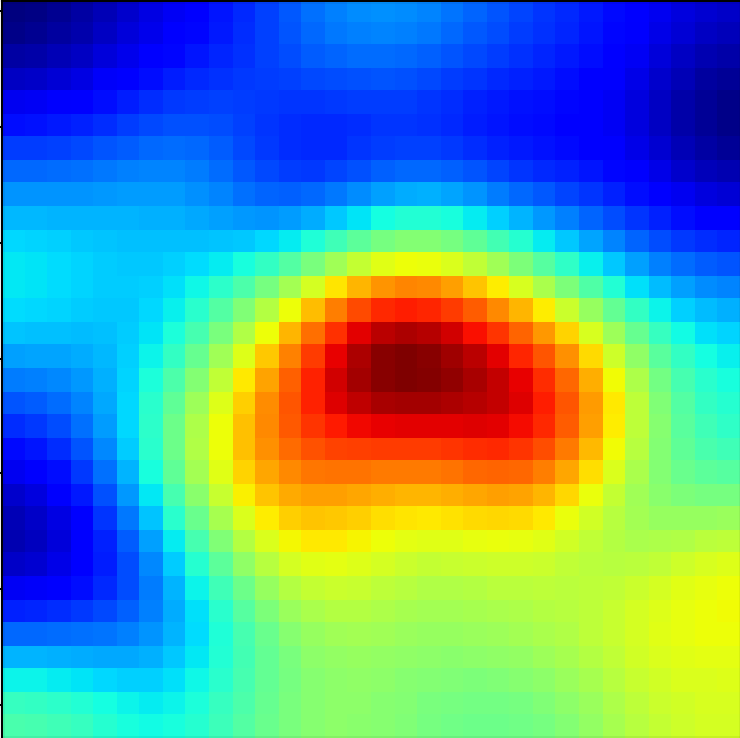}
\end{minipage}
\begin{minipage}[t]{0.18\textwidth}
\centering
\includegraphics[width=1.3cm]{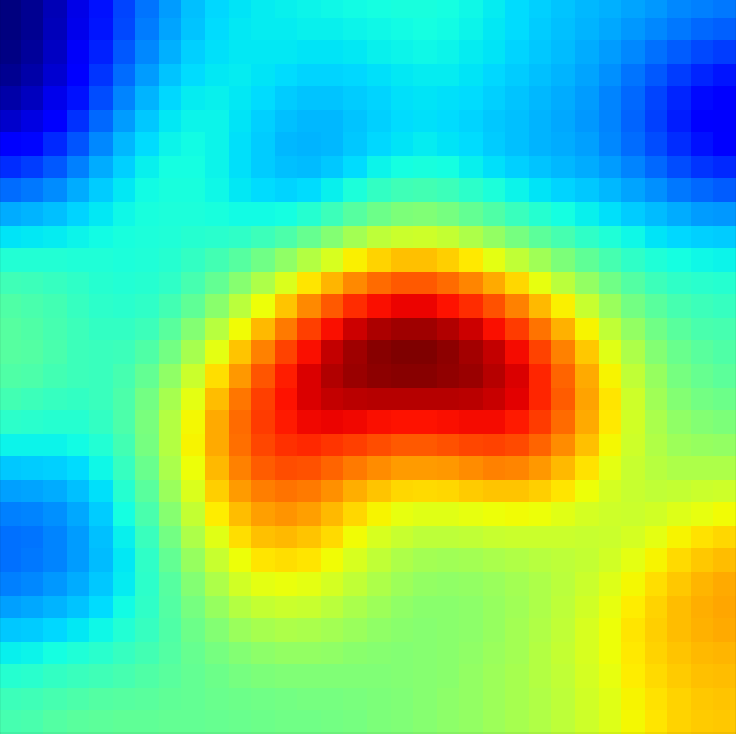}
\end{minipage}
\begin{minipage}[t]{0.18\textwidth}
\centering
\includegraphics[width=1.3cm]{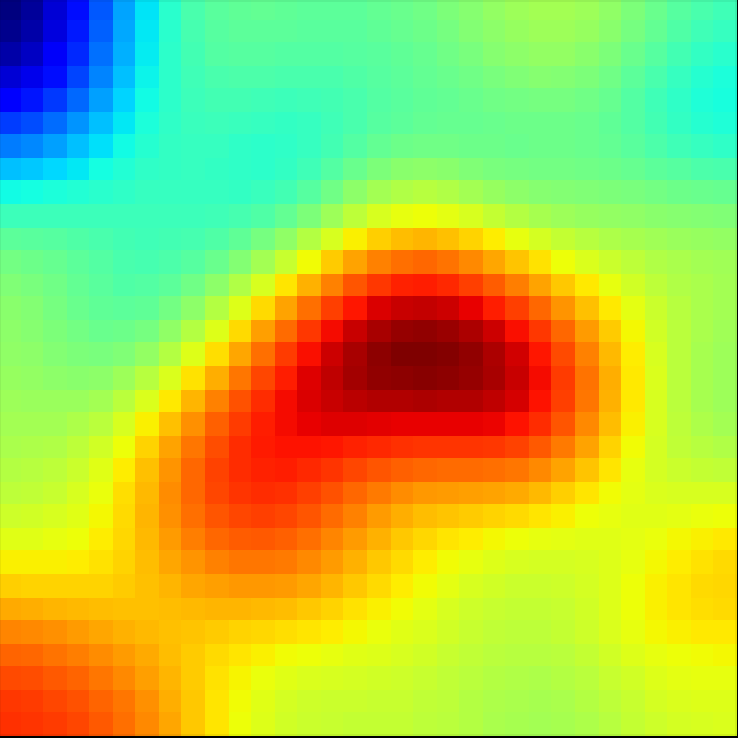}
\end{minipage}
\begin{minipage}[t]{0.18\textwidth}
\centering
\includegraphics[width=1.3cm]{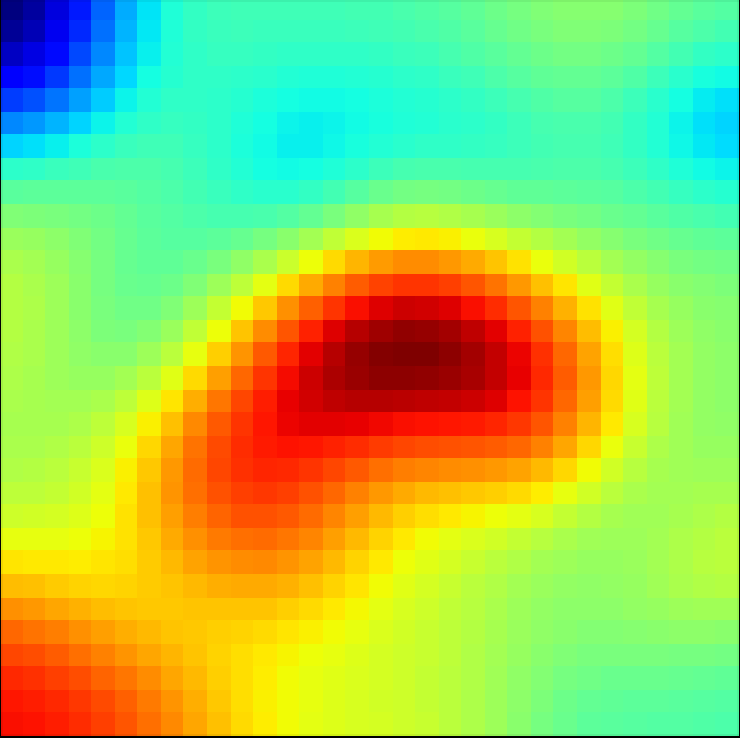}
\end{minipage}

\centering
\begin{minipage}[t]{0.18\textwidth}
\centering
\scriptsize PSNR: 10dB
\end{minipage}
\begin{minipage}[t]{0.18\textwidth}
\centering
\scriptsize 12dB
\end{minipage}
\begin{minipage}[t]{0.18\textwidth}
\centering
\scriptsize 16dB
\end{minipage}
\begin{minipage}[t]{0.18\textwidth}
\centering
\scriptsize 18dB
\end{minipage}
\begin{minipage}[t]{0.18\textwidth}
\centering
\scriptsize Ground Truth
\end{minipage}
\subcaption{\small DLG \cite{zhu2019deep} with a 6-layer AlexNet (vanilla architecture).} 
\end{subfigure}

\begin{subfigure}[t]{1\linewidth}
\centering
\begin{minipage}[t]{0.18\textwidth}
\centering
\includegraphics[width=1.3cm]{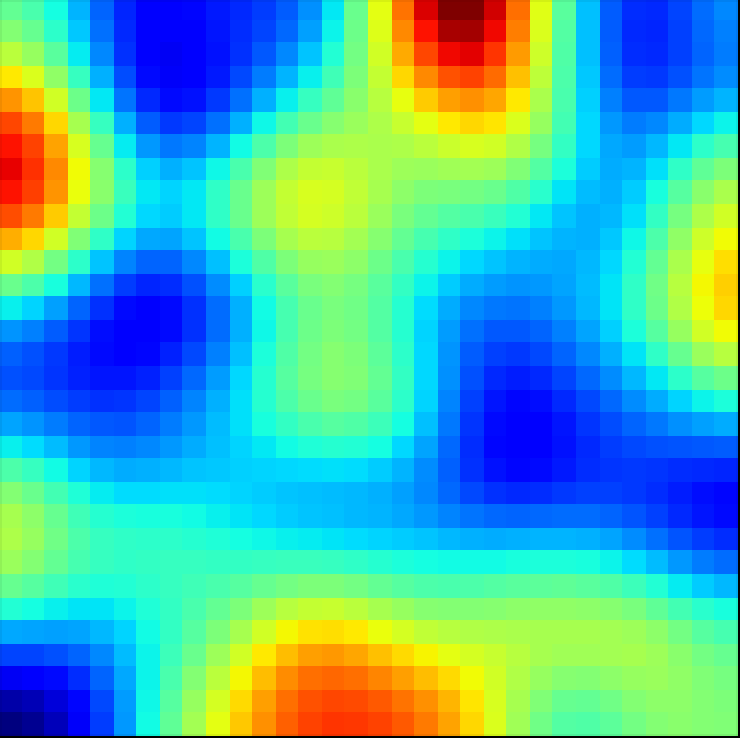}
\end{minipage}
\begin{minipage}[t]{0.18\textwidth}
\centering
\includegraphics[width=1.3cm]{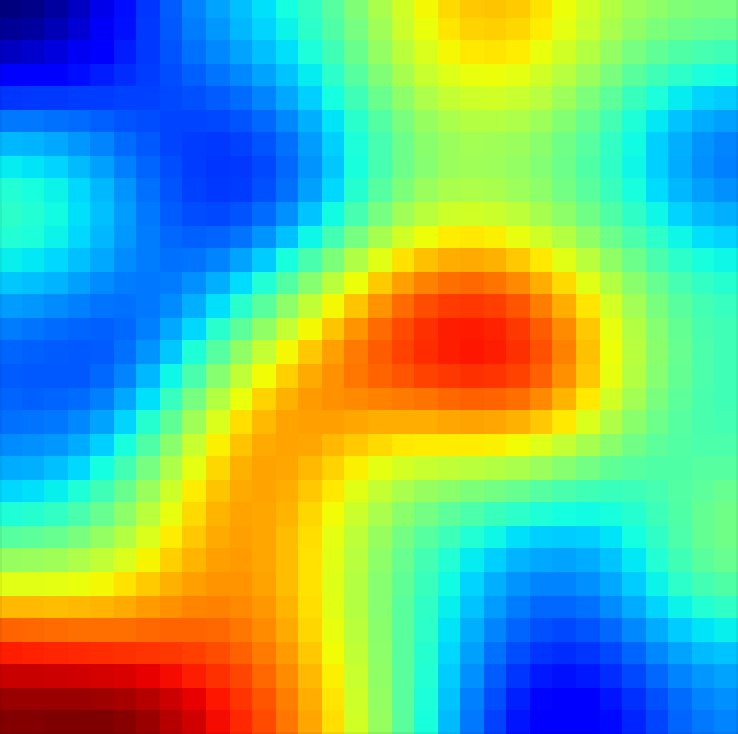}
\end{minipage}
\begin{minipage}[t]{0.18\textwidth}
\centering
\includegraphics[width=1.3cm]{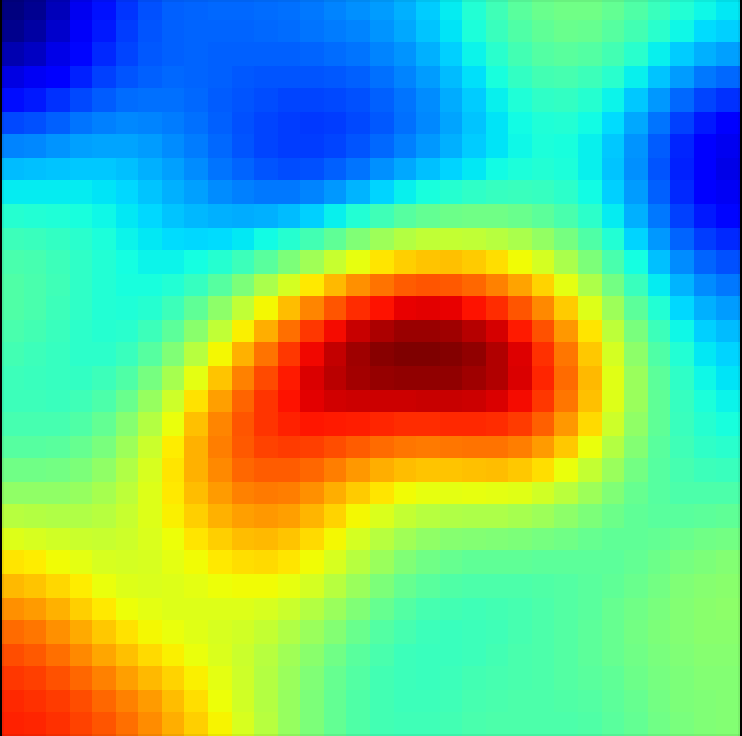}
\end{minipage}
\begin{minipage}[t]{0.18\textwidth}
\centering
\includegraphics[width=1.3cm]{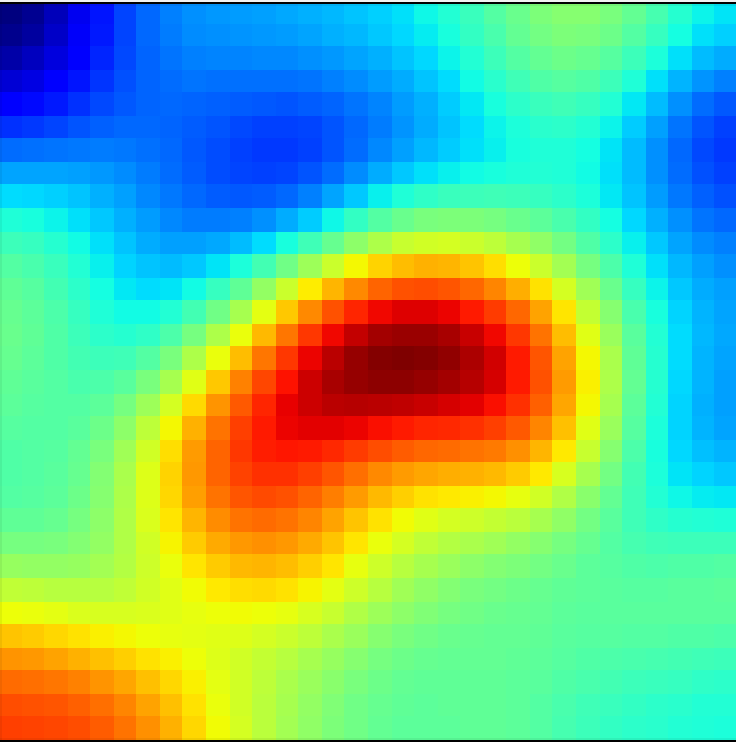}
\end{minipage}
\begin{minipage}[t]{0.18\textwidth}
\centering
\includegraphics[width=1.3cm]{imgs/heatmaps/gt.png}
\end{minipage}
\begin{minipage}[t]{0.18\textwidth}
\centering
\scriptsize 10dB
\end{minipage}
\begin{minipage}[t]{0.18\textwidth}
\centering
\scriptsize 12dB
\end{minipage}
\begin{minipage}[t]{0.18\textwidth}
\centering
\scriptsize 14dB
\end{minipage}
\begin{minipage}[t]{0.18\textwidth}
\centering
\scriptsize 14dB
\end{minipage}
\begin{minipage}[t]{0.18\textwidth}
\centering
\scriptsize Ground Truth 
\end{minipage}
\subcaption{\small DLG with a modified AlexNet (5  Normalization layers added).} 
\end{subfigure}

\begin{subfigure}[t]{1\linewidth}
\centering
\begin{minipage}[t]{0.18\textwidth}
\centering
\includegraphics[width=1.3cm]{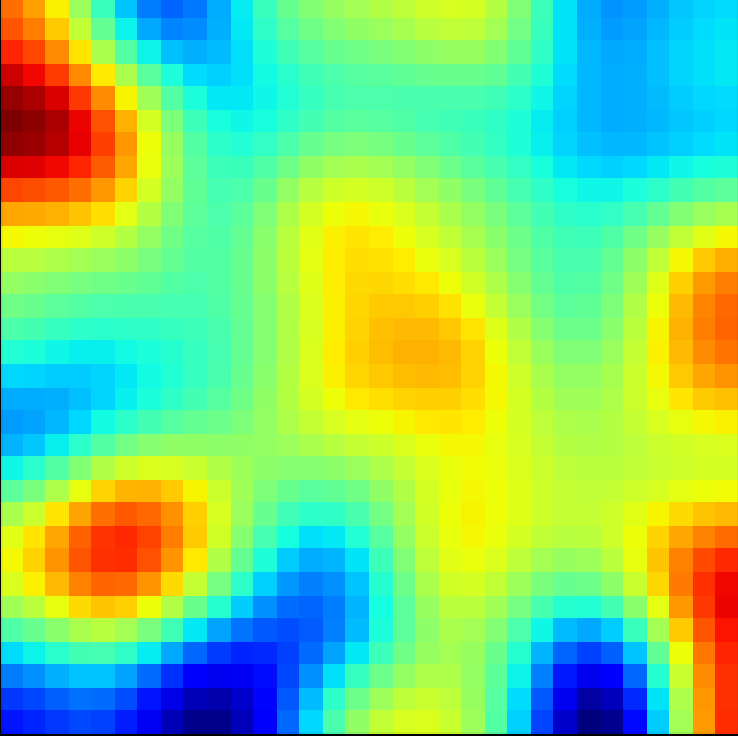}
\end{minipage}
\begin{minipage}[t]{0.18\textwidth}
\centering
\includegraphics[width=1.3cm]{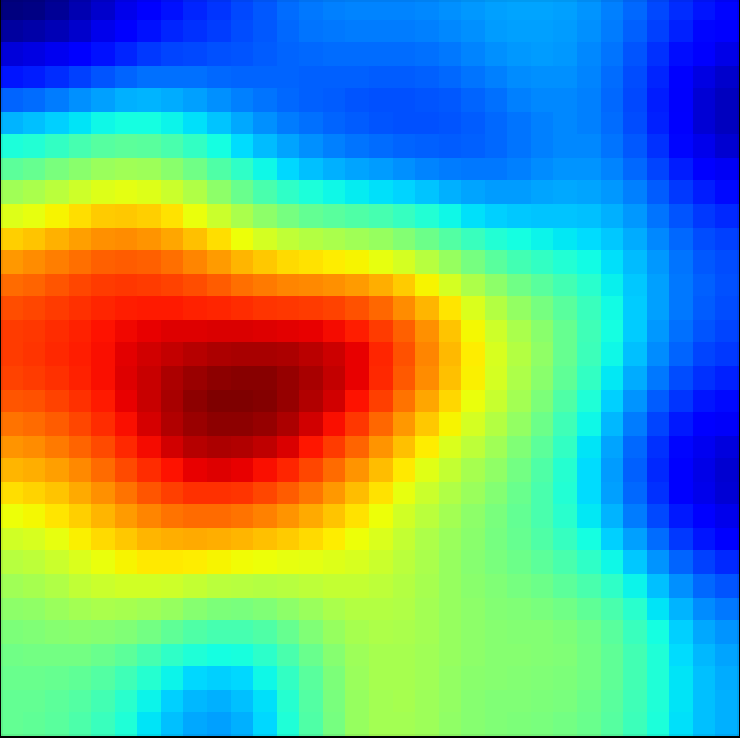}
\end{minipage}
\begin{minipage}[t]{0.18\textwidth}
\centering
\includegraphics[width=1.3cm]{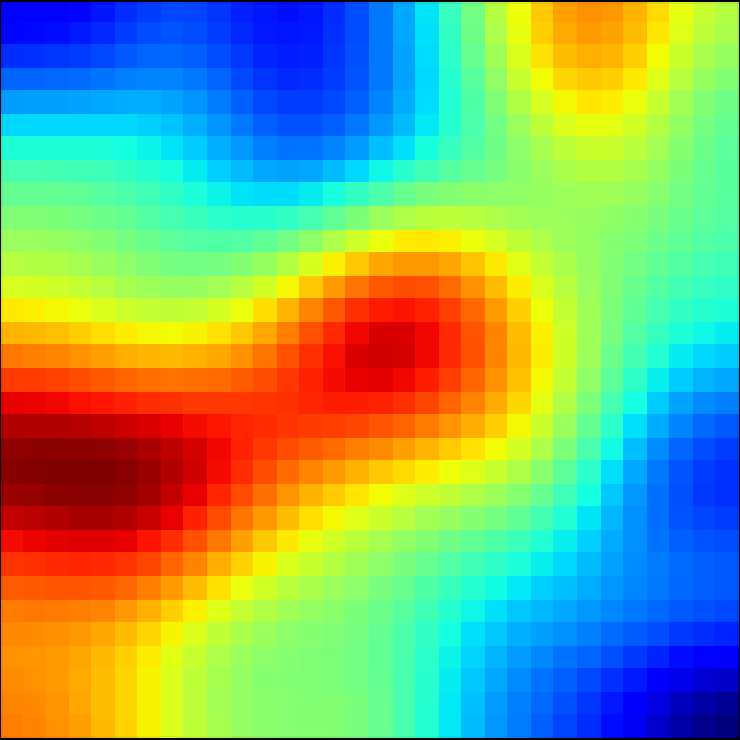}
\end{minipage}
\begin{minipage}[t]{0.18\textwidth}
\centering
\includegraphics[width=1.3cm]{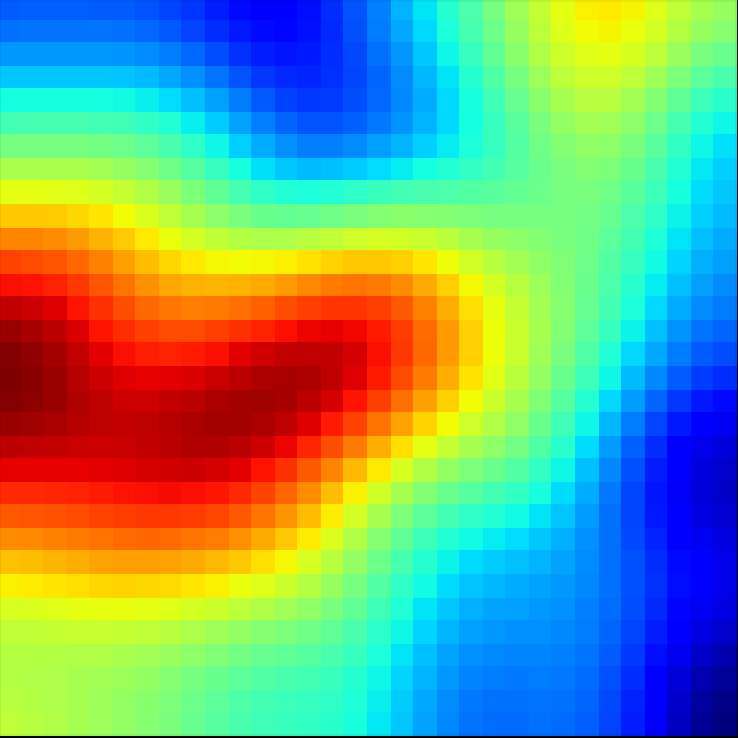}
\end{minipage}
\begin{minipage}[t]{0.18\textwidth}
\centering
\includegraphics[width=1.3cm]{imgs/heatmaps/gt.png}
\end{minipage}
\begin{minipage}[t]{0.18\textwidth}
\centering
\scriptsize 10dB
\end{minipage}
\begin{minipage}[t]{0.18\textwidth}
\centering
\scriptsize 11dB
\end{minipage}
\begin{minipage}[t]{0.18\textwidth}
\centering
\scriptsize 12dB
\end{minipage}
\begin{minipage}[t]{0.18\textwidth}
\centering
\scriptsize 12dB
\end{minipage}
\begin{minipage}[t]{0.18\textwidth}
\centering
\scriptsize Ground Truth%
\end{minipage}
\subcaption{\small DLG with a modified AlexNet (5  Normalization and ReLU layers added).} 
\end{subfigure}

 \caption{\small The optimization of the first channel in restored image in GIAs with different model architectures. } 
  \label{fig:heatmap1}

\end{figure}

\subsubsection{Complex Model Architectures} 
Complex model architectures like Normalization and ReLU layers would lead the optimization process (Eq. \eqref{eq:gradient-inversion-single}) to stuck in the local minima. For instance, \cite{zhu2020r} pointed out the convolution layers give rise to the existence of invalid gradient constraints. \cite{geiping2020inverting} analyzed that it is challenging to optimize Eq. \eqref{eq:gradient-inversion-single} using ReLU since higher-order derivatives are discontinuous. Specifically, Fig. \ref{fig:heatmap1} shows that the data recovery via GIAs with single-temporal gradients stuck in the bad local minima and restored data is increasingly far away from the ground truth when we add the Normalization and ReLU layers in sequence. We report more failure cases in Appendix B.6. 
\subsubsection{Invalid Gradient Information} 
\begin{figure} [h!]
\begin{subfigure}[t]{1\linewidth}
\centering
\begin{minipage}[t]{0.18\textwidth}
\centering
\scriptsize Initialization
\end{minipage}
\begin{minipage}[t]{0.18\textwidth}
\centering
\scriptsize Iteration 1
\end{minipage}
\begin{minipage}[t]{0.18\textwidth}
\centering
\scriptsize Iteration 2
\end{minipage}
\begin{minipage}[t]{0.18\textwidth}
\centering
\scriptsize Iteration 3
\end{minipage}
\begin{minipage}[t]{0.18\textwidth}
\centering
\scriptsize Ground Truth
\end{minipage}
\end{subfigure}

\begin{subfigure}[t]{1\linewidth}
\centering
\begin{minipage}[t]{0.18\textwidth}
\centering
\includegraphics[width=1.3cm]{imgs/heatmaps/1.1.1.png}
\end{minipage}
\begin{minipage}[t]{0.18\textwidth}
\centering
\includegraphics[width=1.3cm]{imgs/heatmaps/1.1.2.png}
\end{minipage}
\begin{minipage}[t]{0.18\textwidth}
\centering
\includegraphics[width=1.3cm]{imgs/heatmaps/1.1.3.png}
\end{minipage}
\begin{minipage}[t]{0.18\textwidth}
\centering
\includegraphics[width=1.3cm]{imgs/heatmaps/1.1.4.png}
\end{minipage}
\begin{minipage}[t]{0.18\textwidth}
\centering
\includegraphics[width=1.3cm]{imgs/heatmaps/gt.png}
\end{minipage}
\begin{minipage}[t]{0.18\textwidth}
\centering
\scriptsize PSNR: 10dB
\end{minipage}
\begin{minipage}[t]{0.18\textwidth}
\centering
\scriptsize 12dB
\end{minipage}
\begin{minipage}[t]{0.18\textwidth}
\centering
\scriptsize 16dB
\end{minipage}
\begin{minipage}[t]{0.18\textwidth}
\centering
\scriptsize 18dB
\end{minipage}
\begin{minipage}[t]{0.18\textwidth}
\centering
\scriptsize Ground Truth
\end{minipage}
\subcaption{\footnotesize DLG \cite{zhu2019deep} with an untrained AlexNet (vanilla architecture).} 
\end{subfigure}

\begin{subfigure}[t]{1\linewidth}
\centering
\begin{minipage}[t]{0.18\textwidth}
\centering
\includegraphics[width=1.3cm]{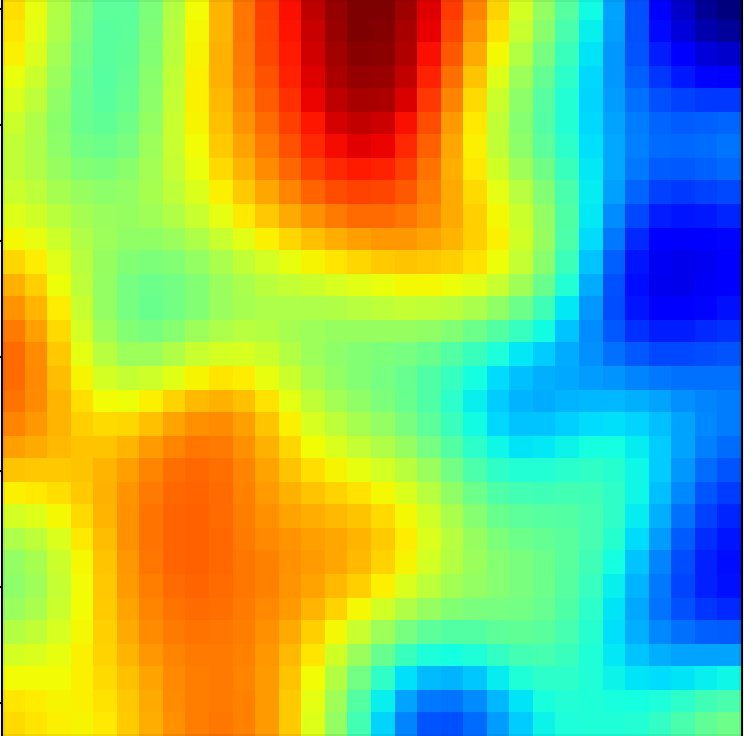}
\end{minipage}
\begin{minipage}[t]{0.18\textwidth}
\centering
\includegraphics[width=1.3cm]{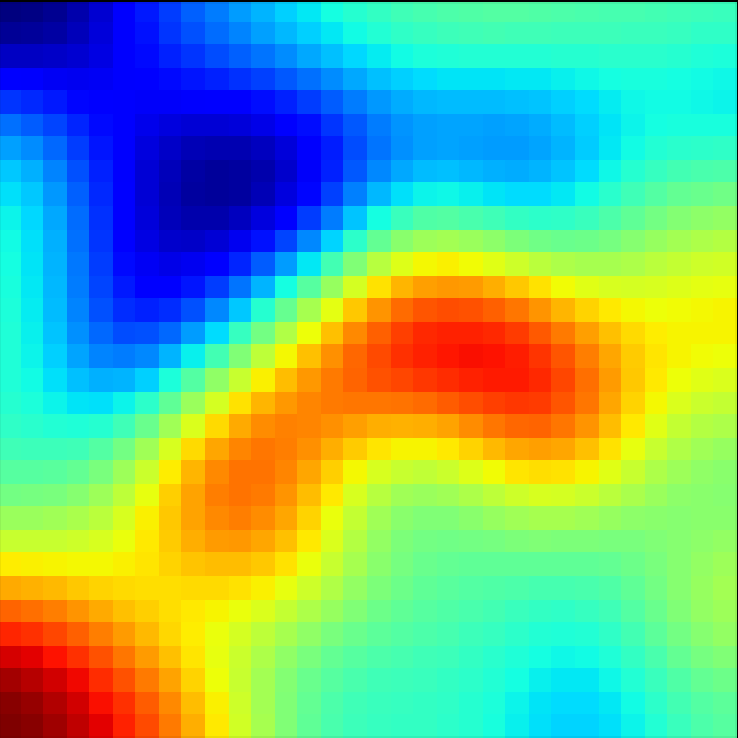}
\end{minipage}
\begin{minipage}[t]{0.18\textwidth}
\centering
\includegraphics[width=1.3cm]{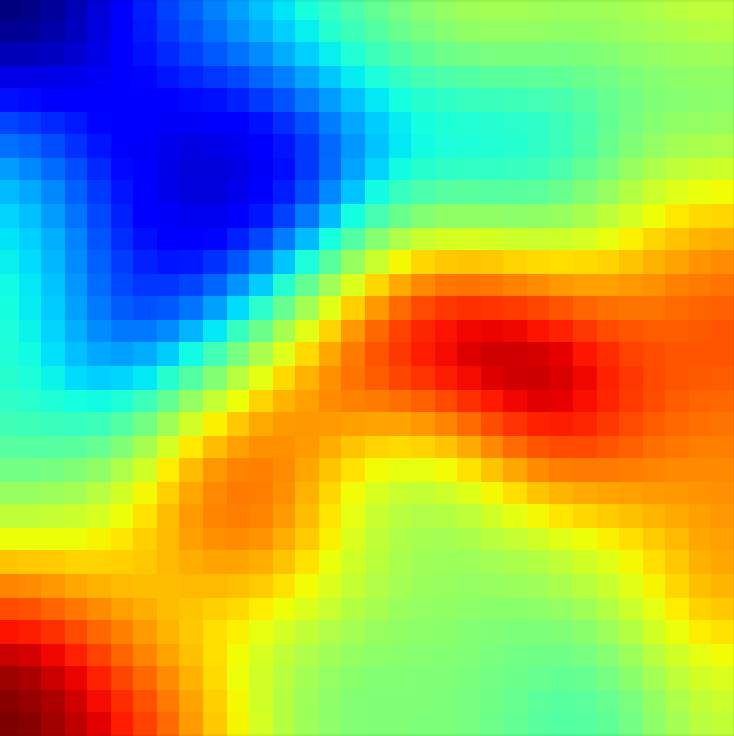}
\end{minipage}
\begin{minipage}[t]{0.18\textwidth}
\centering
\includegraphics[width=1.3cm]{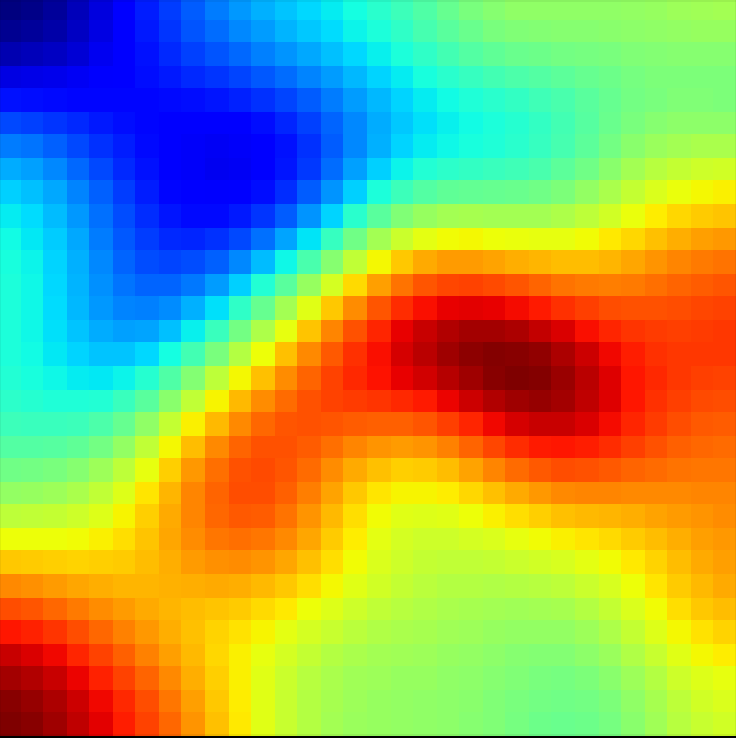}
\end{minipage}
\begin{minipage}[t]{0.18\textwidth}
\centering
\includegraphics[width=1.3cm]{imgs/heatmaps/gt.png}
\end{minipage}
\begin{minipage}[t]{0.18\textwidth}
\centering
\scriptsize 10dB
\end{minipage}
\begin{minipage}[t]{0.18\textwidth}
\centering
\scriptsize 12dB
\end{minipage}
\begin{minipage}[t]{0.18\textwidth}
\centering
\scriptsize 13dB
\end{minipage}
\begin{minipage}[t]{0.18\textwidth}
\centering
\scriptsize 14dB
\end{minipage}
\begin{minipage}[t]{0.18\textwidth}
\centering
\scriptsize Ground Truth
\end{minipage}
\subcaption{\footnotesize DLG with an AlexNet trained for 20 training epochs.} 
\end{subfigure}

\begin{subfigure}[t]{1\linewidth}
\centering
\begin{minipage}[t]{0.18\textwidth}
\centering
\includegraphics[width=1.3cm]{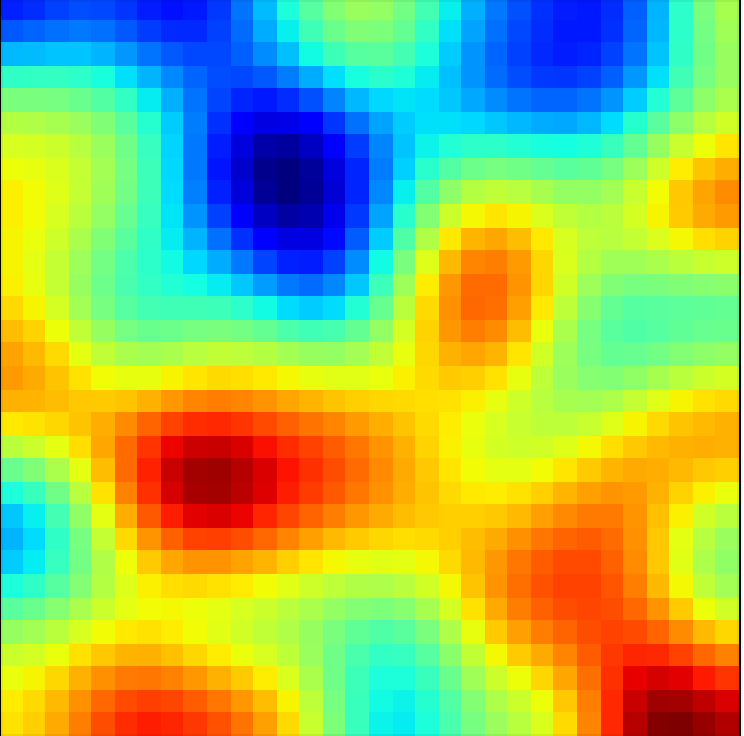}
\end{minipage}
\begin{minipage}[t]{0.18\textwidth}
\centering
\includegraphics[width=1.3cm]{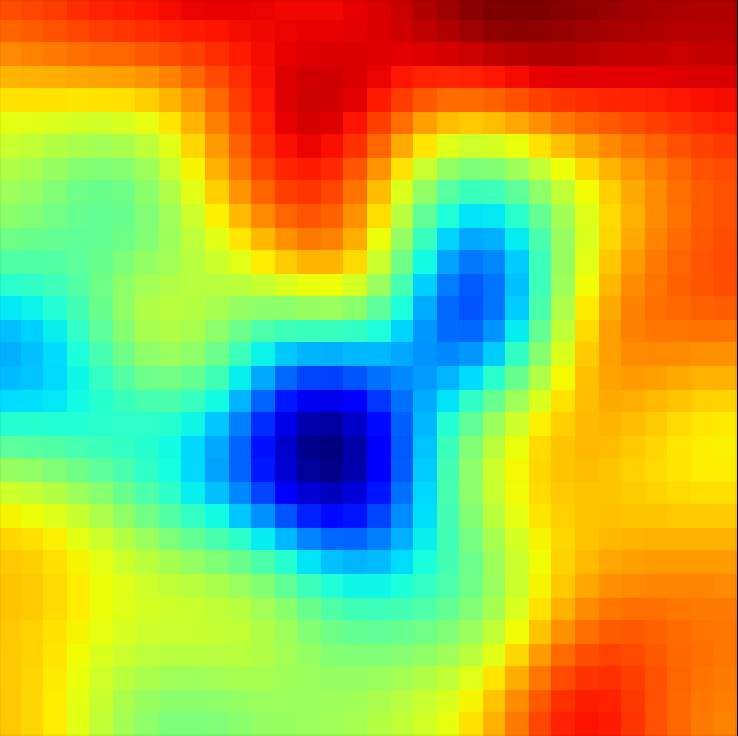}
\end{minipage}
\begin{minipage}[t]{0.18\textwidth}
\centering
\includegraphics[width=1.3cm]{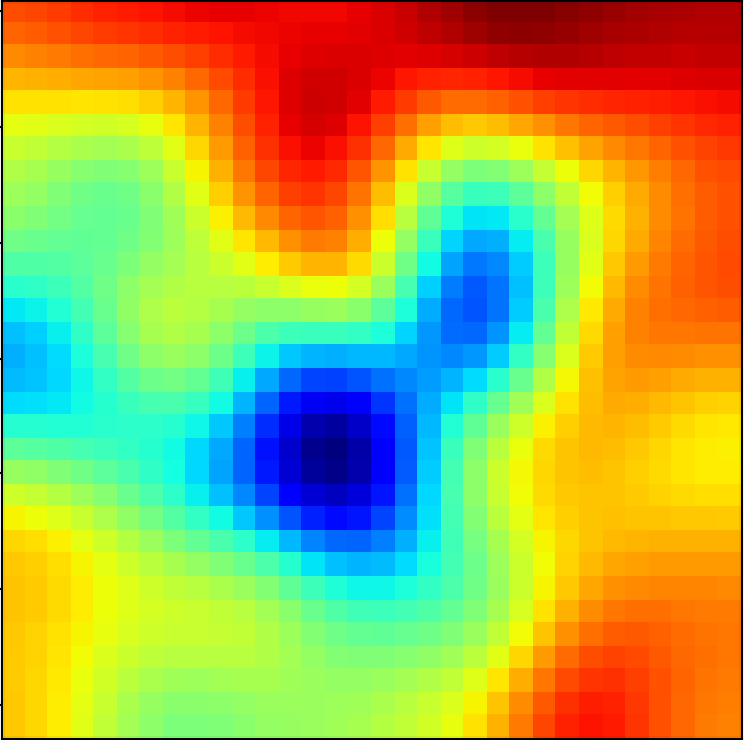}
\end{minipage}
\begin{minipage}[t]{0.18\textwidth}
\centering
\includegraphics[width=1.3cm]{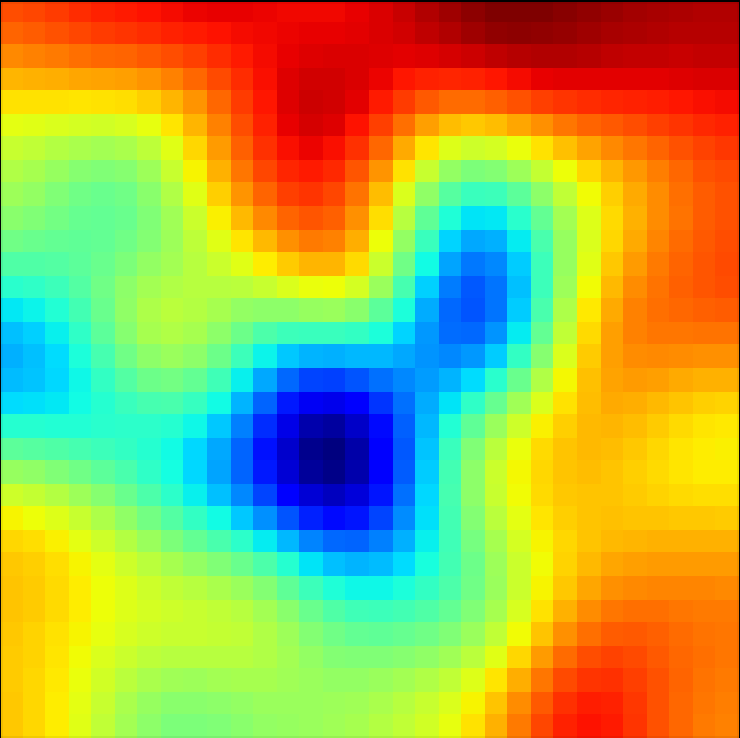}
\end{minipage}
\begin{minipage}[t]{0.18\textwidth}
\centering
\includegraphics[width=1.3cm]{imgs/heatmaps/gt.png}
\end{minipage}
\begin{minipage}[t]{0.18\textwidth}
\centering
\scriptsize 10dB
\end{minipage}
\begin{minipage}[t]{0.18\textwidth}
\centering
\scriptsize 8dB
\end{minipage}
\begin{minipage}[t]{0.18\textwidth}
\centering
\scriptsize 8dB
\end{minipage}
\begin{minipage}[t]{0.18\textwidth}
\centering
\scriptsize 8dB
\end{minipage}
\begin{minipage}[t]{0.18\textwidth}
\centering
\scriptsize Ground Truth
\end{minipage}
\subcaption{\footnotesize DLG with an AlexNet trained for 40 training epochs.} 

\end{subfigure}

 \caption{The optimization of the first channel in restored image in GIAs with models trained in different epochs. } 
  \label{fig:heatmap2}
\end{figure}

The leaked gradients are closed to zero when the model is trained or even convergent, which makes it difficult to recover private data by comparing the gradients value. Moreover, the Proposition \ref{prop1} illustrates the invalid gradient information generated by convergent model cannot help data recovery. In addition, our empirical findings in Sect. \ref{sec:ablation} validate the hardness, and visually, Fig. \ref{fig:heatmap2} shows that when the complex model is trained with more iterations, the restored data is collapsed seriously (PSNR is just 8dB), which is far away from ground truth.
\begin{prop} \label{prop1}
When the model on the main task is convergent for a series of batch data $\calD_1, \calD_2, \cdots$, i.e., $\nabla_{\bw}\calL(\bw, \calD_1) = \nabla_{\bw}\calL(\bw, \calD_2) = \cdots= 0$, then recovering $\calD_1$ using Eq. \eqref{eq:gradient-inversion-single} is impossible almost surely.
\end{prop}

We also provide the evidence in Tab. \ref{tab: gradientNorm}. Moreover, the large-scale (more complex) network provided in \cite{geiping2020inverting} leads the optimization process to stuck in the local minima (3.2.2 in the main text), although the gradient norm does not tend to zero.

\begin{table}[H]
\caption{Gradient norms of models in different training epochs.}\label{tab: gradientNorm}
\renewcommand{\arraystretch}{1.1}
 \centering
\resizebox{0.48\textwidth}{!}{
\begin{tabular}{ccccccc}
\toprule 
Epoch  Number & 1 & 2 & 10& 40  & 60 & 100\\ \midrule
LeNet & 2.22$e^{-4}$ & 6.45$e^{-5}$  &  2.53$e^{-5}$ & 1.38$e^{-5}$ &1.88$e^{-6}$ & 2.18$e^{-6}$\\
ResNet18 & 1.22$e^{-5}$  & 1.05$e^{-5}$   & 7.31$e^{-6}$  & 1.79$e^{-6}$ &3.71$e^{-7}$ & 4.05$e^{-7}$ \\ 
 \bottomrule
\end{tabular}
}
\end{table}


\begin{figure*}[t]
	\centering
        
	\includegraphics[width=6.3in]{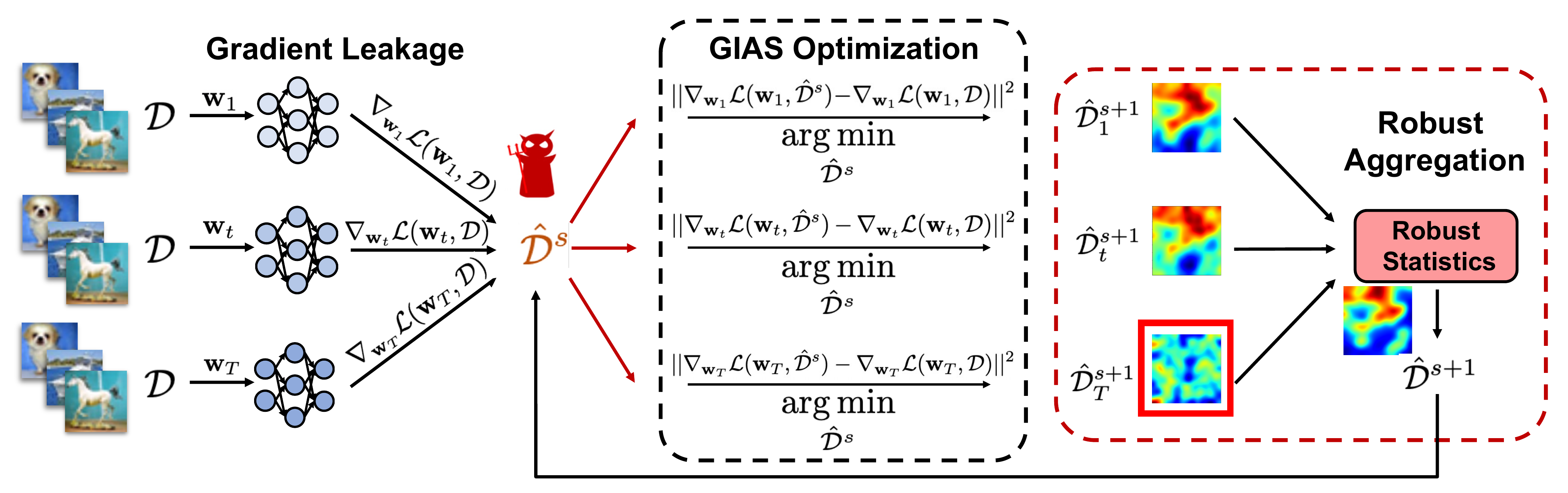}
	\caption{\small An overview of TGIAs-RO consisting of two recursive stages: 1) (the middle panel) the adversary implements gradient inversion optimizations with temporal gradients from multiple iterations; 2) (the right panel) the intermediate restored data is aggregated via robust statistics, where those collapsed data (red boxed ones) are filtered.}\vspace{-10pt}
	\label{fig: framework}
\end{figure*}
\subsection{TGIAs-RO: Temporal Gradient Inversion Attacks with Robust Optimization} \label{sec:method}
The main idea of proposed TGIAs-RO is to leverage the multiple temporal information (leaked gradients of multiple temporal iterations) and robust statistics to recover private data.
In order to address the challenges mentioned above, we firstly leverage multiple temporal information to 1) increase the volume of compared parameters ($p$) and 2) help the data recovery process converge to the global minima by keeping the loss of GIAs with multiple temporal information consistent. Moreover, we formulate the GIAs with multiple temporal gradients as one collaborative optimization problem \cite{konevcny2016federated} with two recursive stages: \\1) Firstly, optimizing each empirical loss function $f_t()$ (defined as Eq. \eqref{each in multi}) for $t_{th}$ temporal information in $R_l$ iterations to obtain updated $\hat{D}_t$.
\begin{equation}\label{each in multi}
 f_t(\hat{\calD}) := ||\nabla_{\bw_t}\calL(\bw_t, \hat{\calD}) - \nabla_{\bw_t}\calL(\bw_t, \calD)||^2.
\end{equation}
2) Secondly, aggregating all $\hat{D}_t$ over the $T$ gradients  as the initialization of the next updating.  
Then the overall collaborative optimization problem to solve $n$ dimension $\hat{D}$ with $T$ temporal gradients can be expressed as:
\begin{equation}\label{eq: multi_temporal}
    \text{min}_{\hat{\calD} \in \RR^p}\sum_{t=1}^T\alpha_t  f_t(\hat{\calD}),
\end{equation}
which aims to learn a common $\calD$, where $\alpha_t$ is the weight of $t_{th}$ local empirical loss function and $\sum_{t=1}^T\alpha_t =1$.


On the aggregation side of the collaborative optimization in Eq. \ref{eq: multi_temporal}, intermediate restored data after optimizing $s$ iterations for $t_{th}$ temporal gradients ($\hat{\calD}^s_t$) may collapse especially when temporal gradients come from model trained multiple iterations (see Sect. \ref{sec:failure}), we thus introduce robust statistics to aggregate $\{\hat{\calD}^s_t\}_{t=1}^T$ (as shown in the right panel of Fig. \ref{fig: framework}). The goal of leveraging the robust statistics is to filter out the outliers in $\{\hat{\calD}^s_t\}_{t=1}^T$, i.e., collapsed restored data during the TGIAs-RO optimization process. We assume the last $m$ temporal optimizations $\{f_t()\}_{t=T-m+1}^{T}$ collapse. Robust statistics such as Krum, coordinated-wise median and trimmed mean guarantee that those collapsed restored data are not selected in each optimization steps if $m<\frac{T}{2}$. We provide the following error bound of data reconstruction for our proposed approach TGIAs-RO via coordinated-wise median.
\begin{thm} \label{thm:thm1}
If $\{f_t\}_{t=1}^{T-m}$ is $\mu$ strong convex and L-smooth, the number of collapsed restored data $m$ is less than $\frac{T}{2}$. Choose step-size $\eta = \frac{1}{L}$
and Algo. \ref{algo:TGIAs-RO} obtains the sequence $\{\hat{\bx}^s=\text{RobustAgg/Median}(\hat{\bx}^s_1,\cdots, \hat{\bx}^s_T) : s\in[0:R_g]\}$
satisfying the following convergence guarantees:
\begin{equation}
    \left\|\hat{\bx}^{s} - \bx^*\right\|_2 \leq (1-\frac{\mu}{\mu+L})^s\left\|\hat{\bx}^{0} - \bx^*\right\|_2 + \frac{2 \Gamma}{\mu}.
\end{equation}
Moreover, we have
\begin{equation}
    \lim_{s \to \infty} \left\|\hat{\bx}^{s} - \bx^*\right\|_2  \leq  \frac{2 \Gamma}{\mu},
\end{equation}
where $\bx^*$ is the global optimal, $\Gamma =\calO(\kappa)$, $\kappa$ is the upper bound of $\|\nabla f_t( \hat{\bx}^s) - \frac{1}{T-m}\sum_{t=1}^{T-m}\nabla f_t(\hat{\bx}^s)\|_2$ for any $s$ and $t=1,\cdots,T-m$ and $\nabla f_t(\hat{\bx}^s)$ is the gradients of the function $f_t$ w.r.t. $\hat{\bx}^s$.
\end{thm}
\begin{proof}
According to Algo. \ref{algo:TGIAs-RO} for $R_l=1$, we obtain
\begin{equation*}
\begin{split}
       \hat{\bx}_t^s &= \hat{\bx}^{s-1} - \eta \text{Median}(\{ \nabla f_t(\hat{\bx}^{s-1})\}_{t=1}^T )\\
       & = (\hat{\bx}^{s-1} -\eta \nabla f^{s-1}) \\
       & \quad + \eta ( \nabla f^{s-1} - \text{Median}(\{ \nabla f_t(\hat{\bx}^{s-1}_t)\}_{t=1}^T)).
\end{split}
\end{equation*}
Thus, 
\begin{equation}
\begin{split}
        & \| \hat{\bx}^s - \bx^* \|_2 \\
        & = \| (\hat{\bx}^{s-1} -\eta \nabla f^{s-1} -\bx^*) \\
        & \quad + \eta ( \nabla f^{s-1} - \text{Median}(\{ \nabla f_t(\hat{\bx}^{s-1})\}_{t=1}^T )) \|_2 \\
        & \leq \underbrace{\|\hat{\bx}^{s-1} -\eta \nabla f^{s-1} -\bx^*\|_2}_{=:U} \\
        & \quad + \underbrace{\| \eta (\nabla f^{s-1} - \text{Median}(\{ \nabla f_t(\hat{\bx}^{s-1})\}_{t=1}^T )) \|_2}_{=:V}.
\end{split}
\end{equation}
Firstly, 
\begin{equation*}
\begin{split}
        U^2 = & \|\hat{\bx}^{s-1} - \bx^* \|_2^2 - 2 \eta<\hat{\bx}^{s-1} - \bx^* ,  \nabla f^{s-1}> \\
        & + \eta^2 \|\nabla f^{s-1} \|_2^2.
\end{split}
\end{equation*}
Since $f$ is $L$-smooth and $\mu$ strong convex, according to the \cite{bubeck2015convex}, we obtain:
\begin{equation}
\begin{split}
    & < \hat{\bx}^{s-1} - \bx^*, \nabla f^{s-1}>  \\
    & \geq \frac{1}{\mu+ L}\|\nabla f^{s-1} \|_2^2 + \frac{\mu L}{\mu+L}||\hat{\bx}^{s-1}-\bx^*||^2.
\end{split}
\end{equation}
Let $\eta = \frac{1}{L}$ and assume $\mu \leq L$, then we get
\begin{equation}
\begin{split}
    U^2 &\leq (1- \frac{2 \mu}{\mu+L})\| \hat{\bx}^{s-1} - \bx^* \|_2^2 - \frac{2}{L(\mu+L)} \|\nabla f^{s-1} \|_2^2 \\
    & + \frac{1}{L^2}\|\nabla f^{s-1} \|_2^2 \\
    & \leq (1- \frac{2 \mu}{\mu+L})\| \hat{\bx}^{s-1} - \bx^* \|_2^2. 
\end{split}
\end{equation}
Since $\sqrt{1-x} \leq 1-\frac{x}{2}$, we get
\begin{equation} \label{eq:U}
    U \leq (1- \frac{ \mu}{\mu+L})\| \hat{\bx}^{s-1} - \bx^* \|_2. 
\end{equation}
Furthermore, according to \textbf{Claim 1} in Appendix A, we have
\begin{equation}\label{eq:V}
    V \leq \calO(\kappa). 
\end{equation}
Combining Eq. \eqref{eq:U} and \eqref{eq:V}, we get
\begin{equation} \label{eq:one-iterate}
    \| \hat{\bx}^s - \bx^* \|_2 \leq  (1- \frac{ \mu}{\mu+L})\| \hat{\bx}^{s-1} - \bx^* \|_2 + \frac{1}{L}\calO(\kappa), 
\end{equation}
then by iterating Eq. \eqref{eq:one-iterate}, we have
\begin{equation}
    \left\|\hat{\bx}^{s} - \bx^*\right\|_2 \leq (1-\frac{\mu}{\mu+L})^s\left\|\hat{\bx}^{0} - \bx^*\right\|_2 + \frac{2 \Gamma}{\mu},
\end{equation}
where $\Gamma = \calO(\kappa)$, which completes the proof.
\end{proof}
\begin{rmk}
Intuitively, Theorem \ref{thm:thm1} provides the convergence analysis and details the quantitative relation between the reconstruction upper bound and $\kappa$ representing the distance of $\nabla f_t( \hat{\bx}^s)$ and averaged gradients. It is noted that $\kappa$ tends to zero if multiple normal temporal information is homogeneous so that we could restore the private data fully \cite{li2019communication}. We provide the proof and extension to the Non-convex condition for $f_t()$ in Appendix A.
\end{rmk}

Moreover, our TGIAs-RO algorithm includes following four steps (see Algo. \ref{algo:TGIAs-RO} and Fig. \ref{fig: framework}): \\

\noindent\textbf{Step 1: Label Restoration and Gradient Alignment.} We firstly restore the label $\hat{\by}$ via label optimization methods according to model gradients and weights of the last fully-connected layer. Then we implement the gradient alignment to choose $T$ gradients with the same batch data from $\{g_1, \cdots, g_{T_0}\}$ (see details in Sect. \ref{label and alignment}). \\
\begin{rmk}
It should be noted that in Sect. \ref{sec:exp-label-batch}, there may be some errors in generating multiple gradients with the same batch data. However, according to Theorem \ref{thm:thm1}, the TGIAs-RO is not affected when the number of mistakes is below 50\%.
\end{rmk}
\noindent\textbf{Step 2: Optimization with Each Temporal Gradients.} Given $\hat{\by}$ fixed, we initialize the restored batch input as normal Gaussian distribution $\calN(0,1)$ and optimize $f_t(\hat{\bx}) = ||\nabla_{\bw_t}\calL(\bw_t, \hat{\bx}) - \nabla_{\bw_t}\calL(\bw_t, \bx)||^2$ over $T$ temporal gradients separately. Each temporal optimization output its own $\hat{\bx}_t$ after updating for $R_l$ iterations.

\begin{rmk}
\textbf{Step 2} is similar to the step of local updating in federated learning. The number of local iteration $R_l$ is regarded as "communications rounds" in FL, our empirical findings in Appendix B.2 show $R_l$ could not be too small and large. Small $R_l$ makes the TGIAs-RO Algorithm hard to converge, and large $R_l$ would lead to consume much time.
\end{rmk}

\noindent\textbf{Step 3: Robust Aggregation for $\{\hat{\bx}_t\}_{t=1}^T$.}
As mentioned in Sect. \ref{sec:failure}, some intermediate restored data $\{\hat{\bx}_t\}_{t=T-m+1}^T$ collapse in the optimization process, which destroys the optimization process when averaging all updated data. In order to solve this issue, we introduce robust statistics such as Krum \cite{blanchard2017machine}, coordinate-wise median and trimmed mean \cite{yin2018byzantine} to conduct robust aggregation. We output \text{Robust Aggregation} of $(\hat{\bx}_1,\cdots, \hat{\bx}_T)$ as follows:

\begin{equation}
    \text{RobustAgg}(\hat{\bx}_1,\cdots, \hat{\bx}_T) \longrightarrow  \hat{\bx}.
\end{equation}


\noindent\textbf{Step 4:} Take $\hat{\bx}$ as initialization, repeat Step 2 and  Step 3 for $R_g$ steps, see \ref{thm:thm1} for convergence analysis.

\begin{algorithm}[t]
\small 
\caption{\textbf{TGIAs-RO Algorithm}}
	\begin{algorithmic}[1]
		\renewcommand{\algorithmicrequire}{\textbf{Input:}}
		\renewcommand{\algorithmicensure}{\textbf{Output:}}
		\Require the loss function $\calL$ of the main task in FL; $T$ leaked gradients $\{\nabla_{\bw_t}\calL(\bw_t, \calD)\}_{t=1}^{T_0}$ and model weights $\{\bw_t\}_{t=1}^T$ for batch data $\{\calD:=(\bx_i, \by_i)\} _{i=1}^b$; TGIAs-RO local iterations $R_l$; TGIAs-RO global iterations $R_g$;
		local learning rate $\eta$.
		\Ensure Restored batch data $\hat{\calD}$. 
	\State \textbf{Step \RomanNumeralCaps{1}}:
		$\text{LabelInfer}(\{\nabla_{\bw_t}\calL(\bw_t, \calD)\}_{ t=1}^T) \rightarrow{} \hat{\by}$
		\State Initialize the image $\hat{\bx}^0 \leftarrow \calN(0,1)$, recovered label $\hat{\by}$.
		
		\For{$s=1,2,\cdots,R_g$}
			\State \textbf{Step \RomanNumeralCaps{2}}: Optimization for each temporal gradients.
		    \ForParallel{$t=1, 2, \cdots, T$}
		    \State $\nabla_{\bw_t} = \nabla_{\bw_t}\calL(\bw_t, \bx)$
		    \State $f_t(\hat{\bx}_t^{s-1}) =||\nabla_{\bw_t}\calL(\bw_t, \hat{\bx}^{s-1}) - \nabla_{\bw_t}||^2 $
		    \State $\hat{\bx}_t^{s-1,0}=\hat{\bx}^{s-1}$
		    \ForParallel{$k=1, 2, \cdots, R_l$}
		    \State $\hat{\bx}_t^{s-1,k} = \hat{\bx}^{s-1,k-1}_t- \eta \nabla_{\hat{\bx}_t^{s-1,k-1}}f_t(\hat{\bx}_t^{s-1,k-1})$
		    \EndFor
		    \State  $\hat{\bx}_t^{s}=\hat{\bx}_t^{s-1, R_l}$
		    \EndFor
		    \State  \textbf{Step \RomanNumeralCaps{3}}: Robust Aggregation for $\{\hat{\bx}_t^s\}_{t=1}^T$
		    \State $\text{RobustAgg}(\hat{\bx}^s_1,\cdots, \hat{\bx}^s_T) \longrightarrow  \hat{\bx}^{s}$.

		\EndFor
		\State  $\hat{\bx}=\hat{\bx}^{R_g}$ \\
		\Return $\hat{\calD}=(\hat{\bx}, \hat{\by})$.
	\end{algorithmic} 
	 \label{algo:TGIAs-RO}
\end{algorithm}



\subsection{Generation of Multiple Gradients}\label{label and alignment}
To obtain $T$ temporal gradients with the same batch data, we follow two procedures: (i) recovering the label, as explained in Sect. \ref{alignment: labels}; and (ii) selecting gradients that have similar labels and high cosine similarity, as described in Sect. \ref{sec:alignment: cos}. By applying these two steps, we obtain multiple temporal gradients, which are presented in Sect. \ref{sec:exp-label-batch}.


\subsubsection{Label Recovery}\label{alignment: labels}
First, we recover label according to last layer weights and gradients via \textit{label optimization}. Specifically, assume that $\bx^{FC}$ is the input to the fully connected layer in the network, which is noted as $\bw^{FC}$. The gradient $\nabla_{\bw^{FC}}\calL(\bw^{FC}, \bx^{FC})$ is known.  

\noindent\textbf{Dummy data initialization in FC layer.}  
We initialize the restored batch input $\hat{\bx}^{FC}$ in the FC layer as a normal Gaussian distribution $\calN(0,1)$. The dimension of $\hat{\bx}^{FC}$ is $\mathbb{R}^{b\times M}$, where $b$ is the batch size of private data $\bx$, and $M$ is the width of fully connected layers. And $\hat{\by}$ is randomly initialized as $\mathbb{R}^{b\times N}$, where $b$ is the batch size and $N$ is the number of classes. 

\noindent\textbf{Dummy label optimization.}  
We minimize 
\begin{equation}
\text{min}_{\hat{\bx}^{FC}, \hat{\by}} ||\nabla_{\bw^{FC}}\calL(\bw^{FC}, \hat{\bx}^{FC}, \hat{\by}) - \nabla_{\bw^{FC}}\calL(\bw^{FC}, \bx^{FC}, \by)||^2
\end{equation}
over fully connected layer $\bw^{FC}$.

\subsubsection{Gradient Alignment}\label{sec:alignment: cos}
Second, we integrate the computations of label and gradients similarity to facilitate the gradient alignment process. Specifically, we select gradients that correspond to the same batch of data, which is achieved by implementing two methods, namely gradient alignment with label similarity and gradient alignment with gradients similarity.

Gradient alignment with label similarity involves the first step of obtaining multiple groups of labels, followed by assigning the same labels to the same batch ids. On the other hand, gradient alignment with gradients similarity is introduced to tackle the issue of gradients from different data batches sharing the same labels. In order to address this problem, we select gradients with high cosine similarity, thereby identifying those that are associated with the same batch of data. Overall, by integrating these two steps, we are able to select multiple temporal gradients that are useful for our downstream tasks. The results of this approach are presented in Sect. \ref{sec:exp-label-batch}.

\section{Experimental Evaluation}\label{sec:exp}

This section evaluates the performance of the proposed TGIAs-RO in terms of image reconstruction quality and robustness under privacy-preserving strategies like differential privacy \cite{DLDP_Abadi16} and gradient sparsification \cite{wangni2018gradient}. The experimental results show that the proposed TGIAs-RO with only 10 temporal gradients  reconstruct private data with higher image quality than the state-of-the-art methods, especially 
for a large batch size (up to 128) and image size (3$\times$224$\times$224 pixels). And the proposed method effectively reconstructs images under strategies like differential privacy \cite{DLDP_Abadi16} and gradient sparsification \cite{wangni2018gradient}. 




\subsection{Experiment Settings}




\noindent\textbf{Baselines.} 
We compare our proposed method with five existing single-temporal GIAs methods: 
(1) \textbf{DLG \cite{zhu2019deep}:}  $l_2$ loss is originally adopted to reconstruct private data from leaked gradients. 
    (2) \textbf{Self-adaptive attack (SAPAG) \cite{wang2020sapag}:} Gaussian kernel based function is adopted as an alternative loss function instead of $l_2$ loss.
    (3) \textbf{Cosine  simlarity \cite{geiping2020inverting}:} Cosine similarity is adopted as an alternative loss function for $l_2$ loss.
    (4) \textbf{BN regularizer \cite{yin2021see}:} Statistics of the batch normalization layers are adopted as a  regularization term.
    (5) \textbf{GC regularizer \cite{yin2021see}:} 
    A group consistency regularizer is adopted as an auxiliary loss. For fair comparisons, all methods were run on the same suites of model architectures and datasets using batch sizes from 1 to 128. We refer reviewers to see more details in Appendix C.


    


\noindent\textbf{Datasets \& Network Architectures.} We conduct experiments on MNIST, CIFAR10, and ImageNet datasets \cite{krizhevsky2009learning, deng2009imagenet} in a horizontal FL setting. We refer readers to more experiments including text data in Appendix B. The network model architectures we investigate include well-known LeNet, AlexNet and ResNet18 \cite{krizhevsky2012imagenet, lecun1998gradient, he2016deep}.  
\noindent \textbf{Implementation Strategies.} (1) \textbf{Label Restoration:} Since label recovery is an independent task that
has been explored in \cite{zhao2020idlg, geiping2020inverting, yin2021see, geng2021towards}, we provide the results of label inference with multiple temporal information in Appendix B.1.  (2) \textbf{Layer Loss Alignment:}
We assign loss (Eq. \ref{eq:gradient-inversion-single}) of each layer with linearly increasing weights in TGIAs-RO \cite{wang2020sapag, geiping2020inverting}, in order to weight the losses in multiple layers.

\noindent\textbf{Evaluation Metrics.} We adopt mean squared error (MSE), peak signal to noise ratio (PSNR), and structural similarity (SSIM) as the metrics for reconstruction attacks on image data \cite{wang2004image}. Lower MSE, higher PSNR and SSIM of reconstructed images indicate better performance.


	
\subsection{Generation of Multiple Gradients} \label{sec:exp-label-batch}
We perform Label Restoration and gradient alignment as a preprocessing step before data recovery, as described in Sect. \ref{label and alignment}. TGIAs-RO is capable of restoring labels with an accuracy of over 97\%, and consequently, we can alignment batch ids with the restored labels and gradient similarity.

\subsubsection{Label Recovery}
The recovery accuracy of labels utilizing label optimization, as described in Sect. \ref{label and alignment}, is presented in Tab. \ref{tab: label}. The results demonstrate a remarkable recovery accuracy surpassing 97\% (e.g., achieving 99.71\% accuracy for a batch size of 32 and 100 classes). Notably, our proposed method outperforms the label recovery accuracy reported in \cite{yin2021see}, where the accuracy is only 88.65\% for the same experimental settings (i.e., a batch size of 32 and 100 classes). The high precision in label recovery significantly contributes to the selection of gradients with the same batch data, as detailed in the subsequent section.



\begin{table}[h!]
\caption{\label{tab: label} Label recovery accuracy with \cite{yin2021see} and with the proposed label optimization on ResNet-ImageNet of different batch sizes.}
 \renewcommand{\arraystretch}{0.9}
\centering
\resizebox{0.48\textwidth}{!}{
\begin{tabular}{ccccc}
\toprule
\multirow{2}{*}{ \tabincell{c}{Batch \\ Size} } &\multicolumn{2}{c}{100 Classes} &\multicolumn{2}{c}{1000 Classes}
\\ \cmidrule(r){2-3} \cmidrule(r){4-5}
& \cite{yin2021see} & Label Opimization &\cite{yin2021see} & Label Opimization    \\ \midrule
8 & 95.89\%    & 100\%    & 99.47\%    & 100\%    \\
16 &  91.84\%   & 99.86\%         & 99.37\%         & 100\%         \\ 
32 & 88.65\%    & 99.71\% & 99.19\% & 99.87\% \\
64 &  $\backslash$    & 98.56\%    & 98.21\%    & 99.47\%    \\
128 & $\backslash$   & 97.32\%    & 98.11\%    & 99.34\%    \\

 \bottomrule
\end{tabular}
}
\end{table}



\subsubsection{Gradient Alignment}
To accomplish gradient alignment, which entails identifying gradients belonging to the same batch data, we employ comparisons utilizing both label similarity and gradient similarity among all gradients. Initially, we cluster the temporal gradients based on the restored batch labels. However, due to the possibility of different batch data sharing the same labels, they may be erroneously grouped together in the same clusters. Therefore, to address this issue, we utilize cosine similarity among gradients to differentiate between those with the same labels but different time. The accuracy of gradient alignment is defined as the ratio of correctly clustered gradients to the total number of gradient clusters.


The experimental findings reported in Tab. \ref{tab: cosine similarity} for ResNet-ImageNet illustrate a decrease in gradient alignment accuracy as the batch size increases, indicating an enhanced ability to identify gradients associated with the same batch data. Notably, our proposed method, which incorporates comparisons of both labels and gradients, achieves a gradient alignment accuracy of 100\% for 100 classes and surpasses 76\% accuracy for 10 classes. Furthermore, the proposed TGIAs-RO is not affected since the error induced the generation of multiple gradients is less than 50\% according to Theorem \ref{thm:thm1}.



\begin{table}[h!]
\caption{\label{tab: cosine similarity} Gradient alignment accuracy with label similarity and gradient similarity. We separate ImageNet (ResNet) including 10000 samples with different batch size and further obtain multiple gradients from 5 different training epochs. For example, when the batch size is 16, the number of gradients we obtain is $10000/16*5=3125$.}
\renewcommand{\arraystretch}{1.1}
\centering
\resizebox{0.49\textwidth}{!}{

\begin{tabular}{ccccc}
\toprule
\multirow{3}{*}{ \tabincell{c}{Batch \\ Size} } &\multicolumn{2}{c}{10 Classes} &\multicolumn{2}{c}{100 Classes}
\\ \cmidrule(r){2-3} \cmidrule(r){4-5}
& \tabincell{c}{Label \\ Similarity} &\tabincell{c}{Label Similarity + \\ Gradient Similarity} &\tabincell{c}{Label \\ Similarity} & \tabincell{c}{Label Similarity + \\ Gradient Similarity}   \\ \midrule
16 &  68.84\%   & \textbf{76.32\%}         & 99.87\%         & \textbf{100\%}         \\ 
32 & 76.54\%    & \textbf{87.93\%}  & 99.99\% & \textbf{100\%} \\
64 &  100\%    & \textbf{100\%}   & 100\%    & \textbf{100\%}    \\

 \bottomrule


\end{tabular}
}
\end{table}

\subsection{TGIAs-RO  v.s. the \textit{State-of-the-art} Methods} \label{subsec:exp-compare}

\begin{figure*}[t] 
	        	
	
	\centering
	 \begin{subfigure}{0.31\textwidth}
		\centering
		\includegraphics[keepaspectratio=true, width=160pt]{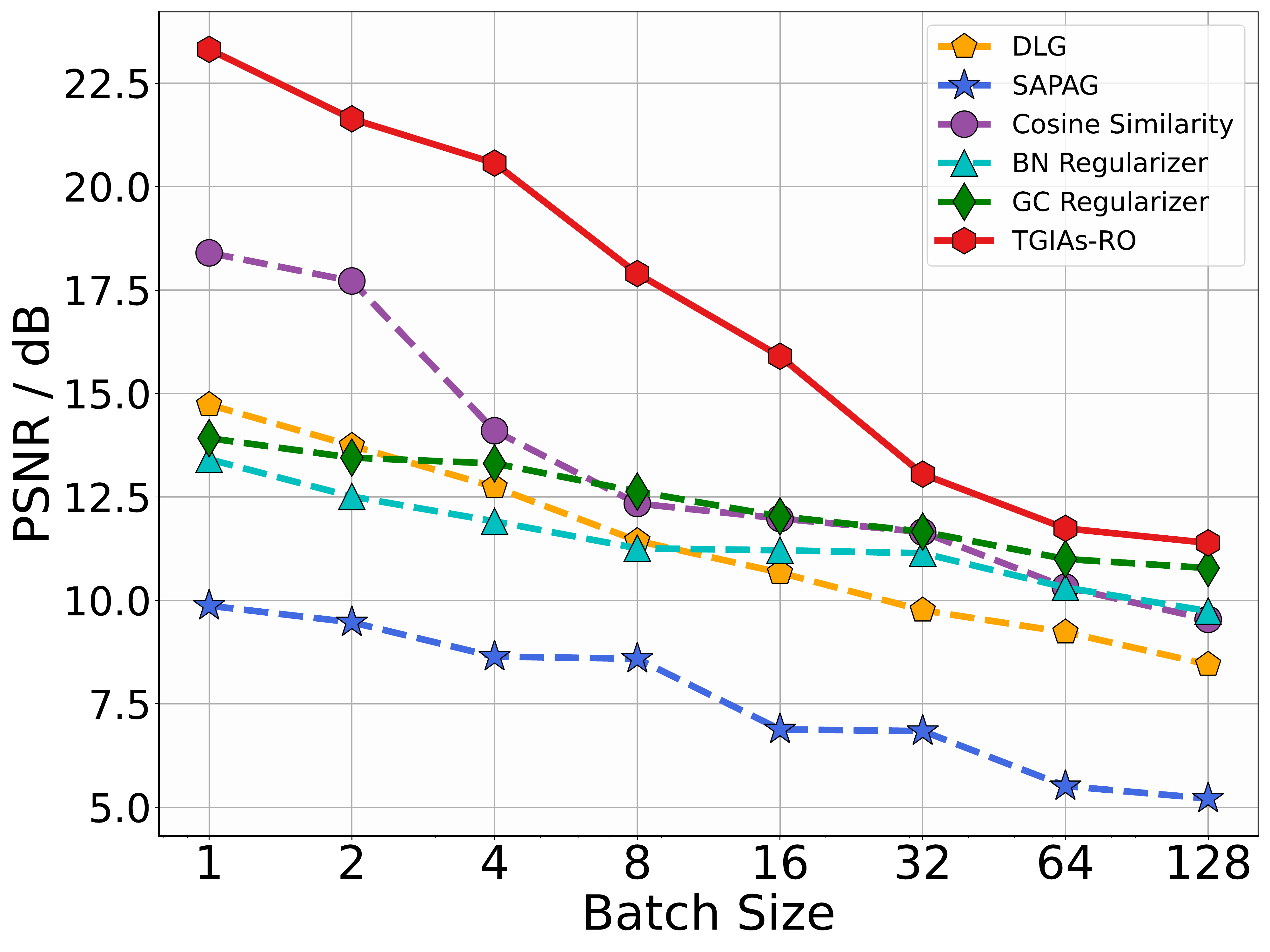}
		\subcaption{\small PSNR of MNIST}
	  \end{subfigure}
	 \begin{subfigure}{0.31\textwidth}
		\centering
		\includegraphics[keepaspectratio=true, width=160pt]{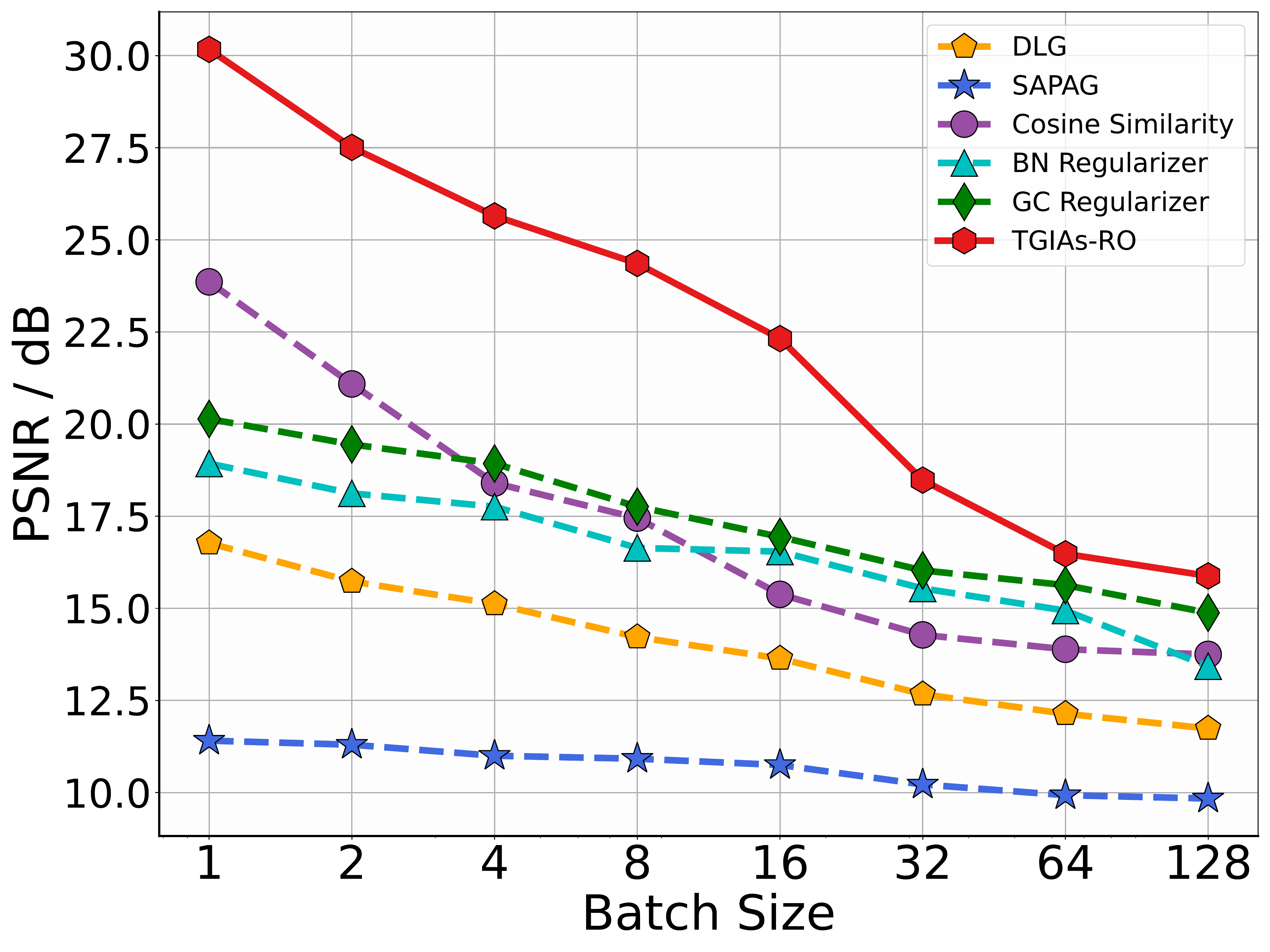}
		\subcaption{\small PSNR of CIFAR10}
	  \end{subfigure}
	  	 \begin{subfigure}{0.31\textwidth}
		\centering
		\includegraphics[keepaspectratio=true, width=160pt]{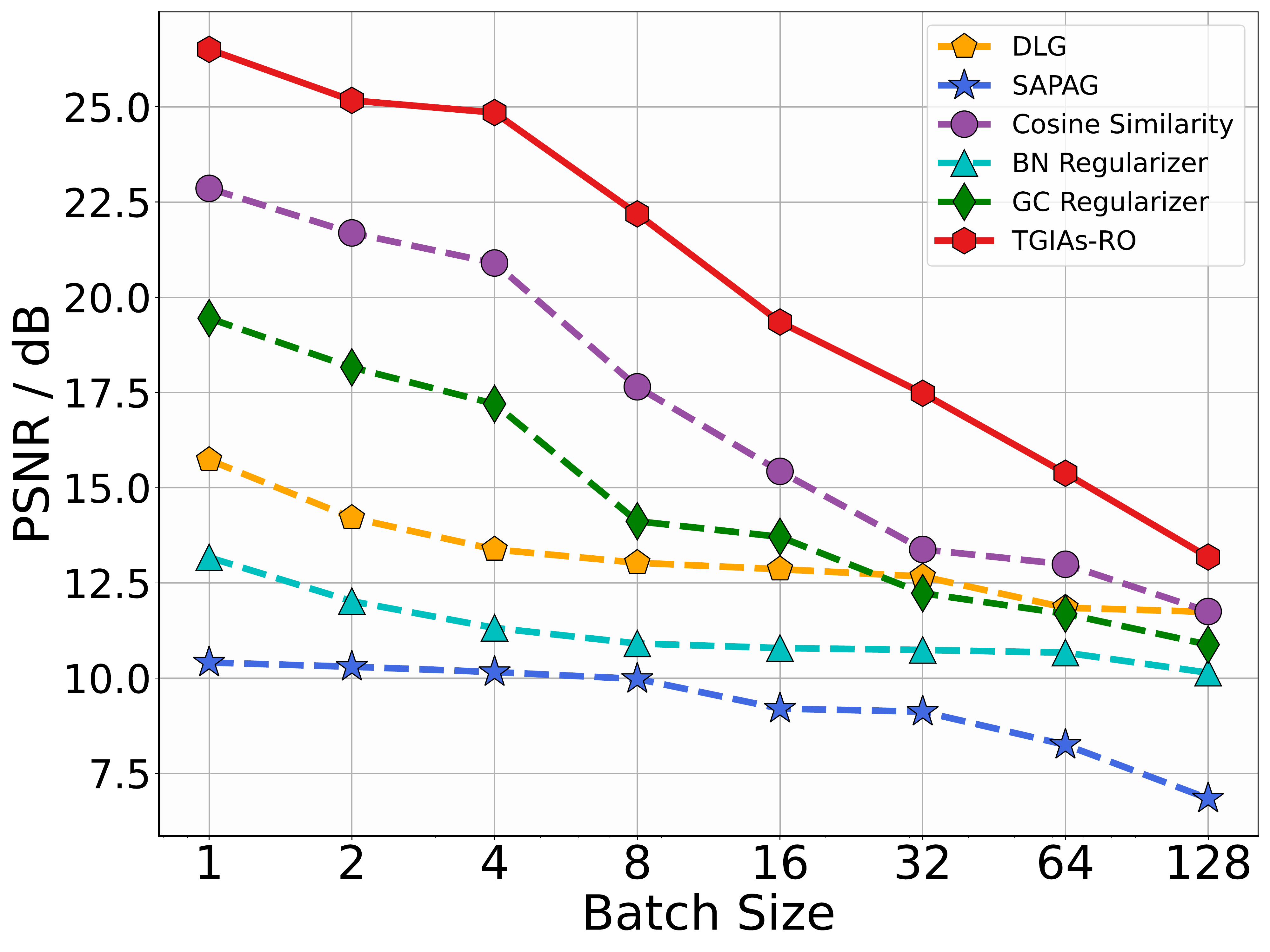}
		\subcaption{\small PSNR of ImageNet}
	  \end{subfigure}

	  \begin{subfigure}{0.31\textwidth}
		\centering
		\includegraphics[keepaspectratio=true, width=160pt]{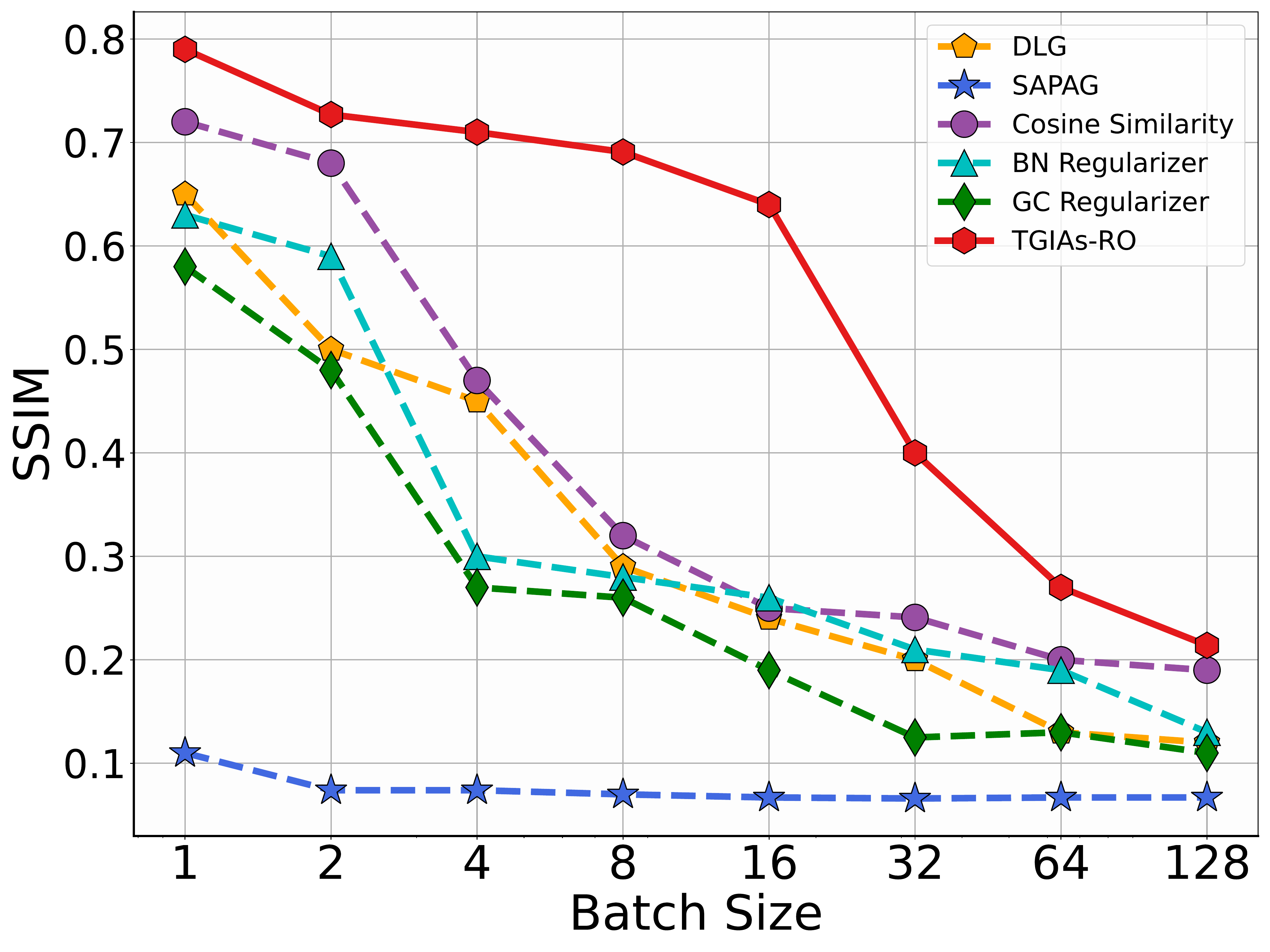}
		\subcaption{\small SSIM of MNIST}
	  \end{subfigure}
	 \begin{subfigure}{0.31\textwidth}
		\centering
		\includegraphics[keepaspectratio=true, width=160pt]{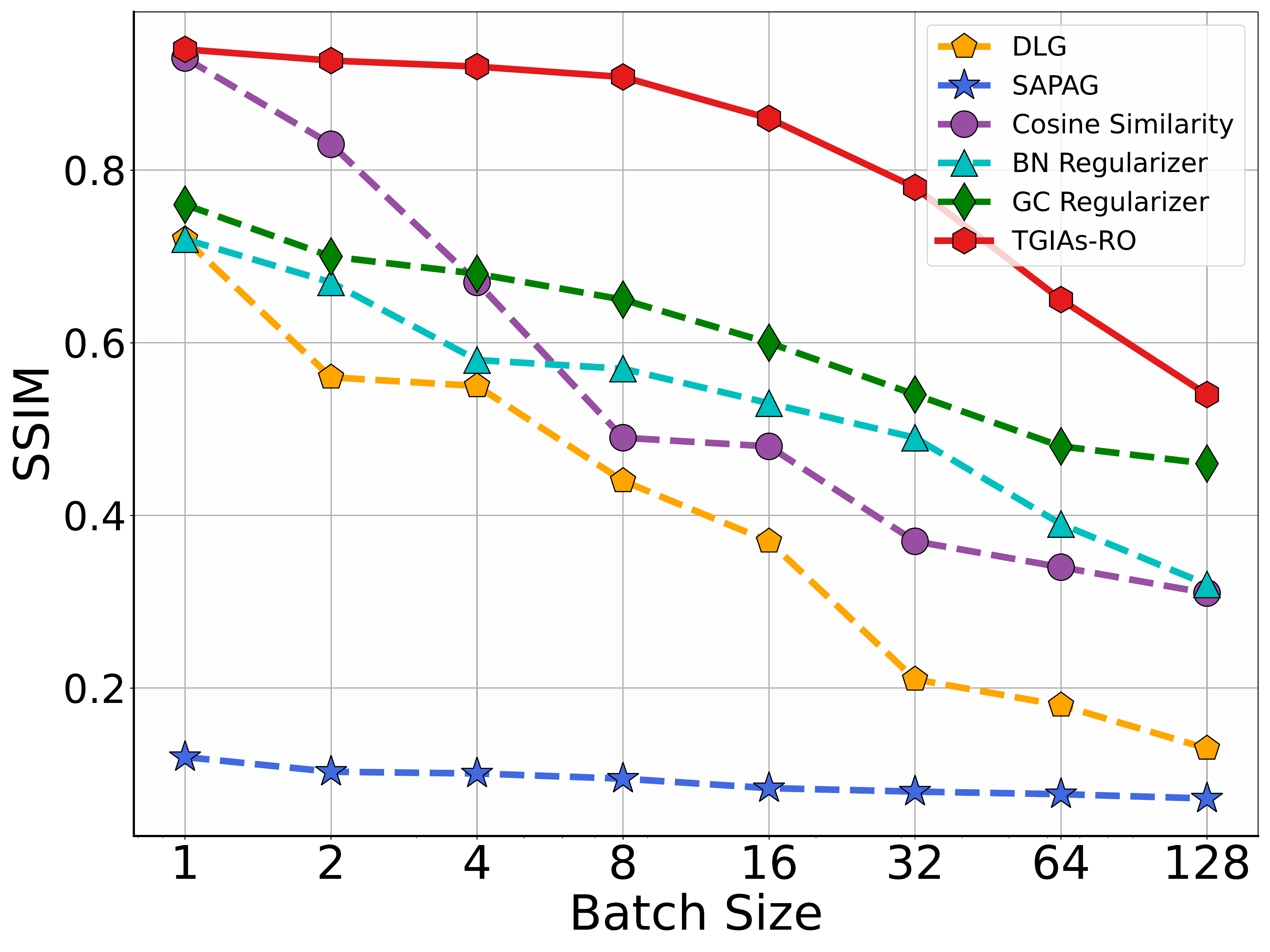}
		\subcaption{\small SSIM of CIFAR10}
	  \end{subfigure}
	  	 \begin{subfigure}{0.31\textwidth}
		\centering
		\includegraphics[keepaspectratio=true, width=160pt]{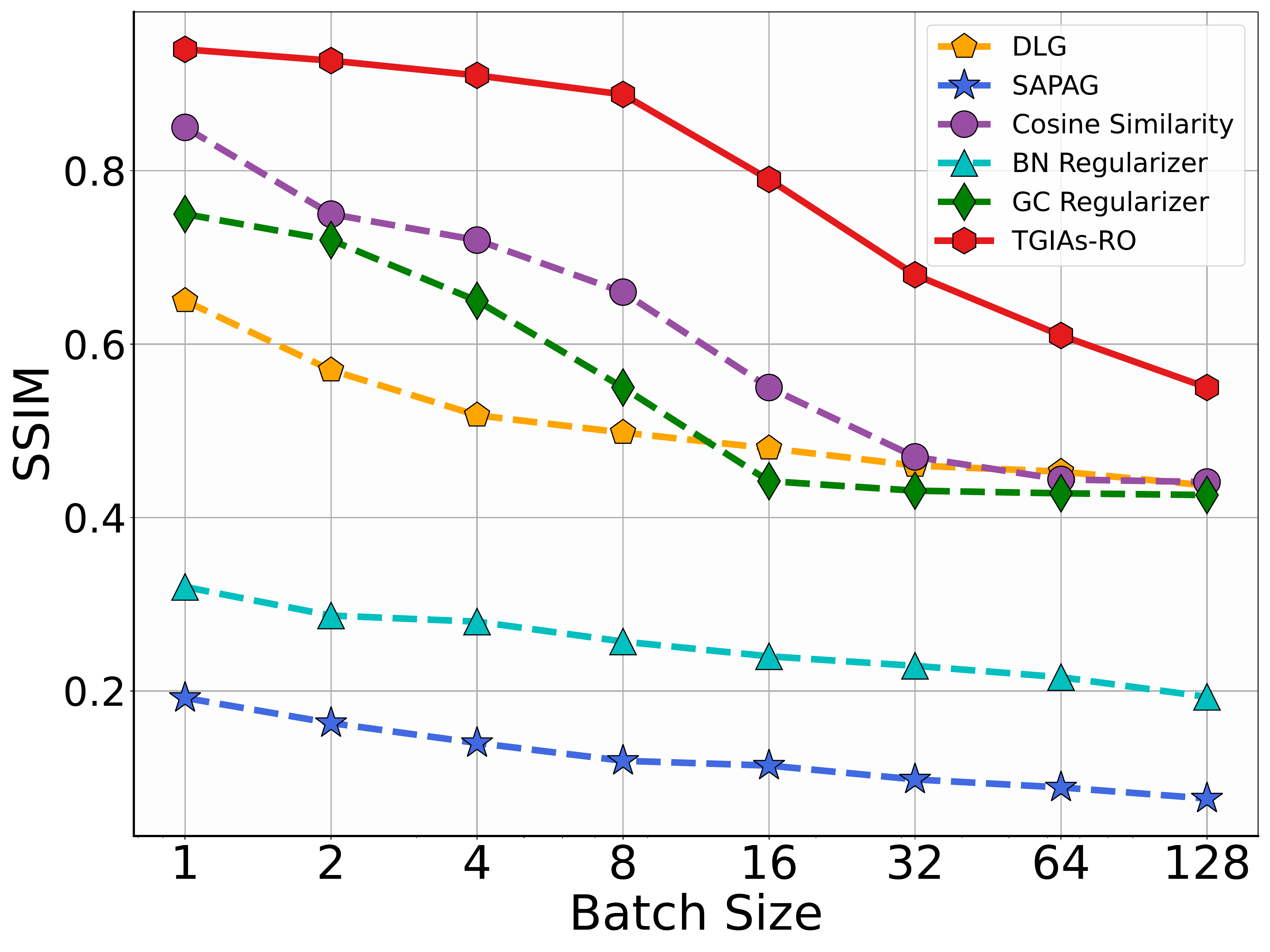}
		\subcaption{\small SSIM of ImageNet}
		
	  \end{subfigure}	
    
	\centering
	\caption{Comparisons between TGIAs-RO (red line) with 10 temporal gradients and the \textit{state-of-the-art} single-temporal GIAs (DLG \cite{zhu2019deep}, Cosine similarity \cite{geiping2020inverting}, SAPAG \cite{wang2020sapag}, BN regularzier \cite{yin2021see} and GC regularizer \cite{yin2021see}) methods with varying batch sizes in different datasets and model architectures.}\label{fig: line}
\end{figure*}

\begin{figure*}[t]

 \begin{subfigure}[t]{1\linewidth}
\centering
\begin{minipage}[c]{0.095\textwidth}
\vspace{-4em} Ground\\ truth 
\end{minipage}
\begin{minipage}[t]{0.095\textwidth}
\centering
\includegraphics[width=1.3cm]{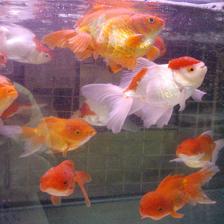}
\end{minipage}
\begin{minipage}[t]{0.095\textwidth}
\centering
\includegraphics[width=1.3cm]{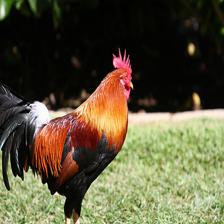}
\end{minipage}
\begin{minipage}[t]{0.095\textwidth}
\centering
\includegraphics[width=1.3cm]{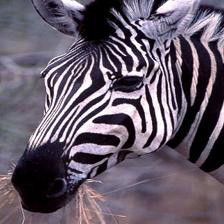}
\end{minipage}
\begin{minipage}[t]{0.095\textwidth}
\centering
\includegraphics[width=1.3cm]{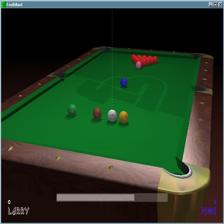}
\end{minipage}
\begin{minipage}[t]{0.095\textwidth}
\centering
\includegraphics[width=1.3cm]{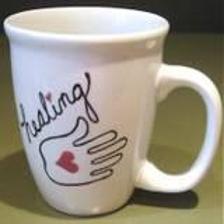}
\end{minipage}
\begin{minipage}[t]{0.095\textwidth}
\centering
\includegraphics[width=1.3cm]{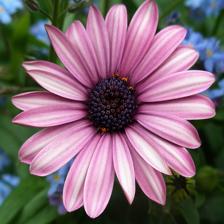}
\end{minipage}
\begin{minipage}[t]{0.095\textwidth}
\centering
\includegraphics[width=1.3cm]{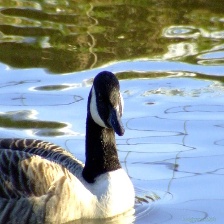}
\end{minipage}
\begin{minipage}[t]{0.095\textwidth}
\centering
\includegraphics[width=1.3cm]{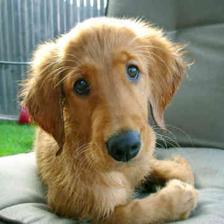}
\end{minipage}
\end{subfigure}
 
 \begin{subfigure}[t]{1\linewidth}
\centering
\begin{minipage}[c]{0.095\textwidth}
\vspace{-4em} TGIAs-RO 
\end{minipage}
\begin{minipage}[t]{0.095\textwidth}
\centering
\includegraphics[width=1.3cm]{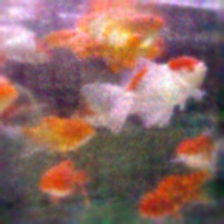}
\end{minipage}
\begin{minipage}[t]{0.095\textwidth}
\centering
\includegraphics[width=1.3cm]{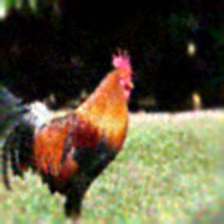}
\end{minipage}
\begin{minipage}[t]{0.095\textwidth}
\centering
\includegraphics[width=1.3cm]{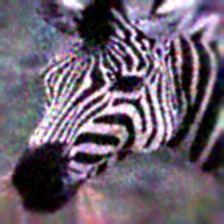}
\end{minipage}
\begin{minipage}[t]{0.095\textwidth}
\centering
\includegraphics[width=1.3cm]{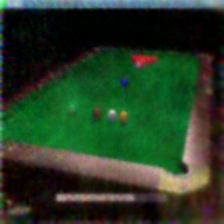}
\end{minipage}
\begin{minipage}[t]{0.095\textwidth}
\centering
\includegraphics[width=1.3cm]{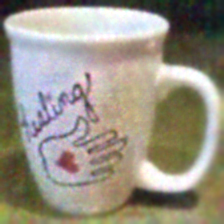}
\end{minipage}
\begin{minipage}[t]{0.095\textwidth}
\centering
\includegraphics[width=1.3cm]{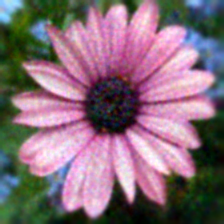}
\end{minipage}
\begin{minipage}[t]{0.095\textwidth}
\centering
\includegraphics[width=1.3cm]{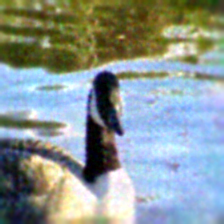}
\end{minipage}
\begin{minipage}[t]{0.095\textwidth}
\centering
\includegraphics[width=1.3cm]{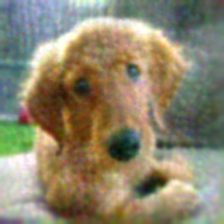}
\end{minipage}
\end{subfigure}

 \begin{subfigure}[t]{1\linewidth}
\centering
\begin{minipage}[c]{0.095\textwidth}
\vspace{-4em} Cosine\\ similarity 
\end{minipage}
\begin{minipage}[t]{0.095\textwidth}
\centering
\includegraphics[width=1.3cm]{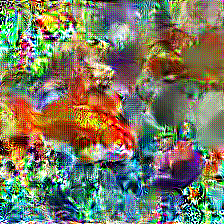}
\end{minipage}
\begin{minipage}[t]{0.095\textwidth}
\centering
\includegraphics[width=1.3cm]{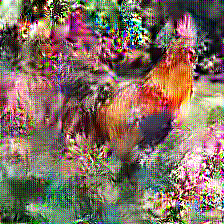}
\end{minipage}
\begin{minipage}[t]{0.095\textwidth}
\centering
\includegraphics[width=1.3cm]{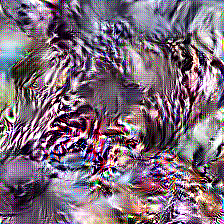}
\end{minipage}
\begin{minipage}[t]{0.095\textwidth}
\centering
\includegraphics[width=1.3cm]{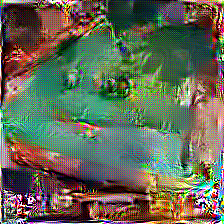}
\end{minipage}
\begin{minipage}[t]{0.095\textwidth}
\centering
\includegraphics[width=1.3cm]{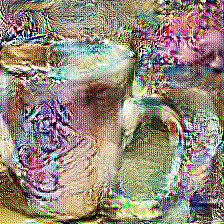}
\end{minipage}
\begin{minipage}[t]{0.095\textwidth}
\centering
\includegraphics[width=1.3cm]{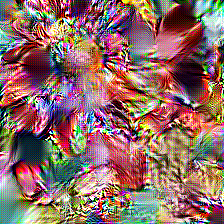}
\end{minipage}
\begin{minipage}[t]{0.095\textwidth}
\centering
\includegraphics[width=1.3cm]{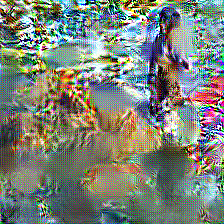}
\end{minipage}
\begin{minipage}[t]{0.095\textwidth}
\centering
\includegraphics[width=1.3cm]{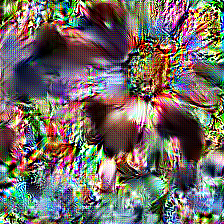}
\end{minipage}
\end{subfigure}

 \begin{subfigure}[t]{1\linewidth}
\centering
\begin{minipage}[c]{0.095\textwidth}
\vspace{-4em} GC\\ regularizer 
\end{minipage}
\begin{minipage}[t]{0.095\textwidth}
\centering
\includegraphics[width=1.3cm]{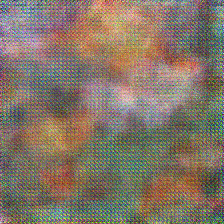}
\end{minipage}
\begin{minipage}[t]{0.095\textwidth}
\centering
\includegraphics[width=1.3cm]{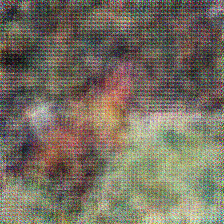}
\end{minipage}
\begin{minipage}[t]{0.095\textwidth}
\centering
\includegraphics[width=1.3cm]{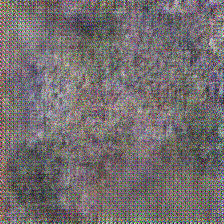}
\end{minipage}
\begin{minipage}[t]{0.095\textwidth}
\centering
\includegraphics[width=1.3cm]{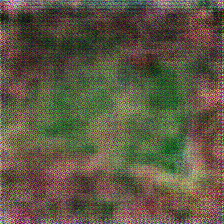}
\end{minipage}
\begin{minipage}[t]{0.095\textwidth}
\centering
\includegraphics[width=1.3cm]{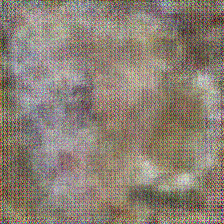}
\end{minipage}
\begin{minipage}[t]{0.095\textwidth}
\centering
\includegraphics[width=1.3cm]{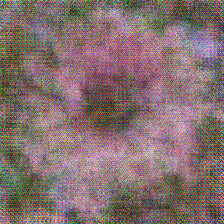}
\end{minipage}
\begin{minipage}[t]{0.095\textwidth}
\centering
\includegraphics[width=1.3cm]{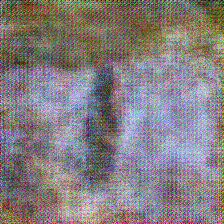}
\end{minipage}
\begin{minipage}[t]{0.095\textwidth}
\centering
\includegraphics[width=1.3cm]{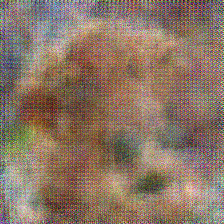}
\end{minipage}
\end{subfigure}

 \begin{subfigure}[t]{1\linewidth}
\centering
\begin{minipage}[c]{0.095\textwidth}
\vspace{-4em} BN\\ Regularizer 
\end{minipage}
\begin{minipage}[t]{0.095\textwidth}
\centering
\includegraphics[width=1.3cm]{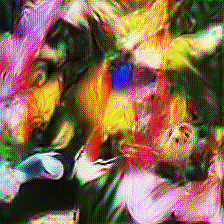}
\end{minipage}
\begin{minipage}[t]{0.095\textwidth}
\centering
\includegraphics[width=1.3cm]{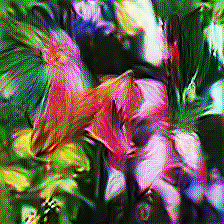}
\end{minipage}
\begin{minipage}[t]{0.095\textwidth}
\centering
\includegraphics[width=1.3cm]{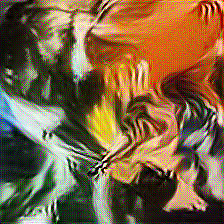}
\end{minipage}
\begin{minipage}[t]{0.095\textwidth}
\centering
\includegraphics[width=1.3cm]{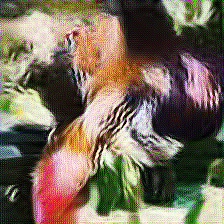}
\end{minipage}
\begin{minipage}[t]{0.095\textwidth}
\centering
\includegraphics[width=1.3cm]{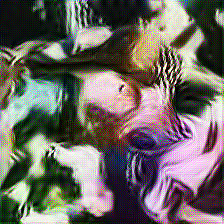}
\end{minipage}
\begin{minipage}[t]{0.095\textwidth}
\centering
\includegraphics[width=1.3cm]{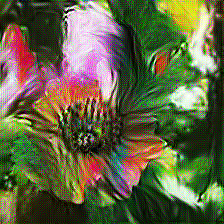}
\end{minipage}
\begin{minipage}[t]{0.095\textwidth}
\centering
\includegraphics[width=1.3cm]{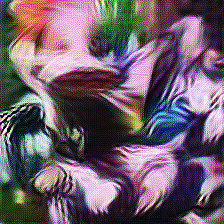}
\end{minipage}
\begin{minipage}[t]{0.095\textwidth}
\centering
\includegraphics[width=1.3cm]{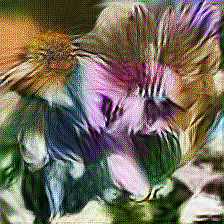}
\end{minipage}
\end{subfigure}

 \begin{subfigure}[t]{1\linewidth}
\centering
\begin{minipage}[c]{0.095\textwidth}
\vspace{-4em} DLG 
\end{minipage}
\begin{minipage}[t]{0.095\textwidth}
\centering
\includegraphics[width=1.3cm]{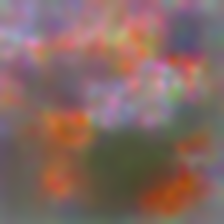}
\end{minipage}
\begin{minipage}[t]{0.095\textwidth}
\centering
\includegraphics[width=1.3cm]{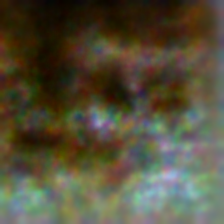}
\end{minipage}
\begin{minipage}[t]{0.095\textwidth}
\centering
\includegraphics[width=1.3cm]{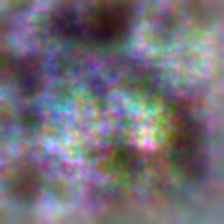}
\end{minipage}
\begin{minipage}[t]{0.095\textwidth}
\centering
\includegraphics[width=1.3cm]{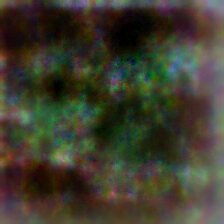}
\end{minipage}
\begin{minipage}[t]{0.095\textwidth}
\centering
\includegraphics[width=1.3cm]{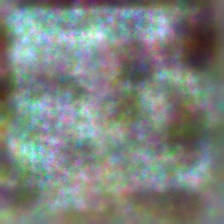}
\end{minipage}
\begin{minipage}[t]{0.095\textwidth}
\centering
\includegraphics[width=1.3cm]{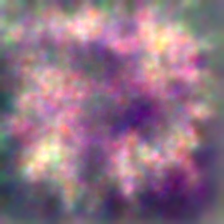}
\end{minipage}
\begin{minipage}[t]{0.095\textwidth}
\centering
\includegraphics[width=1.3cm]{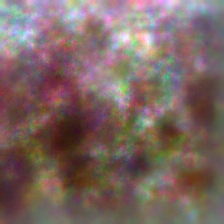}
\end{minipage}
\begin{minipage}[t]{0.095\textwidth}
\centering
\includegraphics[width=1.3cm]{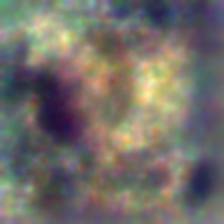}
\end{minipage}
\end{subfigure}

 \begin{subfigure}[t]{1\linewidth}
\centering
\begin{minipage}[c]{0.095\textwidth}
\vspace{-4em} SAPAG 
\end{minipage}
\begin{minipage}[t]{0.095\textwidth}
\centering
\includegraphics[width=1.3cm]{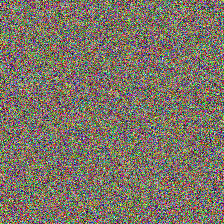}
\end{minipage}
\begin{minipage}[t]{0.095\textwidth}
\centering
\includegraphics[width=1.3cm]{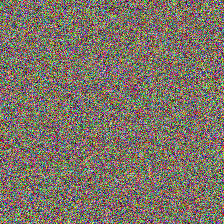}
\end{minipage}
\begin{minipage}[t]{0.095\textwidth}
\centering
\includegraphics[width=1.3cm]{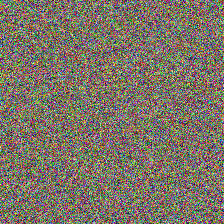}
\end{minipage}
\begin{minipage}[t]{0.095\textwidth}
\centering
\includegraphics[width=1.3cm]{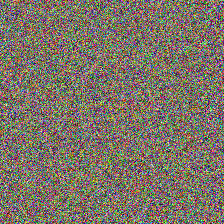}
\end{minipage}
\begin{minipage}[t]{0.095\textwidth}
\centering
\includegraphics[width=1.3cm]{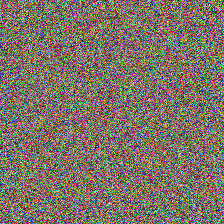}
\end{minipage}
\begin{minipage}[t]{0.095\textwidth}
\centering
\includegraphics[width=1.3cm]{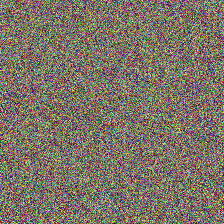}
\end{minipage}
\begin{minipage}[t]{0.095\textwidth}
\centering
\includegraphics[width=1.3cm]{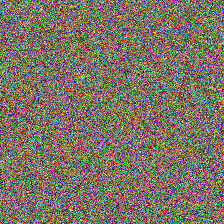}
\end{minipage}
\begin{minipage}[t]{0.095\textwidth}
\centering
\includegraphics[width=1.3cm]{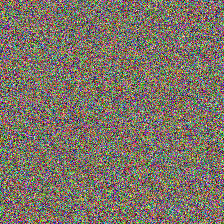}
\end{minipage}
\end{subfigure}

  \caption{Visual comparisons between TGIAs-RO (our method with 10 temporal gradients) with the state-of-the-art single-temporal gradient inversion 
attacks including DLG \cite{zhu2019deep}, Cosine similarity \cite{geiping2020inverting}, SAPAG \cite{wang2020sapag}, BN regularizer and GC
regularizer \cite{yin2021see} on ImageNet dataset (batch size = 8, image size $= 3 \times 224 \times 224$).} 
  \label{fig:rec_imgs}
  
\end{figure*}

Fig. \ref{fig: line} and \ref{fig:rec_imgs} indicates that the performance of TGIAs-RO with 10 temporal gradients and the median strategy surpasses all the \textit{state-of-the-art} methods. For example, in Fig. \ref{fig: line}(f), when reconstructing a batch of 64 ImageNet images, the SSIM and PSNR of baseline methods fall below 0.6 and 15dB respectively, whereas our proposed TGIAs-RO keeps SSIM above 0.6 and PSNR greater than 15dB. As an example, our proposed method TGIAs-RO has the highest image quality and the least amount of noise. Furthermore, we incorporate prior knowledge, such as image smoothness \cite{geiping2020inverting} and group consistency \cite{yin2018byzantine}, into our TGIAs-RO framework, which improves the reconstruction in Appendix B.5.


\subsection{Robustness of TGIAs-RO under Various Strategies in FL} \label{subsec:exp-robust}
In this section, we provide empirical results to validate our proposed TGIAs-RO is robust under various strategies in FL, i.e., client sampling to improve the communication efficiency and privacy-preserving mechanisms.

\subsubsection{TGIAs-RO under Client Sampling}
In the realistic FL scenarios, for communication efficiency concerns, only a small fraction of clients are selected at each iteration to update the global model \cite{mcmahan2017communication}, and each client only adopts a single batch for one iteration training. To collect enough temporal gradients for a single batch, we propose a two-stage batch data alignment technique before image reconstruction: 
\begin{itemize}
    \item  We first conduct label restoration \cite{yin2021see} according to multiple temporal gradients. Results in Sect. B.1 show that the ground truth label of private batch data are reconstructed with almost 100\% accuracy. 
    \item In order to align the multiple temporal gradients for one common batch data, we then implement batch id alignment according to the ground truth data label, and we refer readers to Sect. B.1 
for details. 
\end{itemize}

For the first 10 gradients with the same data batch (not consecutive), we conduct TGIAs-RO according to the aligned gradients. We simulate a federated learning setting with 20 clients, each client with 100 data points. The results presented in Tab. \ref{tab: fraction} show that our proposed TGIAs-RO is persistent under the client sampling strategy. The PSNR of image reconstruction only decades from 24dB to 22dB (less than 2 dB) when the client sampling ratio goes small as 0.2. 

\begin{table}[htbp]
\caption{Comparisons of restored images with different ratios of client sampling, the model architecture adopted is AlexNet and dataset is CIFAR10 (batch size = 8).}\label{tab: fraction}
 \renewcommand{\arraystretch}{0.95}
 \centering
\resizebox{0.46\textwidth}{!}{
\begin{tabular}{ccccc}
\toprule
\multirow{2}{*}{ Client Number}  &
\multirow{2}{*}{ Fraction Ratio}  &\multicolumn{3}{c}{Metrics} \\ \cmidrule(r){3-5}
& & MSE & PSNR/dB & SSIM         \\ \midrule
20 & 1 & 5$e^{-3}$  &  24.36  &  0.92\\
20 & 0.8 & 5$e^{-3}$  &  24.33   &  0.91\\
20 & 0.5 & 5$e^{-3}$  &  24.12   &  0.91\\
20 & 0.2 & 7$e^{-3}$ &  22.32   &  0.86\\
20 & 0.1 & 7$e^{-3}$ &  21.80   &  0.84\\
 \bottomrule
\end{tabular}
}
\end{table}

\subsubsection{TGIAs-RO under Privacy-preserving Methods}
\begin{figure}[htbp] 

	 \begin{subfigure}{0.24\textwidth}
  \centering
\includegraphics[keepaspectratio=true, width=120pt]{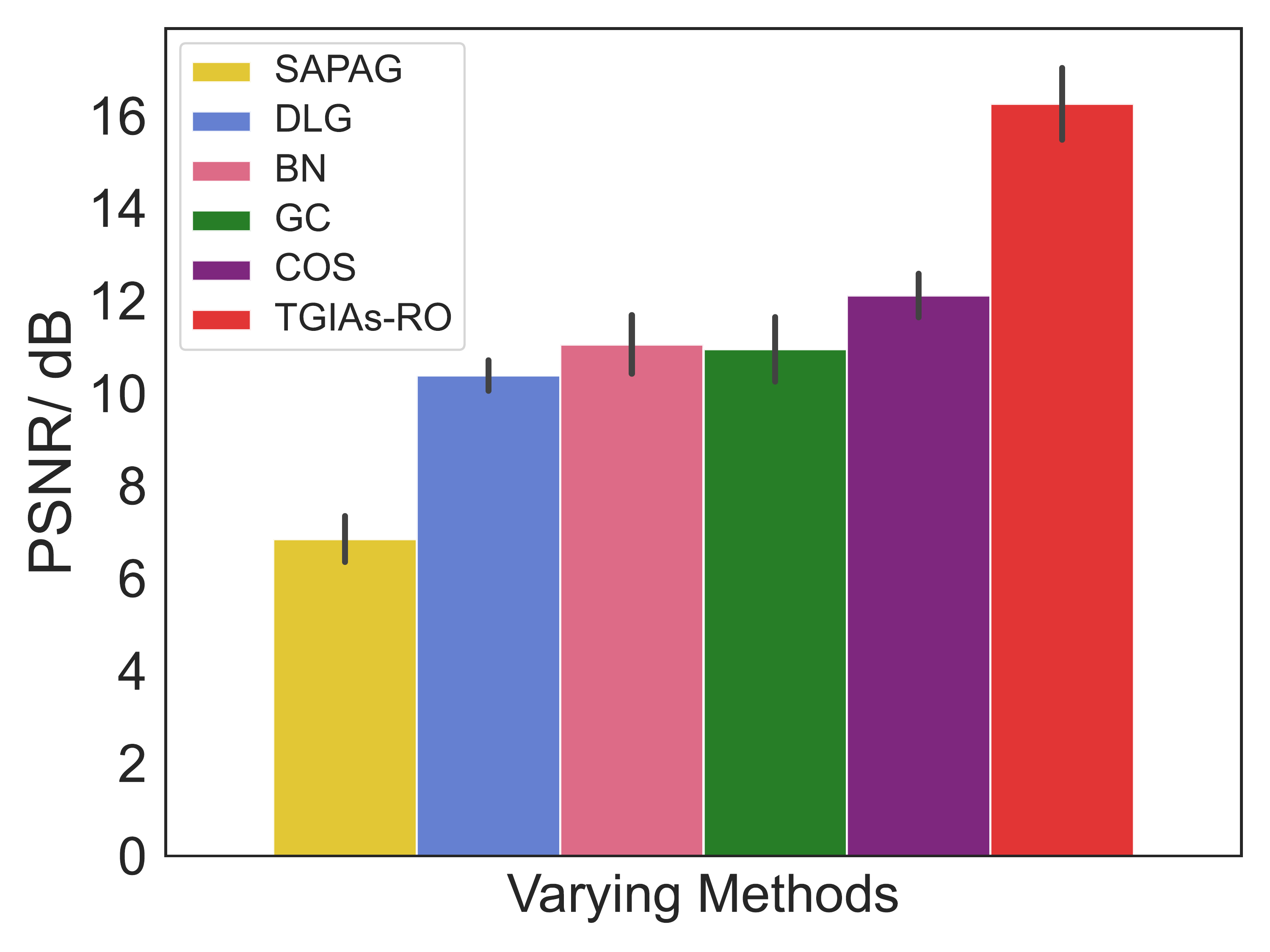}
		\subcaption{\small PNSR of DP }
	  \end{subfigure}
	 \begin{subfigure}{0.24\textwidth}
  \centering
\includegraphics[keepaspectratio=true, width=120pt]{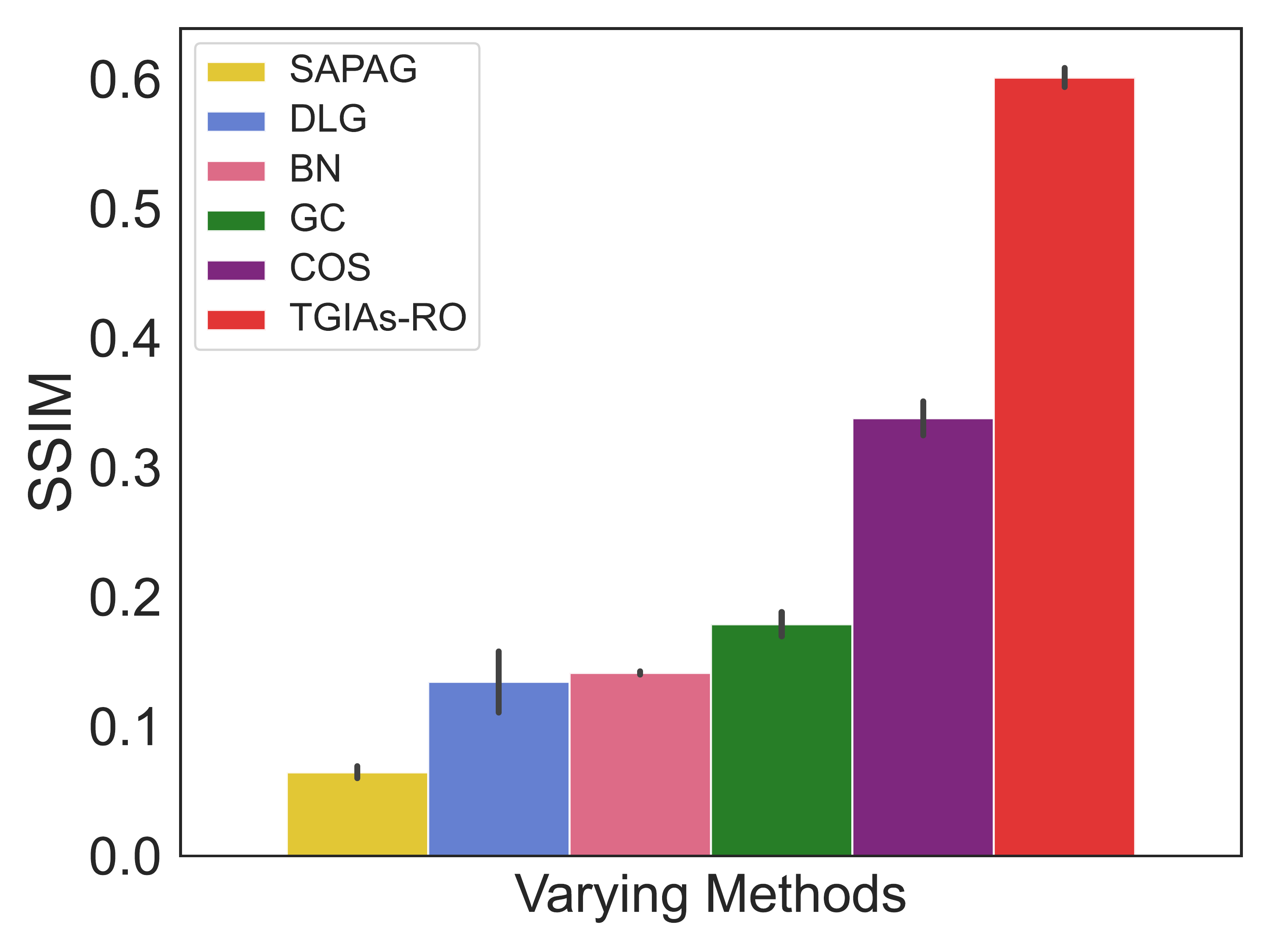}
		\subcaption{\small SSIM of DP}
	  \end{subfigure}
   	 \begin{subfigure}{0.24 \textwidth}
     
     \centering
\includegraphics[keepaspectratio=true, width=120pt]{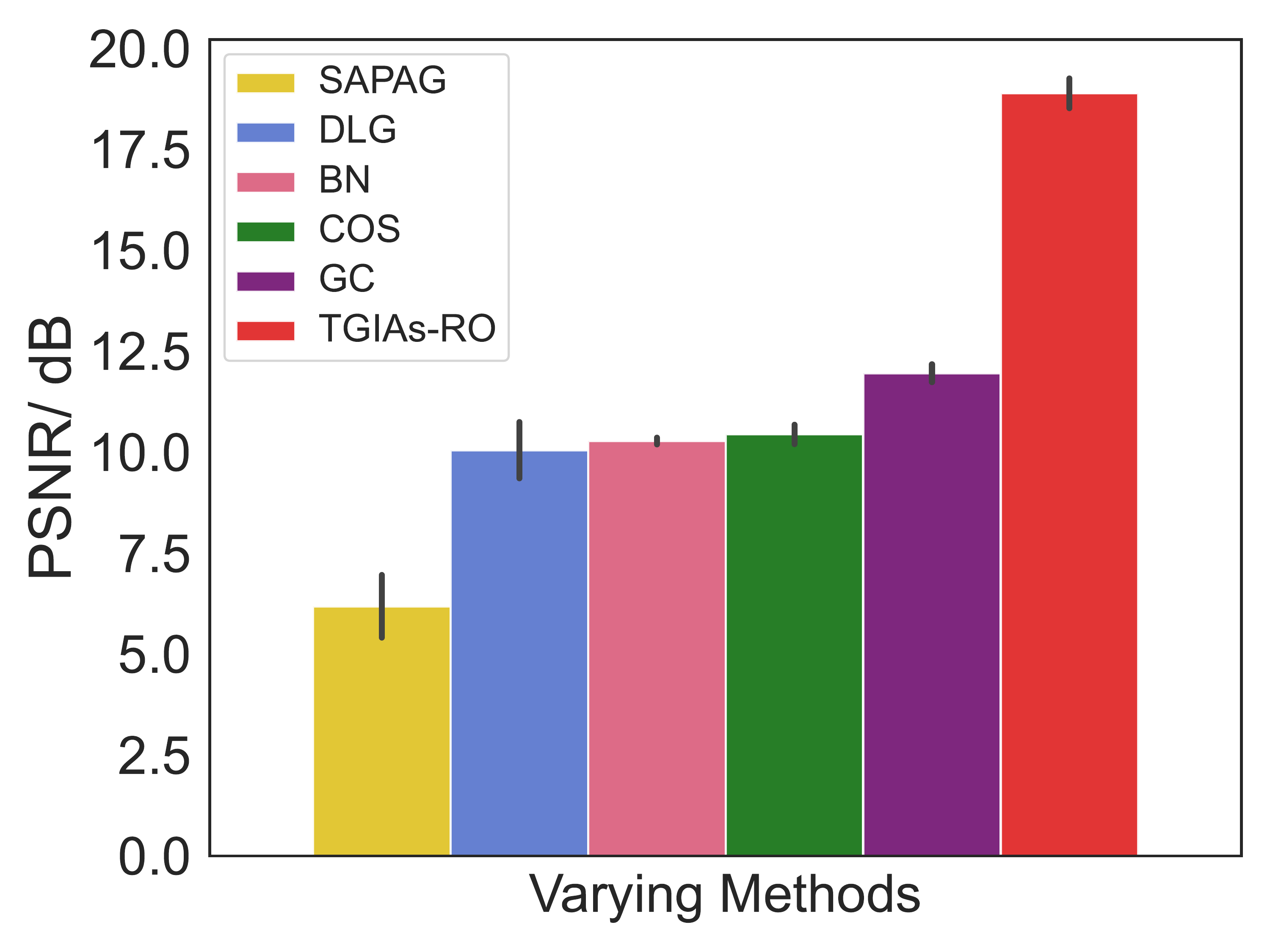}
		\subcaption{\small PNSR of sparsification }
	  \end{subfigure}
	 \begin{subfigure}{0.24\textwidth}
            \centering
		\includegraphics[keepaspectratio=true, width=120pt]{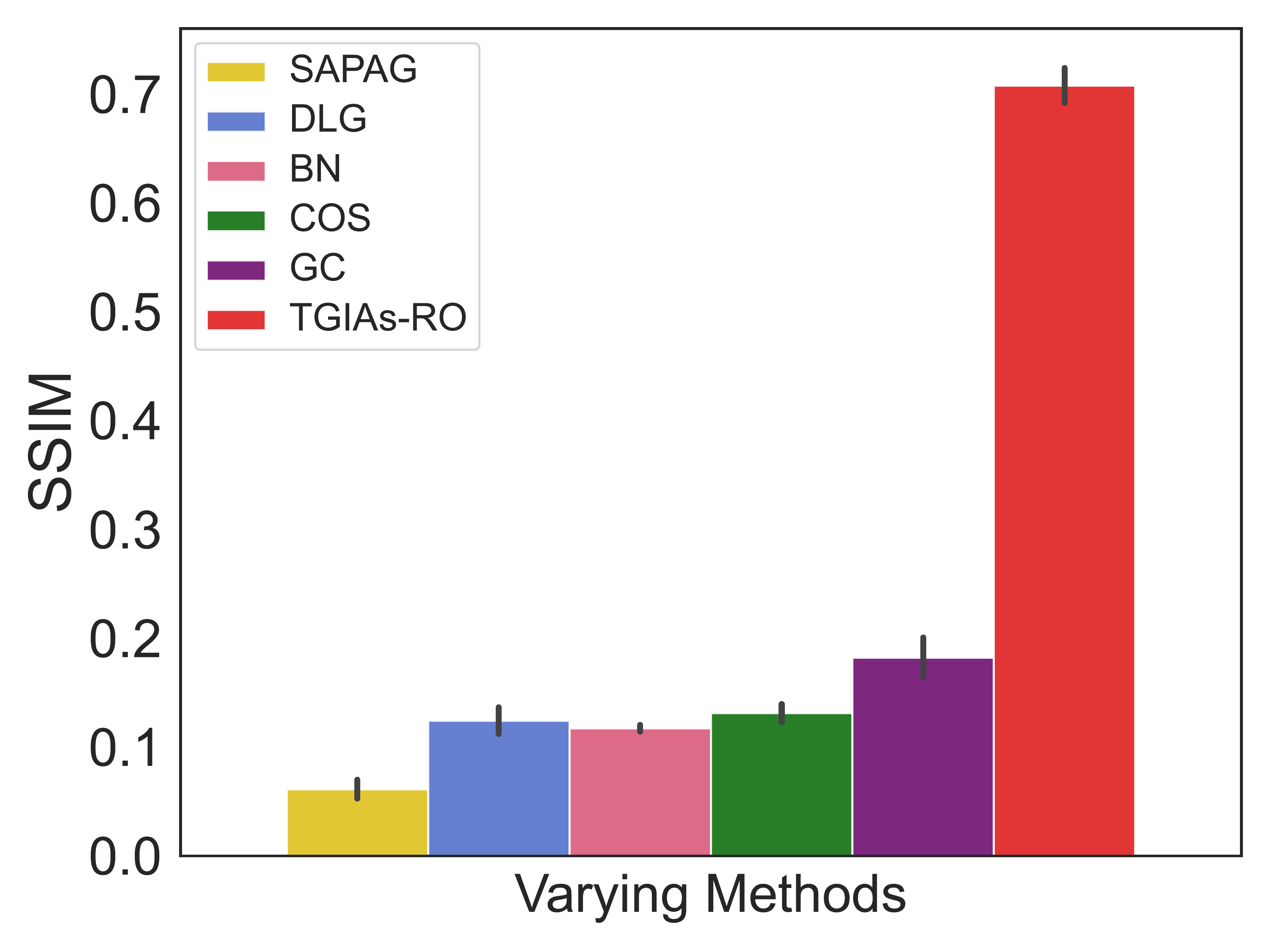}
		\subcaption{\small SSIM of sparsification}
	  \end{subfigure}
	\centering
   
	\centering
	\caption{Figure (a)-(b) present different GIAs methods under differential privacy ($\calN(0, e^{\text{-4}})$). Figure (c)-(d) present the comparisons against gradient sparsification techniques ($p = 40\%$ of the gradient parameters are masked to 0). 
 The model is AlexNet and the dataset is ImageNet in a batch size of 8.}\label{fig: bar_dp}
\end{figure}

We report the performance of TGIAs-RO and the baseline methods against privacy-preserving methods including differential privacy \cite{DLDP_Abadi16} (uploading the model gradients with Gaussian noise $\calN(0,\sigma^2)$) and gradient sparsification \cite{wangni2018gradient} (uploading partial gradients). 
Fig. \ref{fig: bar_dp}(a)-(b) show TGIAs-RO achieves higher PSNR and SSIM than the other methods under Gaussian noise. And Fig. \ref{fig: bar_dp}(c)-(d) present that the proposed TGIAs-RO is much more robust against gradient sparsification (e.g., the SSIM of is beyond 0.7 while others are less than 0.2 when 40\% gradients are masked to zero), we ascribe the robustness of TGIAs-RO against privacy-preserving methods to the temporal information adopted.

\begin{figure} [h]

\begin{subfigure}{0.50\textwidth}
 \centering
  \includegraphics[keepaspectratio=true, width=240pt]{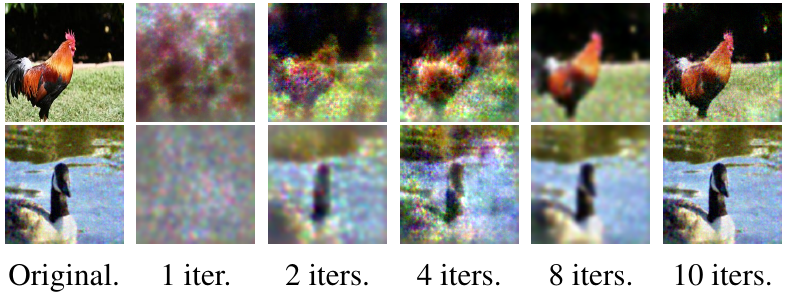}
		\subcaption{\small Visual comparisons of restored images}
	  \end{subfigure}
	 \begin{subfigure}{0.24\textwidth}
		\centering
      
  \includegraphics[keepaspectratio=true, width=120pt]{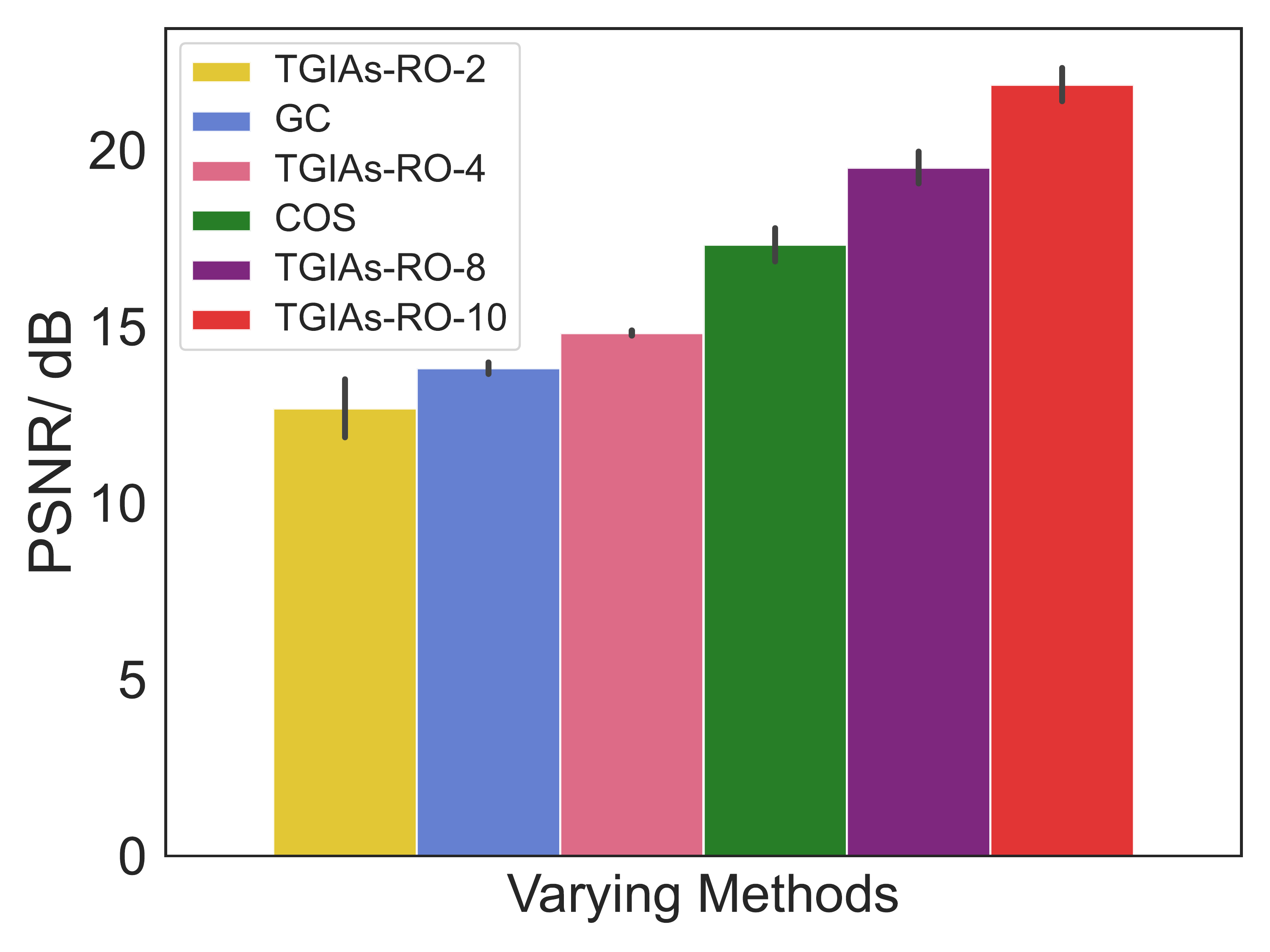}
		\subcaption{\small  PNSR Comparisons}
	  \end{subfigure}
	 \begin{subfigure}{0.24\textwidth}
		\centering
 
\includegraphics[keepaspectratio=true, width=120pt]{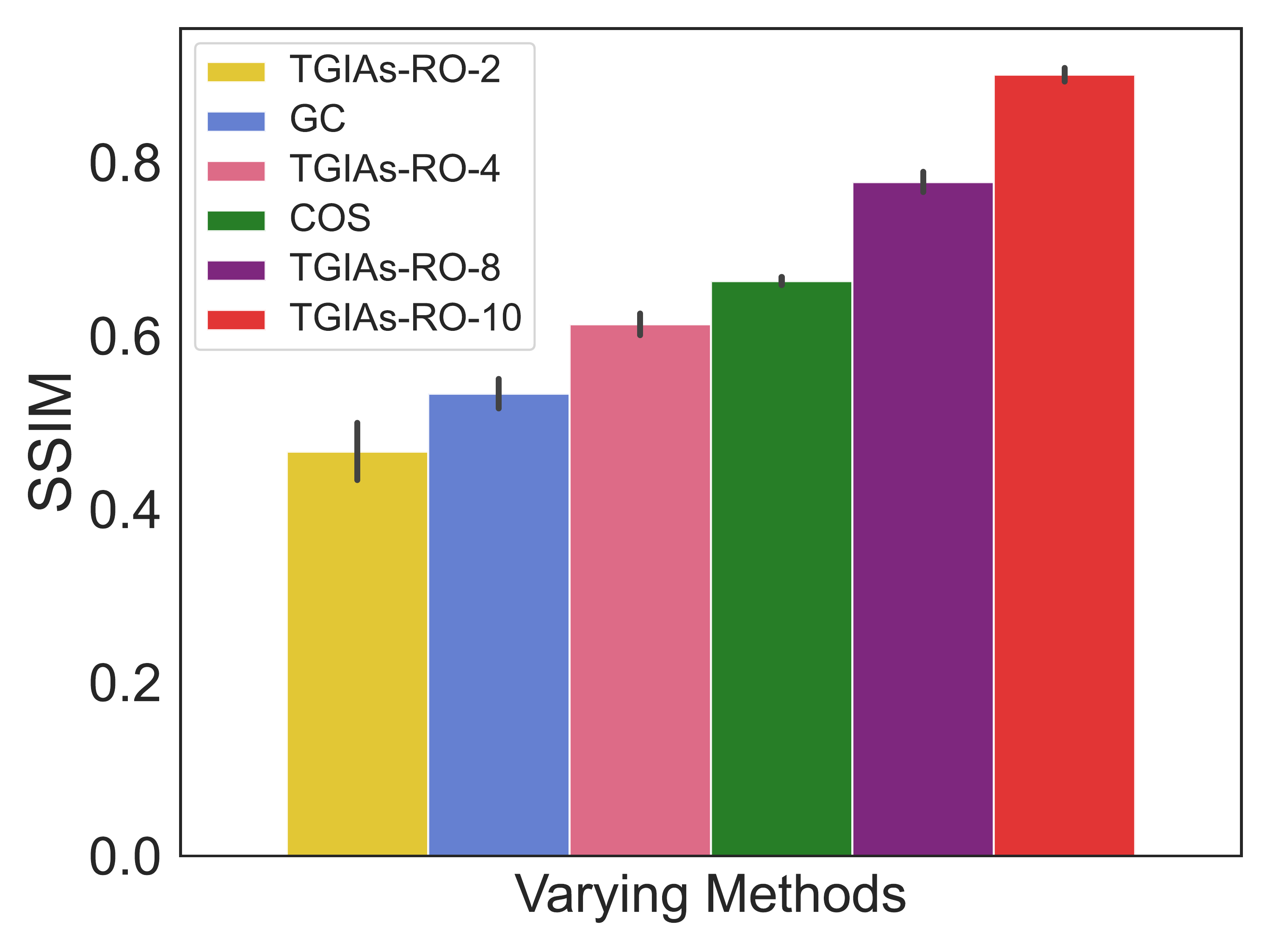}
		\subcaption{\scriptsize  SSIM Comparisons}
	  \end{subfigure}

 \caption{
 Comparisons of restored images with varying number of temporal gradients (e.g., TGIAs-RO-2 represent 2 gradients are used), the images are ImageNet in a  batch size 8 and the model architecture is ResNet-18. } 
  \label{fig:epochs}
\end{figure}

\subsection{Ablation Study} \label{sec:ablation}

\subsubsection{TGIAs-RO with Varying Number of Temporal Gradients}
Fig. \ref{fig:epochs} shows that more temporal gradients lead to higher PSNR and SSIM (e.g., the PSNR and SSIM of 10 temporal gradients are beyond 20 and 0.8 separately). Moreover, the visualization results in Fig. \ref{fig:epochs} also validates that sufficient gradient information leads to better image restoration.

\subsubsection{Effects of Robust Statistics}
\begin{figure}[htbp] 
    \centering

	\centering
	 \begin{subfigure}{0.24\textwidth}
		\centering
	
		\includegraphics[keepaspectratio=true, width=120pt]{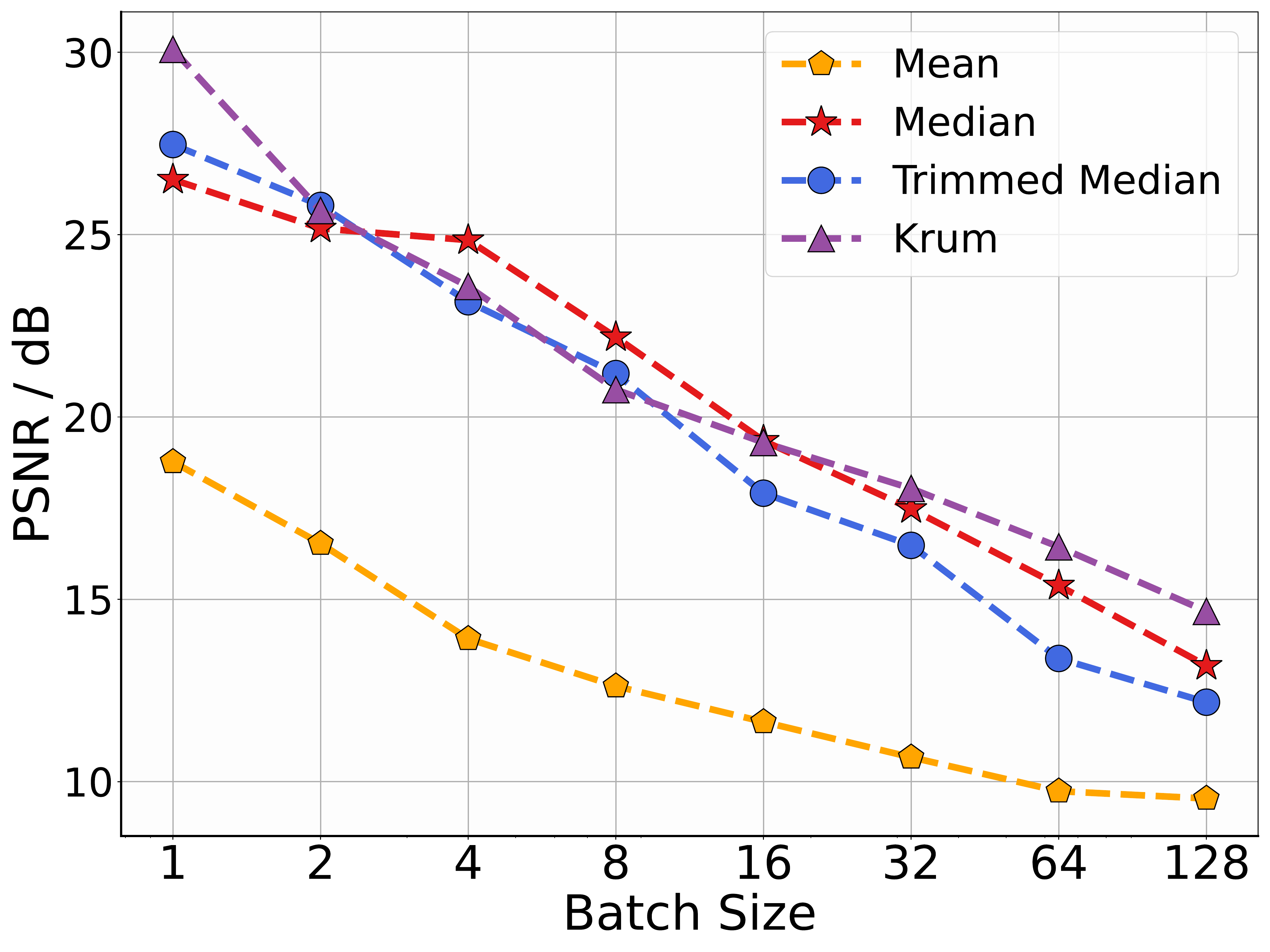}
		\subcaption{\small Comparisons of PNSR }
	  \end{subfigure}
	 \begin{subfigure}{0.24\textwidth}
		\centering
		\includegraphics[keepaspectratio=true, width=120pt]{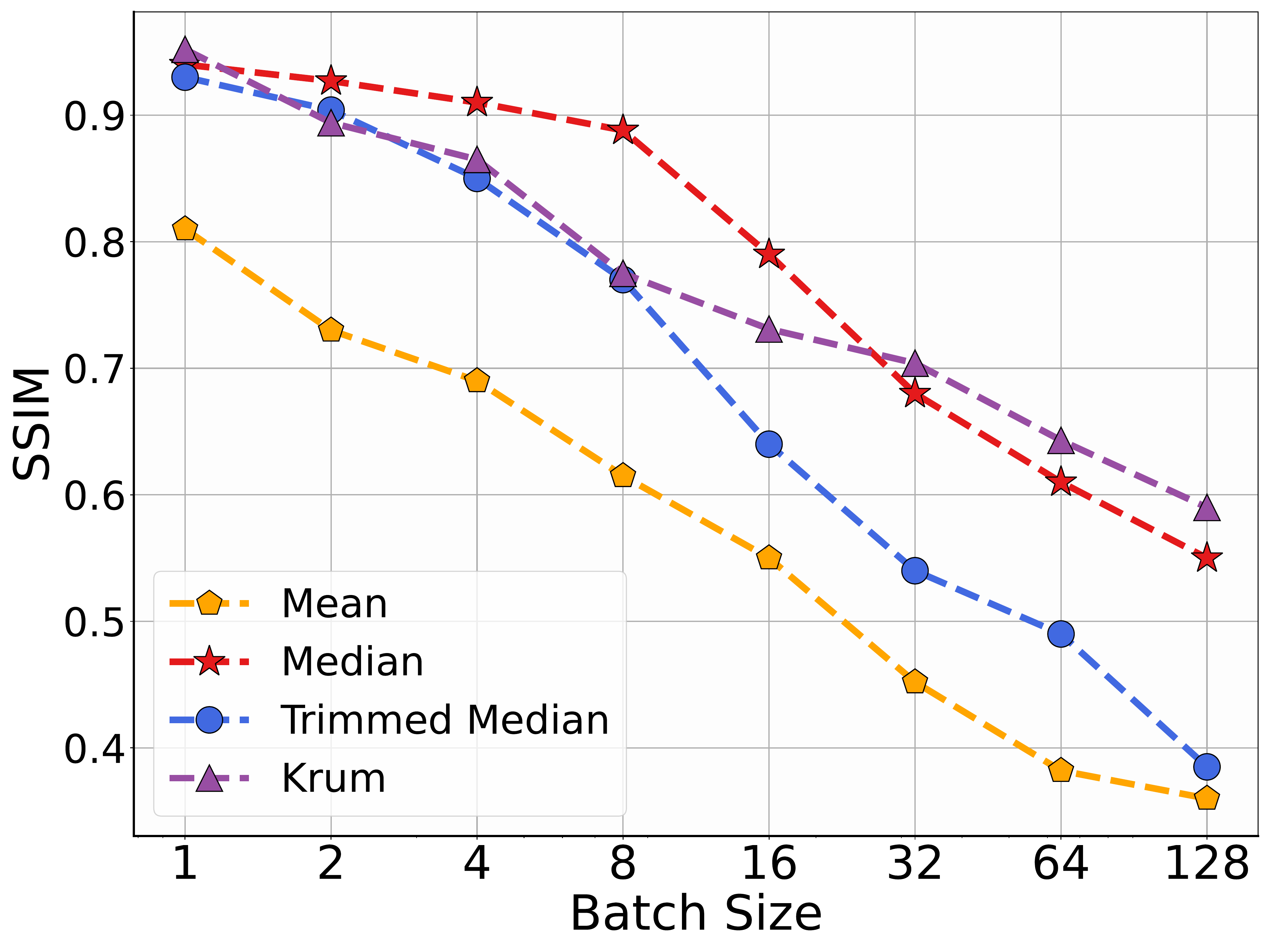}
		\subcaption{\small Comparisons of SSIM}
	  \end{subfigure}
	
	\caption{Comparisons of data recovery with four aggregation strategies: mean, Krum, coordinate-wised median and trimmed mean \cite{yin2018byzantine, blanchard2017machine}.}\label{fig: avg}
	
\end{figure}

To demonstrate the effects of robust statistics, we adopt four strategies to aggregate the intermediate data from different temporal gradients, i.e., mean, Krum, coordinate-wised median, trimmed mean \cite{yin2018byzantine, blanchard2017machine}. Fig. \ref{fig: avg} shows TGIAs-RO with median, Krum, and trimmed mean achieve higher-quality restoration than the mean strategy. The performance decades with mean strategy imply that some invalid gradients (see Sect. \ref{sec:failure}) destroy the data recovery. Thus it is necessary to leverage robust statistics to filter out the outliers in the global TGIAs-RO optimization. 

Tab. \ref{tab: robust} shows that robust aggregation under a single gradient is insufficient to recover images, although it improves the reconstruction quality without the robust aggregation. The reason is that some gradients are invalid (Sect. \ref{sec:failure} in the main text), so recovery is collapsed no matter what the input seed is. Moreover, multiple temporal gradients adopt more information and leverage the robust aggregation to restore the images clearly if more than half of the gradients are useful (Theorem \ref{thm:thm1} in the main text).

\begin{table}[H]
\caption{Robust Aggregation (RA) on single v.s. multiple gradients. }\label{tab: robust}
 \renewcommand{\arraystretch}{0.9} 
 \centering
\resizebox{0.493\textwidth}{!}{
\begin{tabular}{ccccccc}
\toprule 
\multirow{2}{*}{ Batch Size } &\multicolumn{2}{c}{\small Single gradient (DLG)} &\multicolumn{2}{c}{\small Single + RA} &\multicolumn{2}{c}{\small Multiple + RA (TGIAs-RO)} \\
\cmidrule(r){2-3} \cmidrule(r){4-5} \cmidrule(r){6-7}
&  PSNR/dB & SSIM     & PSNR/dB & SSIM  & PSNR/dB & SSIM   \\\midrule
4& 13.38 & 0.51 & 16.47 & 0.55 & \textbf{24.85} &\textbf{0.91} \\
8& 13.03 & 0.49 & 13.59 & 0.49 &\textbf{22.19} & \textbf{0.91}\\ 
    16& 12.86 & 0.36 & 13.10 & 0.38 &\textbf{19.35} & \textbf{0.79}\\ 
 \bottomrule
\end{tabular}
}
\end{table}

\begin{table*}[htbp]
      \caption{ Best case examples of restored texts of TGIAs-RO with different number of temporal gradients on text classification (the model adopted is TextCNN \cite{zhang2015sensitivity} with 1 convolution layer, and the texts are from Reuters R8 dataset with sentence length 32 in batch size of 8), the bold font indicates the error texts.}\label{tab: text_egs}
  \renewcommand{\arraystretch}{0.99}
     \centering
     \resizebox{0.99\textwidth}{!}{
      \begin{tabular}[l]{cp{7.5cm}p{7.5cm}}
        \toprule
        Methods & Original Text  & Restored Text   \\ \midrule
        
 \multirow{4}{*}{ 
 \tabincell{c}{DLG (TGIAs-RO\\with 1 temporal gradients)}}
 & champion products ch approves stock split champion products inc said its board of directors approved a two for one stock split of its common shares for shareholders of record as of april  & champion products ch approves stock split champion products inc said its board of directors approved a two for one stock split of its common shares for shareholders of record as of april  \\
 \cmidrule(r){2-3}
& investment firms cut cyclops cyl stake a group of affiliated new york investment firms said they lowered their stake in cyclops corp to shares or pct of the total outstanding common stock & investment systems cut cyclops cyl stake a systems of affiliated new has bought firms \textbf{of the stock of stake in cyclops corp exchange for} or pct its the \textbf{following} outstanding \textbf{common compensated} \\\cmidrule(r){2-3}
 & circuit systems csyi buys board maker circuit systems inc said it has bought all of the stock of ionic industries inc in exchange for shares of its common following the exchange there & circuit systems cut cyclops board \textbf{stake circuit group inc affiliated it york investment all said they lowered their stake} industries inc in to shares \textbf{shares of its the total the exchange compensate}  \\
\midrule
  \multirow{4}{*}{ 
  \tabincell{c}{TGIAs-RO with \\5 temporal gradients}}  
& champion products ch approves stock split champion products inc said its board of directors approved a two for one stock split of its common shares for shareholders of record as of april  &  champion products ch approves stock split champion products inc said its board of directors approved a two for one stock split of its common shares for shareholders of record as of april \\\cmidrule(r){2-3}
& investment firms cut cyclops cyl stake a group of affiliated new york investment firms said they lowered their stake in cyclops corp to shares or pct of the total outstanding common stock  & investment firms csyi cyclops cyl stake a group of affiliated new york investment firms said they lowered their stake in cyclops corp to shares or pct of the total outstanding \textbf{exchange there} \\
\cmidrule(r){2-3}
& circuit systems csyi buys board maker circuit systems inc said it has bought all of the stock of ionic industries inc in exchange for shares of its common following the exchange there & circuit systems cut buys board maker circuit systems inc said it has bought all of the stock of ionic industries inc in exchange for shares of its common following the \textbf{common stock}  \\
\midrule

\multirow{4}{*}{\tabincell{c}{TGIAs-RO with \\10 temporal gradients}} 
& champion products ch approves stock split champion products inc said its board of directors approved a two for one stock split of its common shares for shareholders of record as of april & champion exchange ch approves stock split champion products inc said its board of directors approved a two for one stock split of its common shares for shareholders of record as of april  \\ 
\cmidrule(r){2-3}
& investment firms cut cyclops cyl stake a group of affiliated new york investment firms said they lowered their stake in cyclops corp to shares or pct of the total outstanding common stock & investment firms cut cyclops cyl stake a system of affiliated new york investment firms said they lowered their stake in cyclops corp to shares or pct of the total outstanding common \textbf{record} \\
\cmidrule(r){2-3}
&circuit systems csyi buys board maker circuit systems inc said it has bought all of the stock of ionic industries inc in exchange for shares of its common following the exchange there &china systems csyi broad buys maker circuit systems inc said it has bought all of the stock of ionic industries inc in exchange for shares of its common following the exchange there\\
      \bottomrule
      \end{tabular}}

\end{table*}

\subsubsection{TGIAs-RO with More Prior Knowledege}
In TGIAs-RO, each temporal gradients conduct a GIAs optimization $\calL_{grad}$, while the optimization can be enhanced with additional prior knowledge. 
\textbf{Batch Normalization Statistics:}
\textit{BN regularizer} \cite{yin2021see} adopted  statistics from the batch normalization layers in network to design an auxiliary regularization term $\calL_{BN}$ for better image restoration. 
\textbf{Total Variation Regulaization Term:} 
\textit{Total variation} is a measure of the complexity of an image with respect to its spatial variation. Total variation can be adopted as prior information to help image reconstruction in the form of a regularization term $\calL_{TV}$. 

The results in Tab. \ref{tab: prior} indicate that total variation regularization term $\calL_{TV}$ improves PSNR by more than 2 dB, while batch normalization regularization term $\calL_{TV}$ slightly improves PSNR within less than 1dB. 
\begin{table}[H]
\caption{Comparisons of reconstruction performance  with different prior knowledge, the model adopted is AlexNet and dataset is CIFAR10 (batch size = 16, image size = 32$\times$32).}\label{tab: prior}
 \renewcommand{\arraystretch}{1}
 \centering
\resizebox{0.48\textwidth}{!}{
\begin{tabular}{cccc}
\toprule
Regularization Terms Adopted & MSE &  PSNR/dB & SSIM     \\ \midrule
$\calL_{grad}$ & 6$e^{-3}$ & 22.32  &  0.86 \\
$\calL_{grad} + \calL_{TV}$  & 3$e^{-3}$  &  24.92  & 0.91  \\ 
$\calL_{grad} + \calL_{BN}$  &4$e^{-3}$ & 23.02 & 0.87 \\
$\calL_{grad} + \calL_{TV} + \calL_{BN}$  &3$e^{-3}$  & 25.47 & 0.92 \\
 \bottomrule
\end{tabular}
}

\end{table}




\subsubsection{TGIAs-GO on Text Classification Tasks}
We also conduct experiments on text-processing tasks. Specifically, we choose a text classification task with a TextCNN with 3 convolution layers \cite{zhang2015sensitivity}. The R8 dataset (full term version) is a subset of the Reuters 21578 datasets, which has 8 categories divided into 5,485 training and 2,189 test documents.  

Different from image restoration tasks where the inputs are continuous values, language models need to preprocess discrete words into embedding (continuous values). We apply TGIAs-RO on embedding space and minimize the gradient distance between dummy embedding and ground truth ones. After optimizing the dummy embedding, we derive original words by finding the closest index entries in the embedding matrix reversely.

\begin{table}[H]
\caption{Recover rate of TGIAs-RO on text classification tasks, the model architecture adopted is TextCNN and dataset is R8 dataset from Reuters 21587 datasets (sentence length = 30).}\label{tab: NLP}

 \renewcommand{\arraystretch}{1.05}
 \centering
\resizebox{0.48\textwidth}{!}{
\begin{tabular}{cccc}
\toprule 
\multirow{2}{*}{ Iteration  Number } & \multirow{2}{*}{ Batch Size } &\multicolumn{2}{c}{Convolution Block Number} 
\\ \cmidrule(r){3-4} 
& &  1 & 3      \\ \midrule
\multirow{3}{*}{ $T=1$ (DLG) }
&2  & 100\%  &  93.75\%  \\
&4    & 93.75\%  & 70.3\% \\ 
&8    & 78.12\%   & 31.25\%   \\ \hline
\multirow{3}{*}{ $T=5$ }&2   & 100\%    & 98.4\% \\
&4   & 97.65\% & 79.7\% \\
&8   & 87.11\% &  41.01\%  \\ \midrule
\multirow{3}{*}{ $T=10$ }&2    & 100\%    & 100\%   \\
& 4  & 100\%  & 87.5\%  \\
&8 & 91.8\%  &  60.94\%  \\
 \bottomrule
\end{tabular}
}

\end{table}
We reconstruct the text in Reuters R8 with two different TextCNN model architectures. One with 1 convolution layer and the other one with 3 convolution layers, each convolution layer is followed by a sigmoid function and a max pooling layer. We adopt \textit{recover rate} \cite{deng2021tag} to measure the text restoration performance, which is defined as the maximum percentage of tokens in ground truth recovered by TGIAs-RO. Results in Tab. \ref{tab: NLP} indicate that TGIAs-RO with 10 temporal gradients reconstruct text embedding better than those with fewer gradients. Moreover, texts in a larger batch size are more difficult to reconstruct, and more complex model architectures increase the failures in GIAs as Tab. \ref{tab: NLP} shows.




\section{Discussion}
In this work, we evaluate the efficacy of our proposed method, TGIAs-RO, in realistic scenarios commonly encountered in federated learning, including client sampling, differential privacy, and gradient sparsification techniques.
Recent advancements have introduced various methods aimed at preserving data privacy in federated learning. Examples include MixUp \cite{zhang2018mixup}, InstaHide \cite{huang2020instahide}, FedPass \cite{gu2023fedpass}, and representation encoding \cite{sun2021soteria}. In future research, we plan to investigate the applicability of TGIAs-RO across a broader spectrum of privacy-preserving strategies. 

Additionally, exploring adaptive privacy-preserving methods to address privacy leakage concerns stemming from multiple iterations is a potential avenue for further investigation.

\section{Conclusion}
In this paper, we demonstrate the risks of data leakage in  horizontal federated learning. 
We point out that gradient inversion attacks adopting single-temporal gradients fail due to the insufficient gradient knowledge, complex model architectures and invalid gradients. And we introduce a gradient inversion attack framework named 
TGIAs-RO leveraging multi-temporal gradients and robust statistics to reconstruct private data with high quality and theoretical guarantee. Extensive experimental and theoretical results demonstrate that our proposed method recovers data successfully even for the large data size, batch size, and complex models.


\section*{Acknowledgement}
This work has been partially supported by the National Key R$\&$D Program of China No. 2020YFB1806700 and the NSFC Grant 61932014. 




\bibliography{related}{}
\bibliographystyle{IEEEtran}
\begin{IEEEbiography}[{\includegraphics[width=0.9in,keepaspectratio]{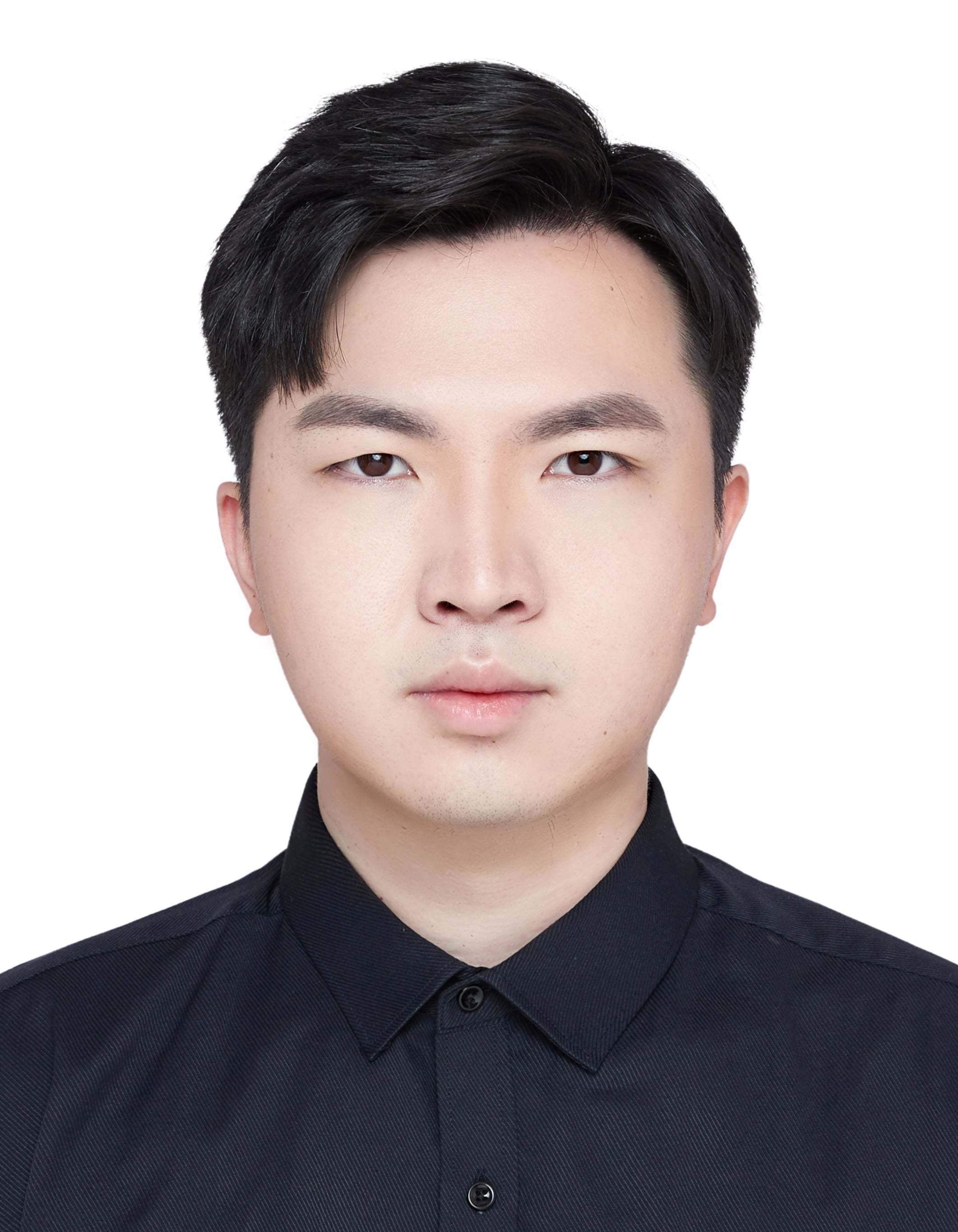}}]{Bowen Li}
received his B.S. degree in Automation from Xi'an Jiaotong University, China in 2019. He is currently pursuing the Ph.D. degree at the Department of Computer  Science and Engineering, Shanghai Jiao Tong University, China. He worked as a research intern at WeBank AI Group
, WeBank, China in 2021. 
His research interests include federated  learning, data privacy and machine learning security. He has published a first-authored paper in IEEE TPAMI.
\end{IEEEbiography}

\begin{IEEEbiography}[{\includegraphics[width=0.9in,keepaspectratio]{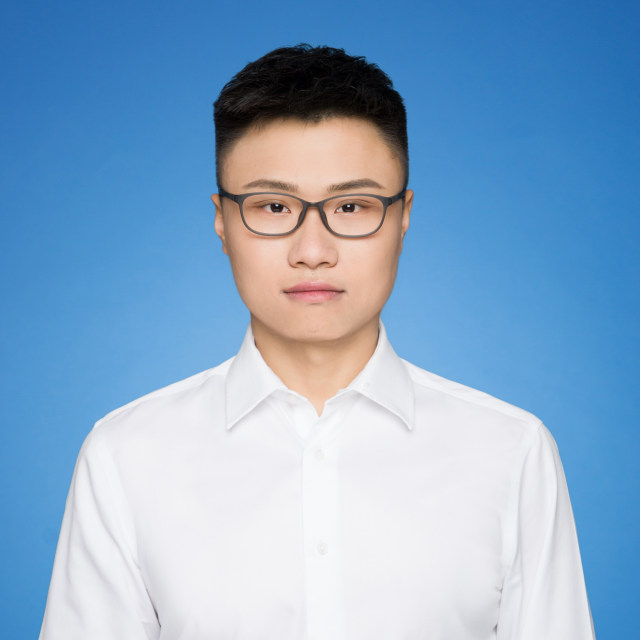}}]{Hanlin Gu}
 received his B.S. degree in Mathematics 
from University of Science and Technology of China in 2017.
He obtained his Ph.D. degree from 
the Department of  Mathematics,
Hong Kong University of Science and Technology.
He is working as a researcher at WeBank AI
Group , WeBank, China in 2021. His research
interests include federated learning, privacy-preserving methodology.
\end{IEEEbiography}

\begin{IEEEbiography}[{\includegraphics[width=1in,height=1.25in,clip,keepaspectratio]{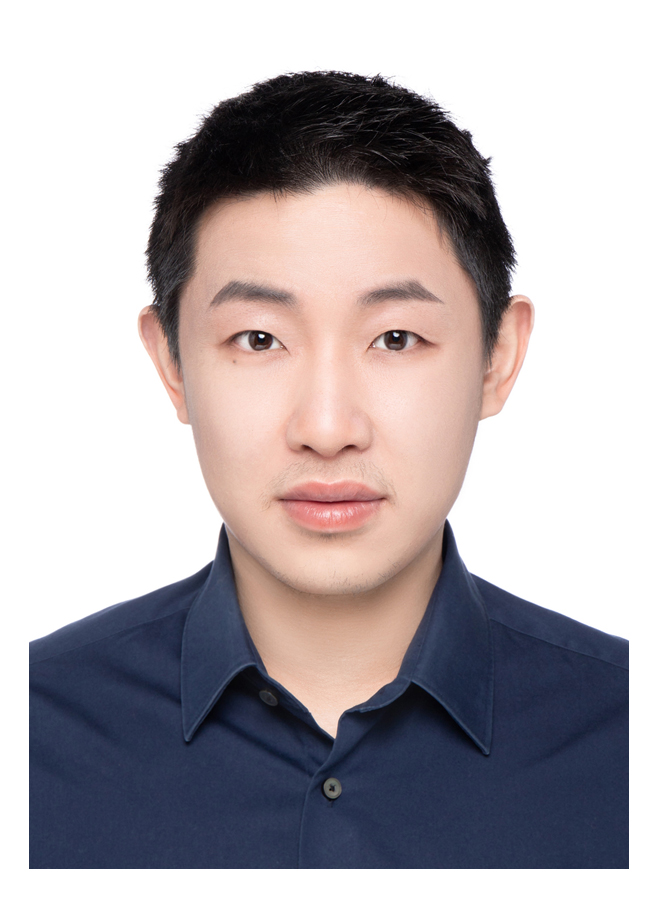}}]{Ruoxin Chen} (S’21-) is currently pursuing the Ph.D. degree in Electronic Information and Electrical Engineering, Shanghai Jiaotong University. He received the B.S. degree in Mathematics and Statistics from Wuhan University, Wuhan, China, in 2014. His main research interest is certified defenses against adversarial examples and data poisoning. He has published first-authored papers in ICML, AAAI.
\end{IEEEbiography}

\begin{IEEEbiography}[{\includegraphics[width=0.9in,clip,keepaspectratio]{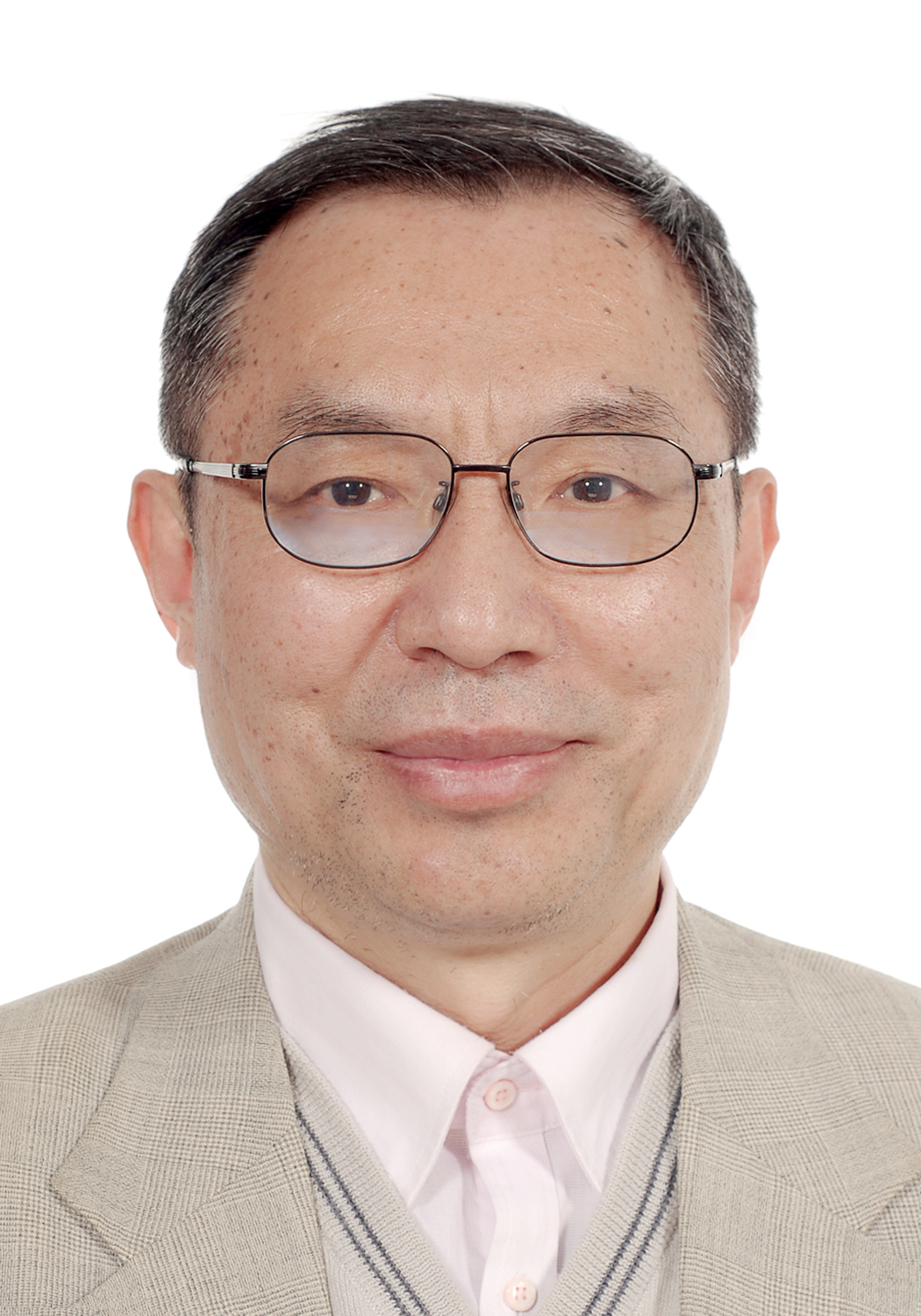}}]{Jie Li} received the B.E. degree in computer science from Zhejiang
University, Hangzhou, China, in 1982, the M.E. degree in electronic engineering and communication systems from China Academy of Posts and
Telecommunications, Beijing, China, in 1985, and the Dr.Eng. degree from the University of ElectroCommunications, Tokyo, Japan, in 1993.
He is with the Department of Computer Science and Engineering, Shanghai Jiao Tong University, Shanghai, China, where he is a Professor. He was a Full Professor with the Department of Computer Science, University of Tsukuba, Tsukuba, Japan. He was a Visiting Professor with Yale University, New Haven, CT, USA; Inria Sophia Antipolis, Biot, France; and Inria Grenoble Rhône-Alpes, Montbonnot-Saint-Martin, France. His current research interests are in big data, IoT, blockchain, edge computing, OS, and modeling and performance evaluation of information systems. Prof. Li is the Co-Chair of the IEEE Technical Community on Big Data and the IEEE Big Data Community, and the Founding Chair of the IEEE ComSoc Technical Committee on Big Data. He serves as an associate editor for many IEEE journals and transactions. He has also served on the program committees for several international conferences.
\end{IEEEbiography}

\begin{IEEEbiography}[{\includegraphics[width=0.9in,keepaspectratio]{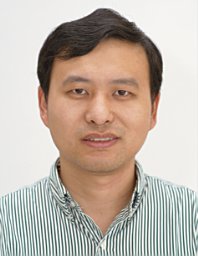}}]{Chentao Wu} 
received the B.S.,
M.E., and Ph.D. degrees in computer science from
the Huazhong University of Science and Technology, Wuhan, China, in 2004, 2006, and 2010,
respectively, and the Ph.D. degree in electrical and
computer engineering from Virginia Commonwealth
University, Richmond, in 2012. He is currently a
Professor with the Department of Computer Science and Engineering, Shanghai Jiao Tong University, Shanghai, China. He has published more than
50 papers in prestigious international conferences
and journals, such as the IEEE TRANSACTIONS ON COMPUTERS, IEEE
TRANSACTIONS ON PARALLEL AND DISTRIBUTED SYSTEMS, HPCA, DSN,
and IPDPS. His research interests include computer architecture and data
storage systems. He is a member of the China Computer Federation.

\end{IEEEbiography}

\begin{IEEEbiography}[{\includegraphics[width=1in,height=1.25in,clip,keepaspectratio]{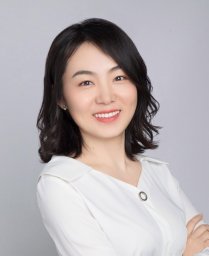}}]{Na Ruan} received the BS degree in information engineering, the MS degree in communication and information system from the China University of Mining and Technology, in 2007 and 2009 respectively, and the DE degree from the Faculty of Engineering, Kyushu University, Japan, in 2012. She is currently an Assistant Professor with the Department of Computer Science and Engineering, Shanghai Jiao Tong University, China. Her research interests include big data, security and privacy, networks, and block chain
\end{IEEEbiography}

\begin{IEEEbiography}[{\includegraphics[width=1in,height=1.25in,clip,keepaspectratio]{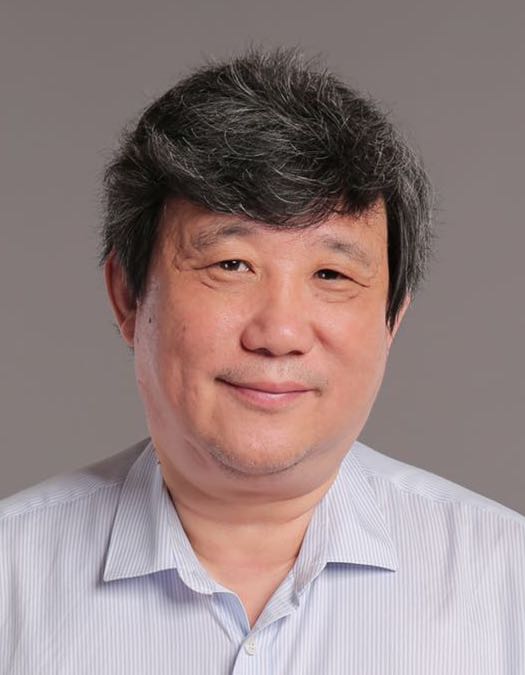}}]{Xueming Si} is the chief scientist of the Blockchain Research Center at Shanghai Jiao Tong University, the director of the Blockchain Special Committee of the China Computer Society, the deputy director of Shanghai Key Laboratory of Data Science (Fudan University), the director of the Blockchain Expert Committee of China Computer Society, and the leader of basic technology of blockchain and application innovation alliance in China. He is specialized in cryptography, data science, computer architecture, network and information system security, and blockchain. He has won the first prize of national science and technology progress for three times.
\end{IEEEbiography}
\begin{IEEEbiography}[{\includegraphics[width=0.9in,keepaspectratio]{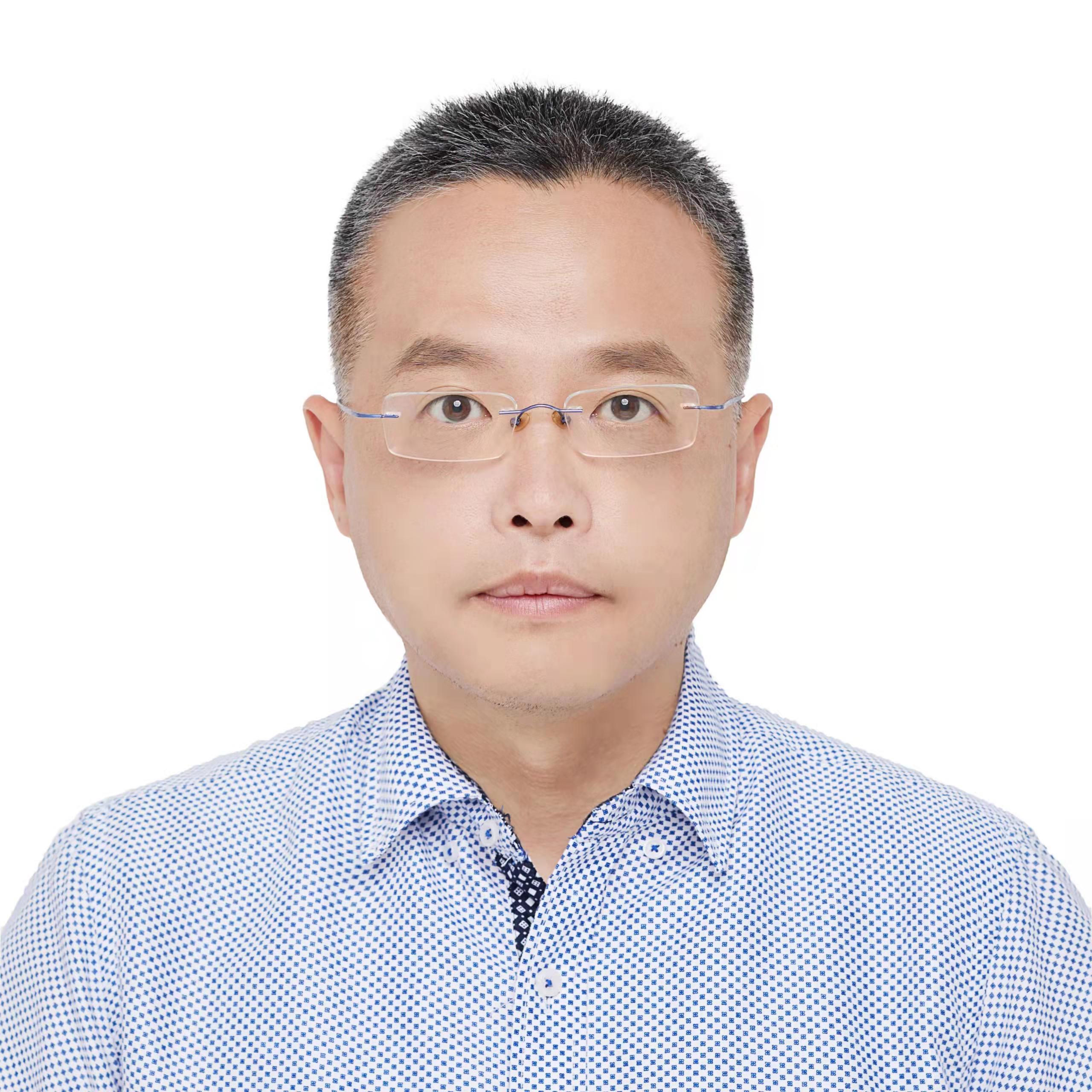}}]{Lixin Fan}

is the Chief Scientist of Artificial Intelligence at WeBank. His research fields include machine learning and deep 
learning, computer vision and pattern recognition,
image and video processing, 3D big data processing and mobile computing. He is the author of more than 70 international journals 
and conference articles. He has worked at Nokia Research Center and Xerox Research Center Europe. He has participated in NIPS/NeurIPS, ICML, CVPR,
ICCV, ECCV, IJCAI and other top artificial intelligence conferences for a
long time, served as area chair of ICPR. He is also the inventor of more than one hundred patents filed in
the United States, Europe and China, and the chairman of the IEEE P2894
Explainable Artificial Intelligence (XAI) Standard Working Group.
\end{IEEEbiography}

\clearpage
\appendices
\setcounter{equation}{0}
\setcounter{thm}{0}
\setcounter{prop}{0}
\setcounter{table}{0}
\setcounter{figure}{0}
\section{Theoretical Proof}
In this section, we provide the theoretical proof and extension for our Proposition 1 and theorem 1. \footnote{We exchange the position of \textbf{Theoretical Proof} and \textbf{Implementation Details}}

\begin{table}[!ht] 
  \renewcommand{\arraystretch}{1.05}
  \centering
  \setlength{\belowcaptionskip}{15pt}
  \caption{Table of Notations}
  \label{table: notation}
    \begin{tabular}{c|p{5.5cm}}
    \toprule
    Notation & Meaning\cr
    \midrule\
    $\calD = (\bx, \by)$ & Original data including input images $\bx$ and label $\by$ \cr \hline
    $b$ & Batch size of  $\calD$ \cr \hline
    $n$ & Size (Dimension) of $\calD$ \cr \hline
    $\hat{\calD}, \hat{\bx}, \hat{\by}$ &  Restored data, input images and label\cr \hline
    $\bw, \calL$ & Model weights and loss of main task\cr \hline
     $\nabla_{\bw} \calL(\bw, \calD)$ & Model Gradients for loss $\calL$ on data $\calD$ w.r.t. $\bw$ \cr \hline

 $p$ & Dimension of Model gradients\cr \hline
 $T$ & Number of temporal gradients (information) \cr \hline
  $()_t$ & Items for $t_{th}$ temporal gradients\cr \hline
  $()^s$ & Items for $s_{th}$ updating\cr \hline
 $f_t$ & Empirical loss for GIAs in $t_{th}$ temporal information as Eq. \eqref{each in multi}\cr \hline
  $\alpha_t$ & The weights of $t_{th}$ local empirical loss function\cr \hline
  $f$ & Total collaborative loss for GIAs for $T$ temporal information as Eq. \eqref{eq: multi_temporal} \cr \hline
  $\nabla f_t(\hat{\bx}_t)$ & Gradients of function $f_t$ w.r.t. $\hat{\bx}_t$\cr \hline
   $R_g$ & Number of steps for global aggregation\cr \hline
 $R_l$ & Number of steps for local optimization\cr \hline
 $m$ & Number of collapsed items\cr \hline
 $\|\cdot \| $ & $\ell_2$ norm \cr 
    \bottomrule
    \end{tabular}
\end{table}

\subsection{Proof for Proposition 1}
Assume that adversaries use the objective loss function as Eq. \eqref{eq:gradient-inversion-single} \cite{zhu2019deep}, aimed at recovering the original batch data $\calD=\{(x_i, y_i)_{i=1}^b \}$ with $b$ batch size by comparing the gradients generated by restored batch data ($\nabla_{\bw}\calL(\bw, \hat{\calD})$ and the leaked gradients of original batch data ($\nabla_{\bw}\calL(\bw, \calD)$)
\begin{equation} \label{eq:gradient-inversion-single-app}
\begin{split}
    \hat{\calD} = \argmin_{\hat{\calD}}||\nabla_{\bw}\calL(\bw, \hat{\calD}) - \nabla_{\bw}\calL(\bw, \calD)||^2, 
\end{split}
\end{equation}

\begin{prop} \label{prop1-app}
When the model on the main task is convergent for a series of batch data $\calD_1, \calD_2, \cdots$, i.e., $\nabla_{\bw}\calL(\bw, \calD_1) = \nabla_{\bw}\calL(\bw, \calD_2) = \cdots= 0$, then recovering $\calD_1$ using Eq. \eqref{eq:gradient-inversion-single-app} is impossible almost surely.
\end{prop}

\begin{proof}
Define $\calD_k = \{x_{i,k},y_{i,k} \}_{i=1}^b$ for $k=1,2 \cdots$. Since $\nabla_{\bw}\calL(\bw, \calD_1) = \nabla_{\bw}\calL(\bw, \calD_2) = \cdots= 0$, Eq. \eqref{eq:gradient-inversion-single-app} can be expressed as:
\begin{equation}
        \hat{\calD} = \argmin_{\hat{\calD}}||\nabla_{\bw}\calL(\bw, \hat{\calD})||^2, 
\end{equation}
thus all $\calD_k$ are the solutions of optimization problem \eqref{eq:gradient-inversion-single-app}. Moreover, if the batch size $b$ is known for adversaries, we have for any $k,s$
\begin{equation}
\begin{split}
       & \nabla_{\bw}\calL(\bw, \calD_k \cup \calD_s)  \\&= \frac{1}{2b}\sum_i^{b}(\frac{\partial\calL(\bw, x_{i,k}, y_{i,k} )}{\partial \bw} + \frac{\partial\calL(\bw, x_{i,s}, y_{i,s} )}{\partial \bw})\\
        &= \frac{\partial\calL(\bw, \calD_k)}{2\partial \bw}+ \frac{\partial\calL(\bw, \calD_s)}{2\partial \bw} \\
        &= \frac{1}{2}(\nabla_{\bw}\calL(\bw, \calD_k)+ \nabla_{\bw}\calL(\bw, \calD_s)) \\
        &= 0.
\end{split}
\end{equation}
Therefore, all $\calD_k \cup \calD_s$ are the solutions of optimization problem \eqref{eq:gradient-inversion-single-app}. Furthermore, for any sequence $\{k_1, k_2,... \}$, we have $\calD_{k_1} \cup \calD_{k_2} \cup \cdots $ are the solutions of optimization problem. Therefore, recovering $\calD_1$ using Eq. \eqref{eq:gradient-inversion-single-app} is impossible almost surely.
\end{proof}

\subsection{Proof of Theorem 1}
Next, suppose the last $m$ temporal optimizations  $\{f_t()\}_{t=T-m+1}^{T}$ collapse in TGIAs-RO. We firstly analyze the property of robust statistics such as median \cite{yin2018byzantine}. Denote $\nabla f^s = \frac{1}{T-m}\sum_{t=1}^{T-m}\nabla f_t(\hat{\bx}^s)$ and $\kappa$ is the upper bound of $\|\nabla f_t( \hat{\bx}^s) - \nabla f^s\|_2$ \footnote{There is one mistake in main text about the definition of $\kappa$, the right definition of $\kappa$ is the upper bound of $\|\nabla f_t( \hat{\bx}^s) - \nabla f^s\|_2$.}, where $\nabla f_t(\hat{\bx}^s_t)$ is the gradients of the function $f_t$ w.r.t. $\hat{\bx}^s_t$.\\
\textbf{Claim 1:} $\| \text{median}( \{ \nabla f_t(\hat{\bx}^s)\}_{t=1}^T)-\nabla f^s \|_2 \leq \calO(\kappa)$.
\begin{proof}
Suppose the dimension of $\nabla f_t^s$ to be $n$ (data size). Since $m < \frac{T}{2}$, we have
\begin{equation}
\begin{split}
        & \| \text{median}( \{ \nabla f_t(\hat{\bx}^s_t)\}_{t=1}^T)-\nabla f^s] \|_2 \\
        & = \sum_{i=1}^n \sqrt{[\text{median}( \{ \nabla f_t(\hat{\bx}^s)\}_{t=1}^T)-\nabla f^s]^2_{(i)}} \\
        &\leq \sum_{i=1}^n \sqrt{[\text{max}( \{ \nabla f_t(\hat{\bx}^s)\}_{t=1}^T)-\nabla f^s]^2_{(i)}} \\
        &\leq  \sum_{i=1}^n\sqrt{\sum_{t=1}^{T-m} [ f_t(\hat{\bx}^s)-\nabla f^s]^2_{(i)}} \\
        & \leq \sqrt{n} \sqrt{\sum_{i=1}^n\sum_{t=1}^{T-m} [ f_t(\hat{\bx}^s)-\nabla f^s]^2_{(i)}} \\
       &  = \sqrt{(T-m)n}\kappa \triangleq \calO(\kappa), 
\end{split}
\end{equation}
\noindent where the last inequality is due to Cauchy–Schwarz inequality.
\end{proof}
\begin{rmk}
\textbf{Explain for $\kappa$.} $\kappa$ represents the calculated gradients $\nabla f_t( \hat{\bx}^s)$ of normal temporal information are closed to each other, it is reasonable because they have the common global solution. On the other hand, the existence of $\kappa$ is also used in collaborative learning \cite{li2019communication}. Specifically, when the temporal information is homogeneous, $\kappa$ tends to zero \cite{li2019communication}.
\end{rmk}
\begin{rmk}
It is noted that the error bound using Krum \cite{blanchard2017machine}, median and trimmed mean \cite{yin2018byzantine} in \textbf{Claim 1} relies on the dimension of parameters $\sqrt{n}$, it is inaccurate when $n$ is large. \cite{guerraoui2018hidden} introduced a new aggregation method, whose error bound reduces to $\frac{1}{\sqrt{n}}$. We leave this novel robust aggregation method in the future work.
\end{rmk}
Furthermore, we provide the proof of Theorem 1 according to \textbf{Claim 1}, the outline of proof follows \cite{yin2018byzantine}.
\begin{thm} \label{thm:thm1-app}
If $\{f_t\}_{t=1}^{T-m}$ is $\mu$ strong convex and L-smooth, the number of collapsed restored data $m$ is less than $\frac{T}{2}$. Choose step-size $\eta = \frac{1}{L}$
and Algo. \ref{algo:TGIAs-RO} obtains the sequence $\{\hat{\bx}^s=\text{RobustAgg/Median}(\hat{\bx}^s_1,\cdots, \hat{\bx}^s_T) : s\in[0:R_g]\}$
satisfying the following convergence guarantees:
\begin{equation}
    \left\|\hat{\bx}^{s} - \bx^*\right\|_2 \leq (1-\frac{\mu}{\mu+L})^s\left\|\hat{\bx}^{0} - \bx^*\right\|_2 + \frac{2 \Gamma}{\mu}.
\end{equation}
Moreover, we have
\begin{equation}
    \lim_{s \to \infty} \left\|\hat{\bx}^{s} - \bx^*\right\|_2  \leq  \frac{2 \Gamma}{\mu},
\end{equation}
where $\bx^*$ is the global optimal and $\Gamma =\calO(\kappa)$.
\end{thm}
\begin{proof}
According to Algo. \ref{algo:TGIAs-RO} for $R_l=1$, we obtain
\begin{equation*}
\begin{split}
       \hat{\bx}_t^s &= \hat{\bx}^{s-1} - \eta \text{Median}(\{ \nabla f_t(\hat{\bx}^{s-1})\}_{t=1}^T )\\
       & = (\hat{\bx}^{s-1} -\eta \nabla f^{s-1}) \\
       & \quad + \eta ( \nabla f^{s-1} - \text{Median}(\{ \nabla f_t(\hat{\bx}^{s-1}_t)\}_{t=1}^T)).
\end{split}
\end{equation*}
Thus, 
\begin{equation}
\begin{split}
        & \| \hat{\bx}^s - \bx^* \|_2 \\
        & = \| (\hat{\bx}^{s-1} -\eta \nabla f^{s-1} -\bx^*) \\
        & \quad + \eta ( \nabla f^{s-1} - \text{Median}(\{ \nabla f_t(\hat{\bx}^{s-1})\}_{t=1}^T )) \|_2 \\
        & \leq \underbrace{\|\hat{\bx}^{s-1} -\eta \nabla f^{s-1} -\bx^*\|_2}_{=:U} \\
        & \quad + \underbrace{\| \eta (\nabla f^{s-1} - \text{Median}(\{ \nabla f_t(\hat{\bx}^{s-1})\}_{t=1}^T )) \|_2}_{=:V}.
\end{split}
\end{equation}
Firstly, 
\begin{equation*}
\begin{split}
        U^2 = & \|\hat{\bx}^{s-1} - \bx^* \|_2^2 - 2 \eta<\hat{\bx}^{s-1} - \bx^* ,  \nabla f^{s-1}> \\
        & + \eta^2 \|\nabla f^{s-1} \|_2^2.
\end{split}
\end{equation*}
Since $f$ is $L$-smooth and $\mu$ strong convex, according to the \cite{bubeck2015convex}, we obtain:
\begin{equation}
\begin{split}
    & < \hat{\bx}^{s-1} - \bx^*, \nabla f^{s-1}>  \\
    & \geq \frac{1}{\mu+ L}\|\nabla f^{s-1} \|_2^2 + \frac{\mu L}{\mu+L}||\hat{\bx}^{s-1}-\bx^*||^2.
\end{split}
\end{equation}
Let $\eta = \frac{1}{L}$ and assume $\mu \leq L$, then we get
\begin{equation}
\begin{split}
    U^2 &\leq (1- \frac{2 \mu}{\mu+L})\| \hat{\bx}^{s-1} - \bx^* \|_2^2 - \frac{2}{L(\mu+L)} \|\nabla f^{s-1} \|_2^2 \\
    & + \frac{1}{L^2}\|\nabla f^{s-1} \|_2^2 \\
    & \leq (1- \frac{2 \mu}{\mu+L})\| \hat{\bx}^{s-1} - \bx^* \|_2^2. 
\end{split}
\end{equation}
Since $\sqrt{1-x} \leq 1-\frac{x}{2}$, we get
\begin{equation} \label{eq:U-app}
    U \leq (1- \frac{ \mu}{\mu+L})\| \hat{\bx}^{s-1} - \bx^* \|_2. 
\end{equation}
Furthermore, according to \textbf{Claim 1}, we have
\begin{equation}\label{eq:V-app}
    V \leq \calO(\kappa). 
\end{equation}
Combining Eq. \eqref{eq:U-app} and \eqref{eq:V-app}, we get
\begin{equation} \label{eq:one-iterate-app}
    \| \hat{\bx}^s - \bx^* \|_2 \leq  (1- \frac{ \mu}{\mu+L})\| \hat{\bx}^{s-1} - \bx^* \|_2 + \frac{1}{L}\calO(\kappa), 
\end{equation}
then by iterating Eq. \eqref{eq:one-iterate-app}, we have
\begin{equation}
    \left\|\hat{\bx}^{s} - \bx^*\right\|_2 \leq (1-\frac{\mu}{\mu+L})^s\left\|\hat{\bx}^{0} - \bx^*\right\|_2 + \frac{2 \Gamma}{\mu},
\end{equation}
where $\Gamma = \calO(\kappa)$, which completes the proof.
\end{proof}
\begin{rmk}
We consider the case that local updating step $R_l=1$ in Theorem \ref{thm:thm1}, the general case the proof is similar to \cite{data2021byzantine}.
\end{rmk}
\begin{rmk}
In optimization of GIAs, second order optimization method LBFGS \cite{zhu1997algorithm} is often used. The proof for Algo. \ref{algo:TGIAs-RO} under LBFGS is similar to \cite{liu2021resource}, which needs the stronger assumption that hessian matrix $H$ of $f_t$ satisfies $H \leq \lambda I$.
\end{rmk}

\subsection{Extension to Non-convex condition for Theorem 1}
In this part, we analyze the convergence rate for the Non-convex condition of $f_t$.
\begin{lem} \cite{boyd2004convex} \label{lem:l-smmoth}
If $f$ is L-smooth, then:
\begin{itemize}
    \item For all $\bx^*$ (optimal solution w.r.t $f$), we have
    \begin{equation} 
    \frac{1}{2L}||\nabla f||^2 \leq f(\bx) -f(\bx^*) \leq \frac{L}{2}||\bx - \bx^*||^2. 
    \end{equation}
    \item For any $x,y$, we have
    \begin{equation}
        |f(y) - f(x) - \nabla f(x)^T(y - x))| \leq \frac{L}{2}\|x-y\|_2^2. 
    \end{equation}
\end{itemize}
\end{lem}
\begin{thm} \label{thm:thm2}
If $\{f_t\}_{t=1}^{T-m}$ is Non-convex and L-smooth, the number of collapsed restored data $m$ is less than $\frac{T}{2}$. Choose step-size $\eta = \frac{1}{L}$
and Algo. \ref{algo:TGIAs-RO} obtains the sequence $\{\hat{\bx}^s=\text{RobustAgg/Median}(\hat{\bx}^s_1,\cdots, \hat{\bx}^s_T) : s\in[0:R_g]\}$, we have
\begin{equation}
   \min_{s=1,\cdots, R_g} \|\sum_{t=1}^{T-m}\nabla f_t(\bx^s)\|_2 \leq  \frac{\sqrt{2}}{R_g}(\sqrt{f(\hat{\bx}^{0}) - f(\bx^*)}  + \calO(\kappa),
\end{equation}
where $\Gamma = \calO(\kappa)$. 
\end{thm}
\begin{proof}
For collaborative optimization problem (Eq. \eqref{eq: multi_temporal}), we define 
\begin{equation*}
    f \triangleq \sum_{t=1}^{T-m} \frac{1}{T-m} f_t,
\end{equation*}
where weights $\alpha_t = \frac{1}{T-m}$. Then 
\begin{equation*}
    \nabla f(\bx) = \sum_{t=1}^{T-m} \frac{1}{T-m} \nabla f_t(\bx). 
\end{equation*}
Note that $\nabla f(\hat{\bx}^s) \triangleq \nabla f^s$. Moreover, since $f_t$ is $L$ smooth, $f$ is $L$ smooth.
According to Lemma \ref{lem:l-smmoth}, we have
\begin{equation} 
    \begin{split}
        &f(\hat{\bx}^s) \\
     &\leq f(\hat{\bx}^{s-1}) + <\nabla  f^{s-1}, \hat{\bx}^{s}-\hat{\bx}^{s-1}> + \frac{L}{2}\|\hat{\bx}^{s}-\hat{\bx}^{s-1}\|_2 \\
      & = f(\hat{\bx}^{s-1}) \\
     & \quad + <\nabla  f^{s-1}, \hat{\bx}^{s-1} - \eta \text{Median}(\{ \nabla f_t(\hat{\bx}^{s-1})\}_{t=1}^T ) -\hat{\bx}^{s-1} 
    \\ & \quad + \eta^2 \frac{L}{2} \|\hat{\bx}^{s-1} - \eta \text{Median}(\{ \nabla f_t(\hat{\bx}^{s-1})\}_{t=1}^T )- \hat{\bx}^{s-1} \|_2^2 \\
    & = f(\hat{\bx}^{s-1}) + <\nabla  f^{s-1},\eta \text{Median}(\{ \nabla f_t(\hat{\bx}^{s-1})\}_{t=1}^T ) 
    \\ & \quad + \eta^2 \frac{L}{2} \|  \text{Median}(\{ \nabla f_t(\hat{\bx}^{s-1})\}_{t=1}^T ) \|_2^2 \\
     &\leq f(\hat{\bx}^{s-1}) - \frac{1}{2L}\|\nabla  f^{s-1} \|_2^2 + \frac{1}{2L} \calO(\kappa^2),
    \end{split}
\end{equation}
where the last inequality is due to $\eta= \frac{1}{L}$ and \textbf{Claim 1}. Consequently, we have
\begin{equation} \label{eq: non-convex}
    \begin{split}
        & f(\hat{\bx}^s) - f(\bx^*) \\
        &\leq f(\hat{\bx}^{s-1}) - f(\bx^*) - \frac{1}{2L}\|\nabla  f^{s-1} \|_2^2 + \frac{1}{2L} \calO(\kappa^2). 
\end{split}
\end{equation}
Sum up Eq. \eqref{eq: non-convex} over $s = 1,\cdots, R_g$, we obtain
\begin{equation} \label{eq:non-covex1}
\begin{split} 
     &f(\hat{\bx}^{R_g}) - f(\bx^*) \\
     & \leq f(\hat{\bx}^{0}) - f(\bx^*) - \frac{1}{2L}\sum_{i=0}^{R_g-1}\|\nabla  f^{R_g-1} \|_2^2 + \frac{R_g}{2L} \calO(\kappa^2). 
\end{split}
\end{equation}
Since, $f(\hat{\bx}^{R_g}) - f(\bx^*) \geq 0$
Simply Eq. \eqref{eq:non-covex1}, we get
\begin{equation*}
    \begin{split}
     \sum_{i=0}^{R_g-1}\|\nabla  f^{R_g-1} \|_2^2 \leq 2L(f(\hat{\bx}^{0}) - f(\bx^*)) + T\calO(\kappa^2). 
    \end{split}
\end{equation*}
Moreover, we obtain
\begin{equation}
    \begin{split}
        &\min_{s=1,\cdots, R_g} \|\sum_{t=1}^{T-m}\frac{1}{T-m}\nabla f_t(\bx^s)\|_2  = \min_{s=1,\cdots, R_g} \|\nabla  f^{s-1}\|_2 \\
        & \leq \sum_{s=1}^{R_g} \frac{1}{R_g} \|\nabla f^{s-1}\|_2 \\
        & \leq \sqrt{\frac{2L}{R_g}(f(\hat{\bx}^{0}) - f(\bx^*)) + \calO(\kappa)} \\
        & \leq \frac{\sqrt{2}}{R_g}\sqrt{f(\hat{\bx}^{0}) - f(\bx^*)} + \calO(\kappa),
    \end{split}
\end{equation}
which completes the proof.
\end{proof}

\newpage
\section{Ablation Study}

\subsection{Label Restoration and Batch Alignment}\label{label and alignment-app}
In our empirical study, we take the label as known \cite{zhao2020idlg, geiping2020inverting} due to the high restoration accuracy. The original label inference method \cite{zhao2020idlg} only works for a single image, \cite{yin2021see} extended it with column-wise inference, but one
limitation is that it assumed there are no repeated labels in the batch, which only holds when the batch size is much smaller than the number of classes. To restore label in a general setting, we propose to restore the label with a GIAs optimization process according to the gradients  in the last layer of network. 

\subsubsection{Label Restoration} Assume that $\bx^{FC}$ is the input to the fully connected layer in the network, which is noted as $\bw^{FC}$. The gradient $\nabla_{\bw^{FC}}\calL(\bw^{FC}, \bx^{FC})$ is known.  

\noindent\textbf{Step 1: Dummy data initialization in FC layer.}  
We initialize the restored batch input $\hat{\bx}^{FC}$ in the FC layer as a normal Gaussian distribution $\calN(0,1)$. The dimension of $\hat{\bx}^{FC}$ is $\mathbb{R}^{b\times M}$, where $b$ is the batch size of private data $\bx$, and $M$ is the width of fully connected layers. And $\hat{\by}$ is randomly initialized as $\mathbb{R}^{b\times N}$, where $b$ is the batch size and $N$ is the number of classes. 

\noindent\textbf{Step 2: Dummy label optimization.}  
We minimize 
\begin{equation}
\text{min}_{\hat{\bx}^{FC}, \hat{\by}} ||\nabla_{\bw^{FC}}\calL(\bw^{FC}, \hat{\bx}^{FC}, \hat{\by}) - \nabla_{\bw^{FC}}\calL(\bw^{FC}, \bx^{FC}, \by)||^2
\end{equation}
over fully connected layer $\bw^{FC}$. And afterwards, $\hat{\by}$ is output. Tab. \ref{tab: label-app} shows the comparisons of label restoration between our TGIAs-RO and \cite{yin2021see}, which demonstrates our proposed method has higher accuracy in label recovery.  





\begin{table}[h!]
\caption{\label{tab: label-app} Average label restoration accuracy for \cite{yin2021see} and TGIAs-RO on ResNet-ImageNet of different batch size.}
 \renewcommand{\arraystretch}{1}
\centering
\resizebox{0.48\textwidth}{!}{
\begin{tabular}{ccccc}
\toprule
\multirow{2}{*}{ \tabincell{c}{Batch \\ Size} } &\multicolumn{2}{c}{100 Classes} &\multicolumn{2}{c}{1000 Classes}
\\ \cmidrule(r){2-3} \cmidrule(r){4-5}
& \cite{yin2021see} & TGIAs-RO  &\cite{yin2021see} & TGIAs-RO   \\ \midrule
8 & 95.89\%    & 100\%    & 99.47\%    & 100\%    \\
16 &  91.84\%   & 99.86\%         & 99.37\%         & 100\%         \\ 
32 & 88.65\%    & 99.71\% & 99.19\% & 99.87\% \\
64 &  $\backslash$    & 98.56\%    & 98.21\%    & 99.47\%    \\
128 & $\backslash$   & 97.32\%    & 98.11\%    & 99.34\%    \\

 \bottomrule
\end{tabular}
}
\end{table}

\subsubsection{Batch Alignment}
Batch alignment is an important procedure to collect multiple temporal gradients for TGIAs-RO. We match batch ids according to the previously restored label in the following. In practical cross-silo federated learning, gradients are often exchanged in large batch size such as more than 32 ($b\geq 32$), as shown in Tab. \ref{tab: label-app}, TGIAs-RO restores label with accuracy higher than 98\%. Thus Batch id alignment accuracy between two epochs is also as high as more than 90\%.

\begin{table*}[htbp]
     \caption{Quantitative comparisons of TGIAs-RO and the  \textit{state-of-the-art} models  with varying batch sizes in different datasets and model architectures. Larger batch 
size increases reconstruction difficulty. TGIAs-RO always
surpasses those baseline methods thanks to the robust optimization enabled by temporal gradients. The bold font  denotes the best performance.}\label{tab: comparison}
  \renewcommand{\arraystretch}{0.95}
     \centering
     \resizebox{0.90\textwidth}{!}{
      \begin{tabular}[l]{ccccccccccc}
        \toprule
        \multirow{2}{*}{ Methods }& 
        \multirow{2}{*}{ Batchsize }&  \multicolumn{3}{c}{LeNet on MNIST} & \multicolumn{3}{c}{AlexNet on CIFAR10}   & \multicolumn{3}{c}{ResNet18 on ImageNet}     \\ \cmidrule(r){3-5} \cmidrule(r){6-8} \cmidrule(r){9-11}
 & & MSE & PSNR & SSIM  & MSE & PSNR & SSIM & MSE & PSNR & SSIM\\ \midrule
        
 \multirow{8}{*}{ DLG } 
 & 1& 0.01 & 14.73 & 0.65 & 0.02 & 16.77 & 0.72 & 0.03 & 15.73 & 0.65 \\
 & 2  & 0.03 & 12.73 & 0.51 
 & 0.03 & 15.73 & 0.56 & 0.04 &14.17 & 0.57 \\
 & 4 & 0.06 & 12.54 & 0.46 & 0.03 & 15.12 & 0.55 & 0.05 & 13.38 & 0.51 \\
 & 8 & 0.08 & 11.44 & 0.38 & 0.04 & 14.22 & 0.44 &  0.05 & 13.03 & 0.49 \\
 & 16 & 0.10 & 10.67 & 0.26 & 0.04 & 13.64 & 0.37 & 0.05 & 12.86 & 0.46\\
 & 32& 0.12 & 9.76 &0.20  & 0.05 & 12.67 & 0.21 & 0.05 & 12.76 & 0.43\\
 & 64& 0.13 & 9.23 &0.13  & 0.06 & 12.14 & 0.18 & 0.07 & 11.84 & 0.42\\
 & 128& 0.14 & 8.45 &0.12  & 0.07 & 11.74 & 0.13 & 0.07 & 11.74 & 0.40\\
\cmidrule(r){2-11}
  \multirow{8}{*}{ 
  \tabincell{c}{Inverting \\ Gradients}}
  & 1&  0.01 & 18.40 & 0.72 & 4$e^{-3}$ & 23.86 & 0.93 & 6$e^{-3}$ & 22.85 & 0.85\\
  & 2 & 0.02 & 17.04 & 0.62 & 5$e^{-3}$ & 23.06 & 0.83 & 0.01 & 22.15 & 0.79 \\
 & 4 & 0.04 & 13.10 & 0.47 & 0.01 & 18.40 & 0.67 & 0.01 & 20.90 & 0.72 \\
 & 8 & 0.10 & 12.34 & 0.32 & 0.03 & 17.45 & 0.48 & 0.03 & 17.65 & 0.66 \\
 & 16 & 0.10 & 11.98 & 0.25 & 0.03 & 15.38 & 0.49 & 0.04 & 15.43 & 0.55 \\
 & 32& 0.11 & 11.65 &0.24  & 0.04 & 14.28 & 0.37 & 0.04 & 13.38 & 0.47\\
 & 64& 0.11 & 10.32 &0.20  &0.04 & 13.89 & 0.32 & 0.05 & 12.99 & 0.44\\
 & 128& 0.12 & 9.54 &0.19  & 0.05 & 13.75 & 0.27 & 0.05 & 11.75 &0.43\\
\cmidrule(r){2-11}
 \multirow{8}{*}{ SAPAG }
 & 1& 0.10 & 9.88 & 0.11 & 0.07 & 11.41 & 0.12 & 0.09 & 10.41 &0.16\\
 & 2 & 0.10 & 9.87 & 0.07 & 0.07 & 11.43 & 0.12 & 0.09 & 10.34 & 0.14 \\
 & 4 & 0.11 & 9.47 &0.06 & 0.07 & 11.38 & 0.11 & 0.10 & 10.16 & 0.14\\
 & 8 & 0.14&8.64 &0.05 & 0.08 & 11.00 & 0.11 & 0.10 & 9.98 & 0.11 \\
 & 16 &0.14 &8.59 &0.05 &0.08 &10.92 & 0.09 &0.12 &9.20 &0.11 \\
 & 32& 0.23 & 6.84 & 0.07 & 0.09 & 10.22 & 0.08 & 0.12 & 9.12 &0.10\\
 & 64& 0.30 & 5.51 & 0.07 & 0.10 & 9.93 & 0.08 & 0.13 & 8.25 &0.09\\
 & 128& 0.31 & 5.21 & 0.07 & 0.11 & 9.84 & 0.07 & 0.21 & 6.84 &0.08\\
\cmidrule(r){2-11}
 \multirow{8}{*}{ BN Regularizer }
 & 1& 0.06 & 13.53 & 0.63  & 0.01 & 18.93 & 0.73 & 0.04 & 13.17 & 0.33\\
 & 2 & 0.07 & 13.41 & 0.33 & 0.01 & 18.34 & 0.73 & 0.07 & 11.55 & 0.32 \\
& 4 & 0.08 & 12.51 & 0.30 &0.02 & 17.30 & 0.68 & 0.07 & 11.32 & 0.28\\
& 8 & 0.08 & 12.50 & 0.28 & 0.02 & 15.94 & 0.65 & 0.08 & 10.91 & 0.26\\
& 16 & 0.08 & 11.26 & 0.26 & 0.02 & 15.70 & 0.60  & 0.09 & 10.79 & 0.24\\
& 32 & 0.09 & 11.14 & 0.21 & 0.03 & 15.54 & 0.49 & 0.09 & 10.74 & 0.23\\
& 64& 0.10 & 10.32 & 0.19 & 0.03 & 14.94 & 0.39 & 0.10 & 10.67 & 0.22\\
& 128& 0.12 & 9.54 & 0.13 & 0.04 & 13.43 & 0.32 & 0.10 & 10.14 & 0.19\\
\cmidrule(r){2-11}       
 \multirow{8}{*}{ GC Regularizer }
 & 1& 0.03 & 13.92 & 0.58 & 8$e^{-3}$ &20.14 & 0.76 & 0.02 & 19.45 &0.75\\
 &2  & 0.04 & 13.47 & 0.29 & 0.01 & 19.77 & 0.70 & 0.03 & 17.43 & 0.69 \\
& 4 & 0.04 & 13.31 & 0.27 & 0.02 & 18.93 & 0.71 & 0.03 & 17.20 & 0.65 \\
& 8 & 0.05 & 12.63 & 0.26 & 0.02 & 17.76 & 0.67 & 0.04 & 14.02 & 0.55 \\
& 16 & 0.07 & 12.03 & 0.23 & 0.03 & 16.64 & 0.58 & 0.04 & 14.01 & 0.44 \\
& 32& 0.08 & 11.66 & 0.13 & 0.03 & 16.03 & 0.54 & 0.05 & 12.23 &0.43\\
& 64& 0.10 & 11.00 & 0.13 & 0.03 & 15.63 & 0.48 & 0.07 & 11.69 & 0.43\\
& 128& 0.11 & 10.78 & 0.11 & 0.04 & 14.88 & 0.46 & 0.09 & 10.88 &0.41\\
\cmidrule(r){2-11}
\multirow{8}{*}{  TGIAs-RO}
& 1& \textbf{4$e^{-3}$} & \textbf{23.32} & \textbf{0.79} & \textbf{1$e^{-3}$} & \textbf{30.17} & \textbf{0.94} & \textbf{2$e^{-3}$} & \textbf{27.53}  &\textbf{0.94}\\
&2 & \textbf{6$e^{-3}$} & \textbf{21.64} & \textbf{0.76} & \textbf{2$e^{-3}$} & \textbf{27.51} & \textbf{0.93} & \textbf{2$e^{-3}$} & \textbf{26.51} & \textbf{0.94} \\ 
& 4 & \textbf{0.01}  &\textbf{20.57} & \textbf{0.71} & \textbf{3$e^{-3}$} & \textbf{25.45} & \textbf{0.92} & \textbf{4$e^{-3}$} & \textbf{24.85} & \textbf{0.91}\\ 
& 8 & \textbf{0.02} & \textbf{17.90} & \textbf{0.69} & \textbf{5$e^{-3}$} & \textbf{24.36} &\textbf{0.91} & \textbf{6$e^{-3}$} & \textbf{22.19} & \textbf{0.91}\\
& 16 & \textbf{0.03} & \textbf{15.90} & \textbf{0.64} & \textbf{6$e^{-3}$} & \textbf{22.32} & \textbf{0.86} & \textbf{0.01} & \textbf{19.35} & \textbf{0.79}\\ 
& 32& \textbf{0.05} & \textbf{13.05} & \textbf{0.40} 
&\textbf{0.02} & \textbf{18.48} & \textbf{0.78} & \textbf{0.02} & \textbf{17.43} &\textbf{0.68}\\
& 64& \textbf{0.09} & \textbf{11.74} & \textbf{0.27} 
& \textbf{0.03} & \textbf{16.48} &\textbf{0.65}  & \textbf{0.03} & \textbf{15.38} &\textbf{0.61}\\
& 128& \textbf{0.10} & \textbf{11.39} & \textbf{0.21}  
& \textbf{0.03} & \textbf{15.88} & \textbf{0.54} & \textbf{0.05} & \textbf{13.18} &\textbf{0.55}\\
      \bottomrule
      \end{tabular}}
 
\end{table*}
\subsection{Data recovery via local updating step $R_l$}

When we implement TGIAs-RO, the local optimization step for each temporal gradients $R_l$ in Algo. \ref{algo:TGIAs-RO} is significant (See Sect. \ref{sec:method}). And Tab. \ref{tab: LBFGS_Step} validates that  larger $R_l$ lead to images with a higher quality.
\begin{figure} [htbp]
\vspace{-1pt}
\begin{subfigure}[t]{1\linewidth}
\centering
\begin{minipage}[t]{0.15\textwidth}
\centering
\includegraphics[width=1.2cm]{imgs/Images/ground_truth/cock.jpeg}
\end{minipage}
\begin{minipage}[t]{0.15\textwidth}
\centering
\includegraphics[width=1.2cm]{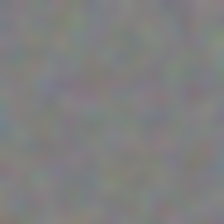}
\end{minipage}
\begin{minipage}[t]{0.15\textwidth}
\centering
\includegraphics[width=1.2cm]{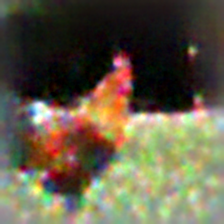}
\end{minipage}
\begin{minipage}[t]{0.15\textwidth}
\centering
\includegraphics[width=1.2cm]{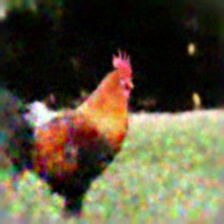}
\end{minipage}
\begin{minipage}[t]{0.15\textwidth}
\centering
\includegraphics[width=1.2cm]{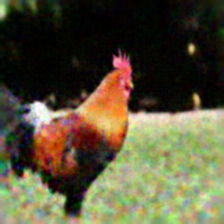}
\end{minipage}
\begin{minipage}[t]{0.15\textwidth}
\centering
\includegraphics[width=1.2cm]{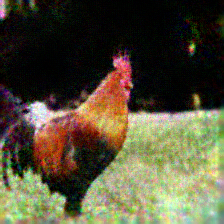}
\end{minipage}
\end{subfigure}

\begin{subfigure}[t]{1\linewidth}
\centering
\begin{minipage}[t]{0.15\textwidth}
\centering
\includegraphics[width=1.2cm]{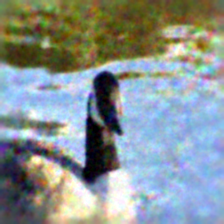}
\end{minipage}
\begin{minipage}[t]{0.15\textwidth}
\centering
\includegraphics[width=1.2cm]{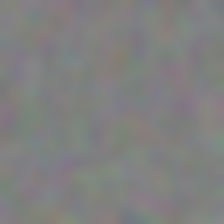}
\end{minipage}
\begin{minipage}[t]{0.15\textwidth}
\centering
\includegraphics[width=1.2cm]{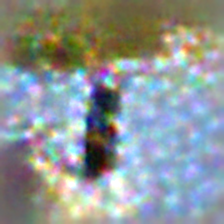}
\end{minipage}
\begin{minipage}[t]{0.15\textwidth}
\centering
\includegraphics[width=1.2cm]{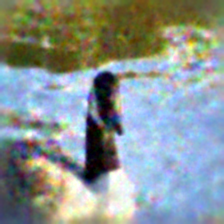}
\end{minipage}
\begin{minipage}[t]{0.15\textwidth}
\centering
\includegraphics[width=1.2cm]{imgs/ablation_study/LBFGS_Steps/10_2.png}
\end{minipage}
\begin{minipage}[t]{0.15\textwidth}
\centering
\includegraphics[width=1.2cm]{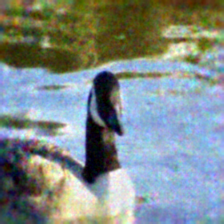}
\end{minipage}
\end{subfigure}

\begin{subfigure}[t]{1\linewidth}
\centering
\begin{minipage}[t]{0.15\textwidth}
\centering
\small Original
\end{minipage}
\begin{minipage}[t]{0.15\textwidth}
\centering
\small 2 steps
\end{minipage}
\begin{minipage}[t]{0.15\textwidth}
\centering
\small 4 steps
\end{minipage}
\begin{minipage}[t]{0.15\textwidth}
\centering
\small 8 steps
\end{minipage}
\begin{minipage}[t]{0.15\textwidth}
\centering
\small 10 steps
\end{minipage}
\begin{minipage}[t]{0.15\textwidth}
\centering
\small 20 steps
\end{minipage}
\end{subfigure}
\vspace{-15pt}
 \caption{\small Visualization of restored images for varying LBFGS Steps with batch size 8 on ImageNet and ResNet-18. } 
  \label{fig: LBFGS_Step}
  \vspace{-10pt}
\end{figure}

\begin{table}[H]
\caption{Comparisons of restored images with different updating steps $R_l$ in GIAs optimization, the model architecture adopted is AlexNet and dataset is ImageNet (batch size = 8).}\label{tab: LBFGS_Step}

 \renewcommand{\arraystretch}{1}
 \centering
\resizebox{0.40\textwidth}{!}{
\begin{tabular}{cccc}
\toprule
LBFGS Steps & MSE   & PSNR/dB & SSIM       \\ \midrule
2 step & 0.047 & 13.21  &  0.43  \\
4 steps & 0.030   & 16.05  & 0.60  \\ 
8 steps & 0.016 & 19.35 & 0.73 \\
10 steps & 0.014   & 20.18    & 0.76   \\
20 steps & 8$e^{-3}$    & 23.19    & 0.91  \\
 \bottomrule
\end{tabular}
}

\end{table}

\subsection{Trained v.s. Untrained Temporal Gradients}
We compare the TGIAs-RO from the first 10 temporal gradients with TGIAs-RO with 10 untrained model gradients. The results in Tab. \ref{tab: Untrained} show that the PSNR and SSIM with gradients of untrained models are greatly higher than 10 trained gradients, which indicates that temporal gradients from the early stages encode more private information, which is consistent with our findings in Sect. \ref{sec:failure}. 

\begin{table}[H]
\caption{Comparisons of restored images with different iterations of temporal gradients, the model architecture adopted is AlexNet and dataset is CIFAR10 (batch size = 8).}\label{tab: Untrained}
 \renewcommand{\arraystretch}{1}
 \centering
\resizebox{0.48\textwidth}{!}{
\begin{tabular}{ccccc}
\toprule
\multirow{2}{*}{ Batch Size } &\multicolumn{2}{c}{First 10 temporal} &\multicolumn{2}{c}{10 Untrained}
\\ \cmidrule(r){2-3} \cmidrule(r){4-5}
&  PSNR/dB & SSIM     & PSNR/dB & SSIM    \\ \midrule
1  & 30.15  &  0.94  & 58.36  &  0.99\\
2    & 27.51   & 0.93  & 46.31    & 0.99 \\ 
4   & 25.45 & 0.92 & 38.55    & 0.99 \\
8    & 24.36 & 0.91  & 31.87    & 0.95  \\
16 &  22.32 & 0.86 & 27.19    & 0.91  \\
32 &  18.48 & 0.78 & 22.19    & 0.91  \\
 \bottomrule
\end{tabular}
}
\end{table}

\subsection{More Experimental Results}
\subsubsection{More Details on TGIAs-RO v.s. the State-of-the-art Methods}
\begin{figure}[H] 

    \vspace{-10pt}
	\centering
	 \begin{subfigure}{0.24\textwidth}
		\centering
		\includegraphics[keepaspectratio=true, width=120pt]{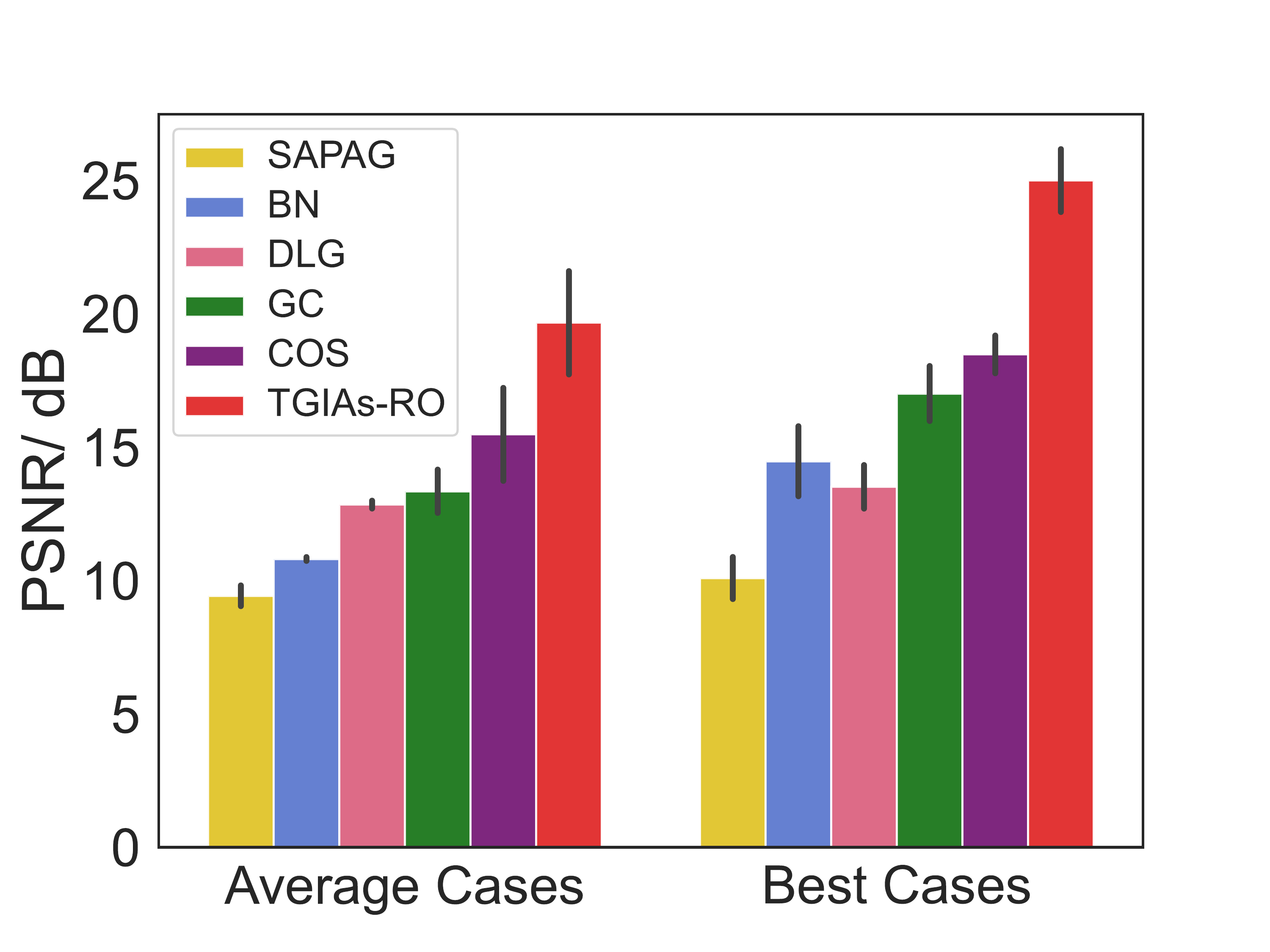}
		\subcaption{\small Comparisons of PNSR }
	  \end{subfigure}
	 \begin{subfigure}{0.24\textwidth}
		\centering
		\includegraphics[keepaspectratio=true, width=120pt]{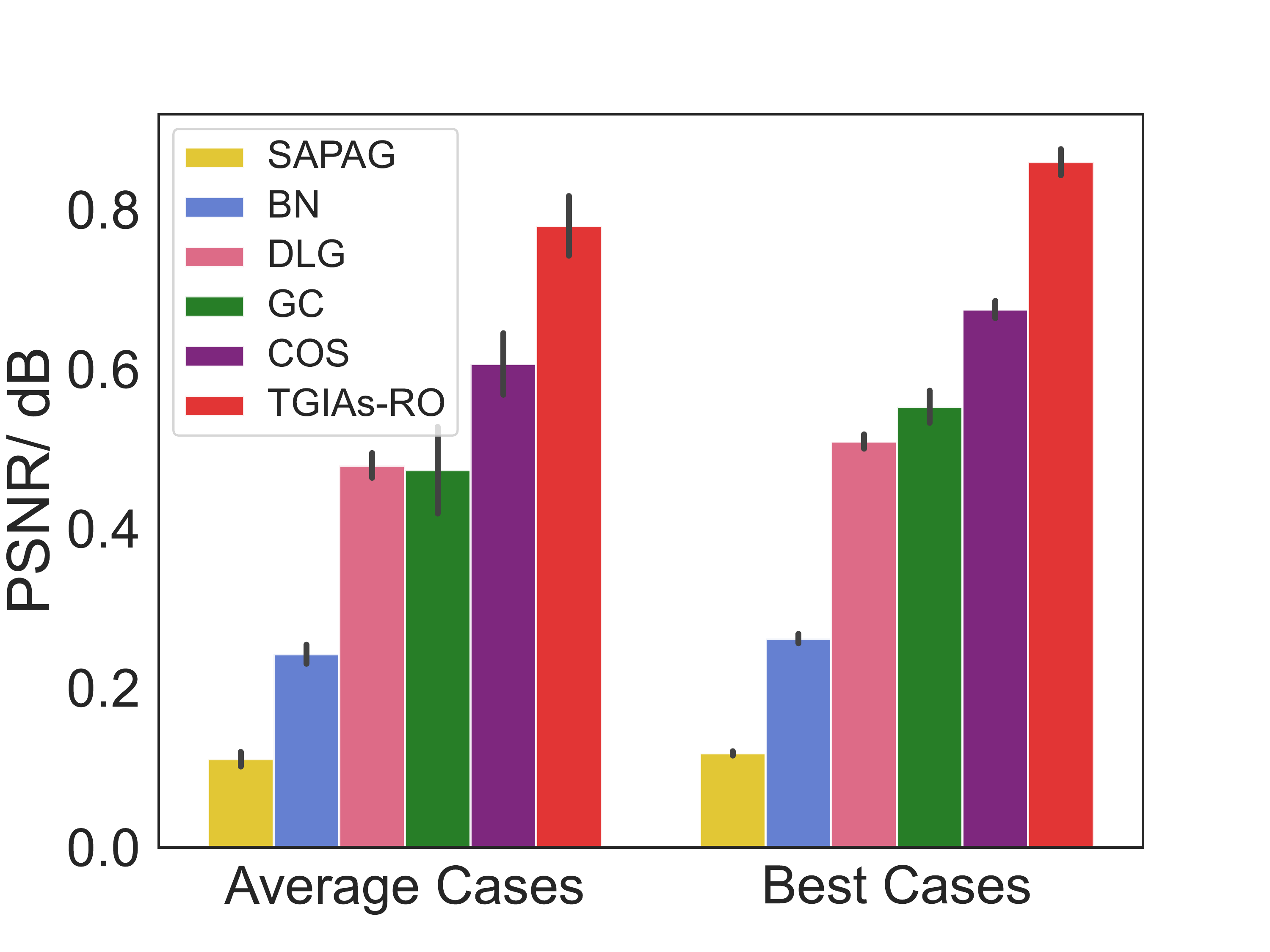}
		\subcaption{\small Comparisons of SSIM}
	  \end{subfigure}
    \vspace{-5pt}
	\centering
	\caption{Comparisons of our proposed TGIAs-RO (with 10 temporal gradients) and the \textit{state-of-the-art} methods against differential privacy. Figure provides the average and the best performances of different methods in a batch size of 16.}\label{fig: bar_comparsion}
\vspace{-10pt}
\end{figure}

\begin{figure*}[h!]
  \centering
 \vspace{-10pt}
 \begin{subfigure}[t]{1\linewidth}
\centering
\begin{minipage}[c]{0.07\textwidth}
\vspace{-4em} \scriptsize Ground\\ Truth 
\end{minipage}
\begin{minipage}[t]{0.105\textwidth}
\centering
\includegraphics[width=1.5cm]{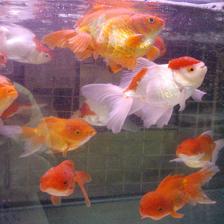}
\end{minipage}
\begin{minipage}[t]{0.105\textwidth}
\centering
\includegraphics[width=1.5cm]{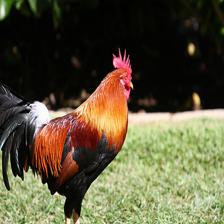}
\end{minipage}
\begin{minipage}[t]{0.105\textwidth}
\centering
\includegraphics[width=1.5cm]{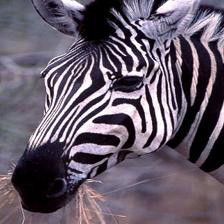}
\end{minipage}
\begin{minipage}[t]{0.105\textwidth}
\centering
\includegraphics[width=1.5cm]{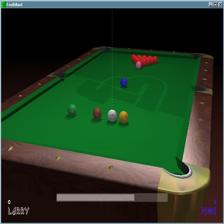}
\end{minipage}
\begin{minipage}[t]{0.105\textwidth}
\centering
\includegraphics[width=1.5cm]{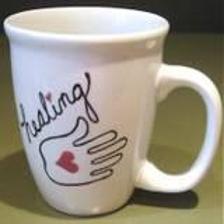}
\end{minipage}
\begin{minipage}[t]{0.105\textwidth}
\centering
\includegraphics[width=1.5cm]{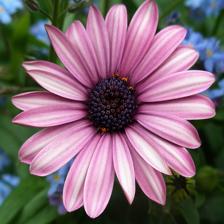}
\end{minipage}
\begin{minipage}[t]{0.105\textwidth}
\centering
\includegraphics[width=1.5cm]{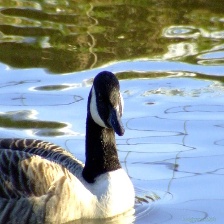}
\end{minipage}
\begin{minipage}[t]{0.105\textwidth}
\centering
\includegraphics[width=1.5cm]{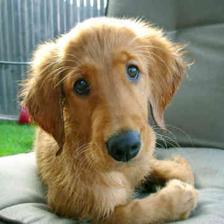}
\end{minipage}
\end{subfigure}
 
 \begin{subfigure}[t]{1\linewidth}
\centering
\begin{minipage}[c]{0.07\textwidth}
\vspace{-4em} \scriptsize TGIAs-RO 
\end{minipage}
\begin{minipage}[t]{0.105\textwidth}
\centering
\includegraphics[width=1.5cm]{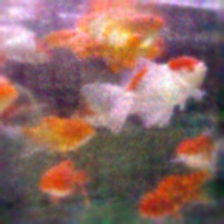}
\end{minipage}
\begin{minipage}[t]{0.105\textwidth}
\centering
\includegraphics[width=1.5cm]{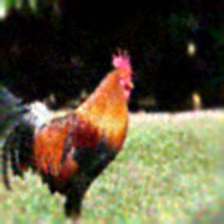}
\end{minipage}
\begin{minipage}[t]{0.105\textwidth}
\centering
\includegraphics[width=1.5cm]{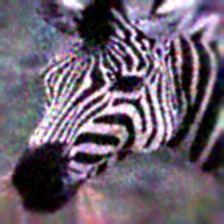}
\end{minipage}
\begin{minipage}[t]{0.105\textwidth}
\centering
\includegraphics[width=1.5cm]{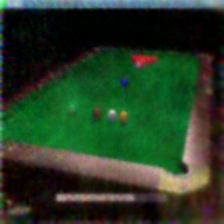}
\end{minipage}
\begin{minipage}[t]{0.105\textwidth}
\centering
\includegraphics[width=1.5cm]{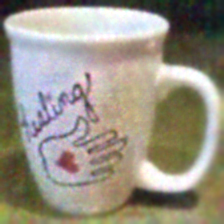}
\end{minipage}
\begin{minipage}[t]{0.105\textwidth}
\centering
\includegraphics[width=1.5cm]{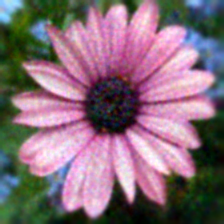}
\end{minipage}
\begin{minipage}[t]{0.105\textwidth}
\centering
\includegraphics[width=1.5cm]{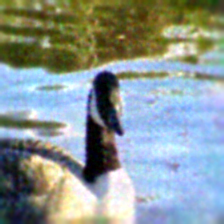}
\end{minipage}
\begin{minipage}[t]{0.105\textwidth}
\centering
\includegraphics[width=1.5cm]{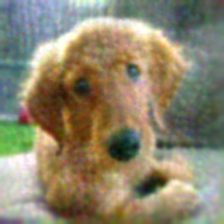}
\end{minipage}
\end{subfigure}

 \begin{subfigure}[t]{1\linewidth}
\centering
\begin{minipage}[c]{0.07\textwidth}
\vspace{-4em} \tiny Cosine\\ Similarity 
\end{minipage}
\begin{minipage}[t]{0.105\textwidth}
\centering
\includegraphics[width=1.5cm]{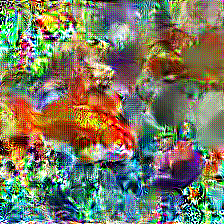}
\end{minipage}
\begin{minipage}[t]{0.105\textwidth}
\centering
\includegraphics[width=1.5cm]{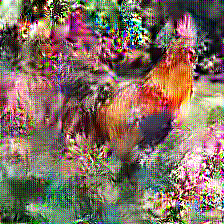}
\end{minipage}
\begin{minipage}[t]{0.105\textwidth}
\centering
\includegraphics[width=1.5cm]{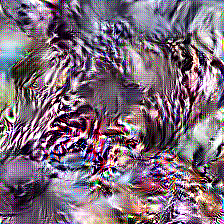}
\end{minipage}
\begin{minipage}[t]{0.105\textwidth}
\centering
\includegraphics[width=1.5cm]{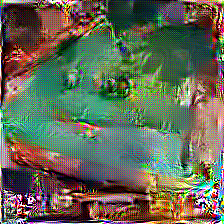}
\end{minipage}
\begin{minipage}[t]{0.105\textwidth}
\centering
\includegraphics[width=1.5cm]{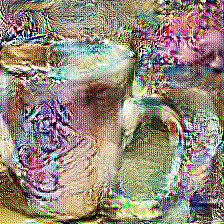}
\end{minipage}
\begin{minipage}[t]{0.105\textwidth}
\centering
\includegraphics[width=1.5cm]{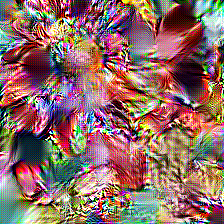}
\end{minipage}
\begin{minipage}[t]{0.105\textwidth}
\centering
\includegraphics[width=1.5cm]{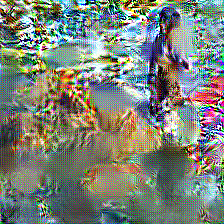}
\end{minipage}
\begin{minipage}[t]{0.105\textwidth}
\centering
\includegraphics[width=1.5cm]{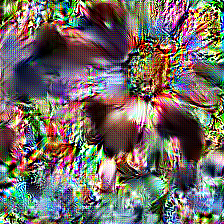}
\end{minipage}
\end{subfigure}

 \begin{subfigure}[t]{1\linewidth}
\centering
\begin{minipage}[c]{0.07\textwidth}
\vspace{-4em} \tiny GC\\ Regularizer 
\end{minipage}
\begin{minipage}[t]{0.105\textwidth}
\centering
\includegraphics[width=1.5cm]{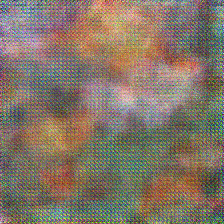}
\end{minipage}
\begin{minipage}[t]{0.105\textwidth}
\centering
\includegraphics[width=1.5cm]{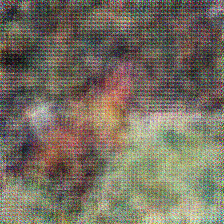}
\end{minipage}
\begin{minipage}[t]{0.105\textwidth}
\centering
\includegraphics[width=1.5cm]{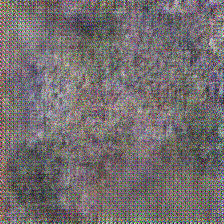}
\end{minipage}
\begin{minipage}[t]{0.105\textwidth}
\centering
\includegraphics[width=1.5cm]{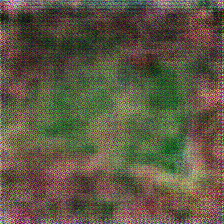}
\end{minipage}
\begin{minipage}[t]{0.105\textwidth}
\centering
\includegraphics[width=1.5cm]{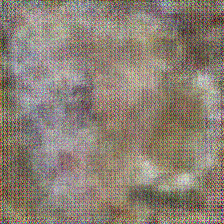}
\end{minipage}
\begin{minipage}[t]{0.105\textwidth}
\centering
\includegraphics[width=1.5cm]{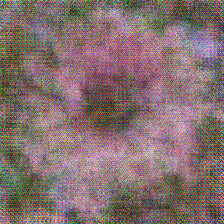}
\end{minipage}
\begin{minipage}[t]{0.105\textwidth}
\centering
\includegraphics[width=1.5cm]{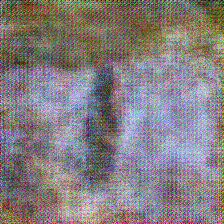}
\end{minipage}
\begin{minipage}[t]{0.105\textwidth}
\centering
\includegraphics[width=1.5cm]{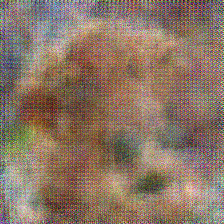}
\end{minipage}
\end{subfigure}

 \begin{subfigure}[t]{1\linewidth}
\centering
\begin{minipage}[c]{0.07\textwidth}
\vspace{-4em} \tiny BN\\ Regularizer 
\end{minipage}
\begin{minipage}[t]{0.105\textwidth}
\centering
\includegraphics[width=1.5cm]{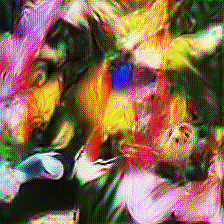}
\end{minipage}
\begin{minipage}[t]{0.105\textwidth}
\centering
\includegraphics[width=1.5cm]{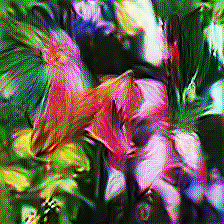}
\end{minipage}
\begin{minipage}[t]{0.105\textwidth}
\centering
\includegraphics[width=1.5cm]{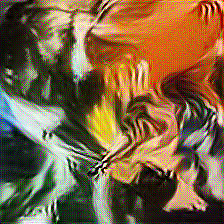}
\end{minipage}
\begin{minipage}[t]{0.105\textwidth}
\centering
\includegraphics[width=1.5cm]{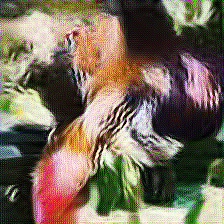}
\end{minipage}
\begin{minipage}[t]{0.105\textwidth}
\centering
\includegraphics[width=1.5cm]{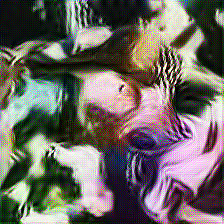}
\end{minipage}
\begin{minipage}[t]{0.105\textwidth}
\centering
\includegraphics[width=1.5cm]{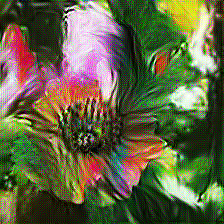}
\end{minipage}
\begin{minipage}[t]{0.105\textwidth}
\centering
\includegraphics[width=1.5cm]{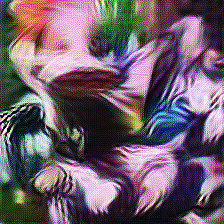}
\end{minipage}
\begin{minipage}[t]{0.105\textwidth}
\centering
\includegraphics[width=1.5cm]{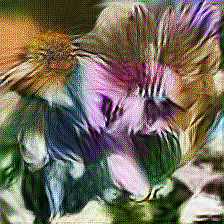}
\end{minipage}
\end{subfigure}

 \begin{subfigure}[t]{1\linewidth}
\centering
\begin{minipage}[c]{0.07\textwidth}
\vspace{-4em} \scriptsize DLG 
\end{minipage}
\begin{minipage}[t]{0.105\textwidth}
\centering
\includegraphics[width=1.5cm]{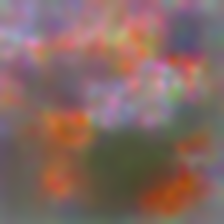}
\end{minipage}
\begin{minipage}[t]{0.105\textwidth}
\centering
\includegraphics[width=1.5cm]{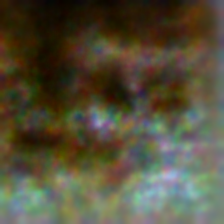}
\end{minipage}
\begin{minipage}[t]{0.105\textwidth}
\centering
\includegraphics[width=1.5cm]{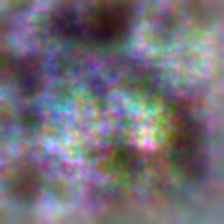}
\end{minipage}
\begin{minipage}[t]{0.105\textwidth}
\centering
\includegraphics[width=1.5cm]{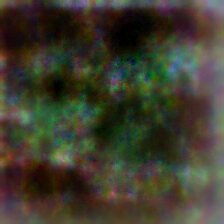}
\end{minipage}
\begin{minipage}[t]{0.105\textwidth}
\centering
\includegraphics[width=1.5cm]{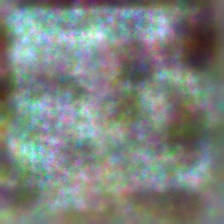}
\end{minipage}
\begin{minipage}[t]{0.105\textwidth}
\centering
\includegraphics[width=1.5cm]{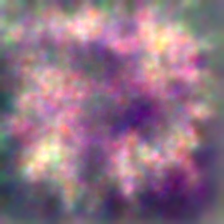}
\end{minipage}
\begin{minipage}[t]{0.105\textwidth}
\centering
\includegraphics[width=1.5cm]{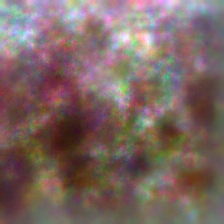}
\end{minipage}
\begin{minipage}[t]{0.105\textwidth}
\centering
\includegraphics[width=1.5cm]{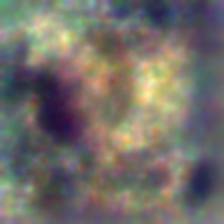}
\end{minipage}
\end{subfigure}

 \begin{subfigure}[t]{1\linewidth}
\centering
\begin{minipage}[c]{0.07\textwidth}
\vspace{-4em} \scriptsize SAPAG 
\end{minipage}
\begin{minipage}[t]{0.105\textwidth}
\centering
\includegraphics[width=1.5cm]{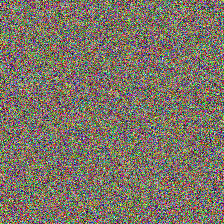}
\end{minipage}
\begin{minipage}[t]{0.105\textwidth}
\centering
\includegraphics[width=1.5cm]{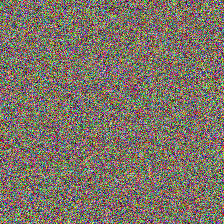}
\end{minipage}
\begin{minipage}[t]{0.105\textwidth}
\centering
\includegraphics[width=1.5cm]{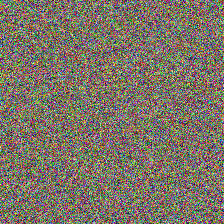}
\end{minipage}
\begin{minipage}[t]{0.105\textwidth}
\centering
\includegraphics[width=1.5cm]{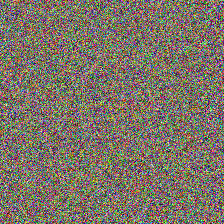}
\end{minipage}
\begin{minipage}[t]{0.105\textwidth}
\centering
\includegraphics[width=1.5cm]{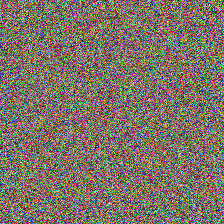}
\end{minipage}
\begin{minipage}[t]{0.105\textwidth}
\centering
\includegraphics[width=1.5cm]{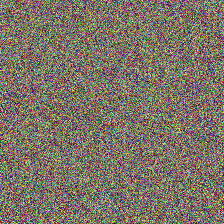}
\end{minipage}
\begin{minipage}[t]{0.105\textwidth}
\centering
\includegraphics[width=1.5cm]{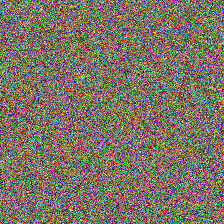}
\end{minipage}
\begin{minipage}[t]{0.105\textwidth}
\centering
\includegraphics[width=1.5cm]{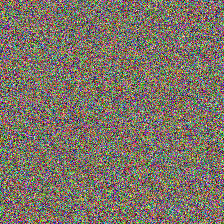}
\end{minipage}
\end{subfigure}

  \caption{ Visual comparisons between TGIAs-RO (our method with 10 temporal gradients) with the state-of-the-art single-temporal gradient inversion 
attacks including DLG \cite{zhu2019deep}, Cosine similarity \cite{geiping2020inverting}, SAPAG \cite{wang2020sapag}, BN regularzier and GC
regularizer \cite{yin2021see} on Imagenet dataset (batch size = 8, image size $= 3 \times 224 \times 224$).} \label{visual1-app}

\vspace{-5pt}
\end{figure*}

\begin{figure*}[htbp]
  \centering
 \vspace{-10pt}
 \begin{subfigure}[t]{1\linewidth}
\centering
\begin{minipage}[c]{0.08\textwidth}
\vspace{-2em} \scriptsize Ground\\ Truth 
\end{minipage}
\begin{minipage}[t]{0.08\textwidth}
\centering
\includegraphics[width=1.2cm]{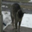}
\end{minipage}
\begin{minipage}[t]{0.08\textwidth}
\centering
\includegraphics[width=1.2cm]{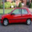}
\end{minipage}
\begin{minipage}[t]{0.08\textwidth}
\centering
\includegraphics[width=1.2cm]{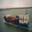}
\end{minipage}
\begin{minipage}[t]{0.08\textwidth}
\centering
\includegraphics[width=1.2cm]{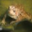}
\end{minipage}
\begin{minipage}[t]{0.08\textwidth}
\centering
\includegraphics[width=1.2cm]{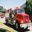}
\end{minipage}
\begin{minipage}[t]{0.08\textwidth}
\centering
\includegraphics[width=1.2cm]{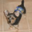}
\end{minipage}
\begin{minipage}[t]{0.08\textwidth}
\centering
\includegraphics[width=1.2cm]{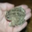}
\end{minipage}
\begin{minipage}[t]{0.08\textwidth}
\centering
\includegraphics[width=1.2cm]{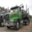}
\end{minipage}
\begin{minipage}[t]{0.08\textwidth}
\centering
\includegraphics[width=1.2cm]{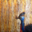}
\end{minipage}
\begin{minipage}[t]{0.08\textwidth}
\centering
\includegraphics[width=1.2cm]{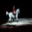}
\end{minipage}
\end{subfigure}
 
  \begin{subfigure}[t]{1\linewidth}
\centering
\begin{minipage}[c]{0.08\textwidth}
\vspace{-2em} \scriptsize Restored\\ Image 
\end{minipage}
\begin{minipage}[t]{0.08\textwidth}
\centering
\includegraphics[width=1.2cm]{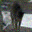}
\end{minipage}
\begin{minipage}[t]{0.08\textwidth}
\centering
\includegraphics[width=1.2cm]{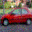}
\end{minipage}
\begin{minipage}[t]{0.08\textwidth}
\centering
\includegraphics[width=1.2cm]{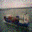}
\end{minipage}
\begin{minipage}[t]{0.08\textwidth}
\centering
\includegraphics[width=1.2cm]{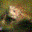}
\end{minipage}
\begin{minipage}[t]{0.08\textwidth}
\centering
\includegraphics[width=1.2cm]{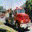}
\end{minipage}
\begin{minipage}[t]{0.08\textwidth}
\centering
\includegraphics[width=1.2cm]{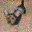}
\end{minipage}
\begin{minipage}[t]{0.08\textwidth}
\centering
\includegraphics[width=1.2cm]{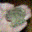}
\end{minipage}
\begin{minipage}[t]{0.08\textwidth}
\centering
\includegraphics[width=1.2cm]{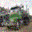}
\end{minipage}
\begin{minipage}[t]{0.08\textwidth}
\centering
\includegraphics[width=1.2cm]{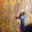}
\end{minipage}
\begin{minipage}[t]{0.08\textwidth}
\centering
\includegraphics[width=1.2cm]{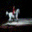}
\end{minipage}
\end{subfigure}

 \begin{subfigure}[t]{1\linewidth}
\centering
\begin{minipage}[c]{0.08\textwidth}
\vspace{-2em} \scriptsize Ground\\ Truth 
\end{minipage}
\begin{minipage}[t]{0.08\textwidth}
\centering
\includegraphics[width=1.2cm]{imgs/visual/gt_8.png}
\end{minipage}
\begin{minipage}[t]{0.08\textwidth}
\centering
\includegraphics[width=1.2cm]{imgs/visual/gt_6.png}
\end{minipage}
\begin{minipage}[t]{0.08\textwidth}
\centering
\includegraphics[width=1.2cm]{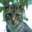}
\end{minipage}
\begin{minipage}[t]{0.08\textwidth}
\centering
\includegraphics[width=1.2cm]{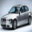}
\end{minipage}
\begin{minipage}[t]{0.08\textwidth}
\centering
\includegraphics[width=1.2cm]{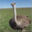}
\end{minipage}
\begin{minipage}[t]{0.08\textwidth}
\centering
\includegraphics[width=1.2cm]{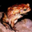}
\end{minipage}
\begin{minipage}[t]{0.08\textwidth}
\centering
\includegraphics[width=1.2cm]{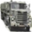}
\end{minipage}
\begin{minipage}[t]{0.08\textwidth}
\centering
\includegraphics[width=1.2cm]{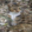}
\end{minipage}
\begin{minipage}[t]{0.08\textwidth}
\centering
\includegraphics[width=1.2cm]{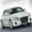}
\end{minipage}
\begin{minipage}[t]{0.08\textwidth}
\centering
\includegraphics[width=1.2cm]{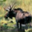}
\end{minipage}
\end{subfigure}
 
  \begin{subfigure}[t]{1\linewidth}
\centering
\begin{minipage}[c]{0.08\textwidth}
\vspace{-2em} \scriptsize Restored\\ Image 
\end{minipage}
\begin{minipage}[t]{0.08\textwidth}
\centering
\includegraphics[width=1.2cm]{imgs/visual/11.png}
\end{minipage}
\begin{minipage}[t]{0.08\textwidth}
\centering
\includegraphics[width=1.2cm]{imgs/visual/12.png}
\end{minipage}
\begin{minipage}[t]{0.08\textwidth}
\centering
\includegraphics[width=1.2cm]{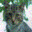}
\end{minipage}
\begin{minipage}[t]{0.08\textwidth}
\centering
\includegraphics[width=1.2cm]{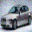}
\end{minipage}
\begin{minipage}[t]{0.08\textwidth}
\centering
\includegraphics[width=1.2cm]{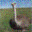}
\end{minipage}
\begin{minipage}[t]{0.08\textwidth}
\centering
\includegraphics[width=1.2cm]{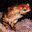}
\end{minipage}
\begin{minipage}[t]{0.08\textwidth}
\centering
\includegraphics[width=1.2cm]{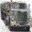}
\end{minipage}
\begin{minipage}[t]{0.08\textwidth}
\centering
\includegraphics[width=1.2cm]{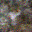}
\end{minipage}
\begin{minipage}[t]{0.08\textwidth}
\centering
\includegraphics[width=1.2cm]{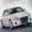}
\end{minipage}
\begin{minipage}[t]{0.08\textwidth}
\centering
\includegraphics[width=1.2cm]{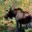}
\end{minipage}
\end{subfigure}

 \begin{subfigure}[t]{1\linewidth}
\centering
\begin{minipage}[c]{0.08\textwidth}
\vspace{-2em} \scriptsize Ground\\ Truth 
\end{minipage}
\begin{minipage}[t]{0.08\textwidth}
\centering
\includegraphics[width=1.2cm]{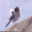}
\end{minipage}
\begin{minipage}[t]{0.08\textwidth}
\centering
\includegraphics[width=1.2cm]{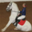}
\end{minipage}
\begin{minipage}[t]{0.08\textwidth}
\centering
\includegraphics[width=1.2cm]{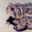}
\end{minipage}
\begin{minipage}[t]{0.08\textwidth}
\centering
\includegraphics[width=1.2cm]{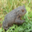}
\end{minipage}
\begin{minipage}[t]{0.08\textwidth}
\centering
\includegraphics[width=1.2cm]{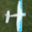}
\end{minipage}
\begin{minipage}[t]{0.08\textwidth}
\centering
\includegraphics[width=1.2cm]{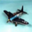}
\end{minipage}
\begin{minipage}[t]{0.08\textwidth}
\centering
\includegraphics[width=1.2cm]{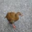}
\end{minipage}
\begin{minipage}[t]{0.08\textwidth}
\centering
\includegraphics[width=1.2cm]{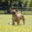}
\end{minipage}
\begin{minipage}[t]{0.08\textwidth}
\centering
\includegraphics[width=1.2cm]{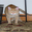}
\end{minipage}
\begin{minipage}[t]{0.08\textwidth}
\centering
\includegraphics[width=1.2cm]{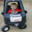}
\end{minipage}
\end{subfigure}
 
  \begin{subfigure}[t]{1\linewidth}
\centering
\begin{minipage}[c]{0.08\textwidth}
\vspace{-2em} \scriptsize Restored\\ Image 
\end{minipage}
\begin{minipage}[t]{0.08\textwidth}
\centering
\includegraphics[width=1.2cm]{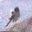}
\end{minipage}
\begin{minipage}[t]{0.08\textwidth}
\centering
\includegraphics[width=1.2cm]{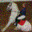}
\end{minipage}
\begin{minipage}[t]{0.08\textwidth}
\centering
\includegraphics[width=1.2cm]{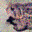}
\end{minipage}
\begin{minipage}[t]{0.08\textwidth}
\centering
\includegraphics[width=1.2cm]{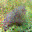}
\end{minipage}
\begin{minipage}[t]{0.08\textwidth}
\centering
\includegraphics[width=1.2cm]{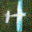}
\end{minipage}
\begin{minipage}[t]{0.08\textwidth}
\centering
\includegraphics[width=1.2cm]{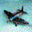}
\end{minipage}
\begin{minipage}[t]{0.08\textwidth}
\centering
\includegraphics[width=1.2cm]{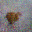}
\end{minipage}
\begin{minipage}[t]{0.08\textwidth}
\centering
\includegraphics[width=1.2cm]{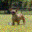}
\end{minipage}
\begin{minipage}[t]{0.08\textwidth}
\centering
\includegraphics[width=1.2cm]{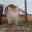}
\end{minipage}
\begin{minipage}[t]{0.08\textwidth}
\centering
\includegraphics[width=1.2cm]{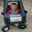}
\end{minipage}
\end{subfigure}

 \begin{subfigure}[t]{1\linewidth}
\centering
\begin{minipage}[c]{0.08\textwidth}
\vspace{-2em} \scriptsize Ground\\ Truth 
\end{minipage}
\begin{minipage}[t]{0.08\textwidth}
\centering
\includegraphics[width=1.2cm]{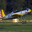}
\end{minipage}
\begin{minipage}[t]{0.08\textwidth}
\centering
\includegraphics[width=1.2cm]{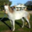}
\end{minipage}
\begin{minipage}[t]{0.08\textwidth}
\centering
\includegraphics[width=1.2cm]{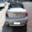}
\end{minipage}
\begin{minipage}[t]{0.08\textwidth}
\centering
\includegraphics[width=1.2cm]{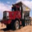}
\end{minipage}
\begin{minipage}[t]{0.08\textwidth}
\centering
\includegraphics[width=1.2cm]{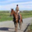}
\end{minipage}
\begin{minipage}[t]{0.08\textwidth}
\centering
\includegraphics[width=1.2cm]{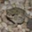}
\end{minipage}
\begin{minipage}[t]{0.08\textwidth}
\centering
\includegraphics[width=1.2cm]{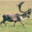}
\end{minipage}
\begin{minipage}[t]{0.08\textwidth}
\centering
\includegraphics[width=1.2cm]{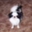}
\end{minipage}
\begin{minipage}[t]{0.08\textwidth}
\centering
\includegraphics[width=1.2cm]{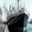}
\end{minipage}
\begin{minipage}[t]{0.08\textwidth}
\centering
\includegraphics[width=1.2cm]{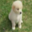}
\end{minipage}
\end{subfigure}
 
  \begin{subfigure}[t]{1\linewidth}
\centering
\begin{minipage}[c]{0.08\textwidth}
\vspace{-2em} \scriptsize Restored\\ Image 
\end{minipage}
\begin{minipage}[t]{0.08\textwidth}
\centering
\includegraphics[width=1.2cm]{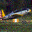}
\end{minipage}
\begin{minipage}[t]{0.08\textwidth}
\centering
\includegraphics[width=1.2cm]{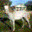}
\end{minipage}
\begin{minipage}[t]{0.08\textwidth}
\centering
\includegraphics[width=1.2cm]{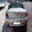}
\end{minipage}
\begin{minipage}[t]{0.08\textwidth}
\centering
\includegraphics[width=1.2cm]{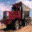}
\end{minipage}
\begin{minipage}[t]{0.08\textwidth}
\centering
\includegraphics[width=1.2cm]{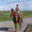}
\end{minipage}
\begin{minipage}[t]{0.08\textwidth}
\centering
\includegraphics[width=1.2cm]{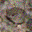}
\end{minipage}
\begin{minipage}[t]{0.08\textwidth}
\centering
\includegraphics[width=1.2cm]{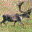}
\end{minipage}
\begin{minipage}[t]{0.08\textwidth}
\centering
\includegraphics[width=1.2cm]{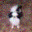}
\end{minipage}
\begin{minipage}[t]{0.08\textwidth}
\centering
\includegraphics[width=1.2cm]{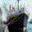}
\end{minipage}
\begin{minipage}[t]{0.08\textwidth}
\centering
\includegraphics[width=1.2cm]{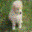}
\end{minipage}
\end{subfigure}

 \begin{subfigure}[t]{1\linewidth}
\centering
\begin{minipage}[c]{0.08\textwidth}
\vspace{-2em} \scriptsize Ground\\ Truth 
\end{minipage}
\begin{minipage}[t]{0.08\textwidth}
\centering
\includegraphics[width=1.2cm]{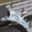}
\end{minipage}
\begin{minipage}[t]{0.08\textwidth}
\centering
\includegraphics[width=1.2cm]{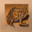}
\end{minipage}
\begin{minipage}[t]{0.08\textwidth}
\centering
\includegraphics[width=1.2cm]{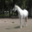}
\end{minipage}
\begin{minipage}[t]{0.08\textwidth}
\centering
\includegraphics[width=1.2cm]{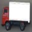}
\end{minipage}
\begin{minipage}[t]{0.08\textwidth}
\centering
\includegraphics[width=1.2cm]{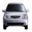}
\end{minipage}
\begin{minipage}[t]{0.08\textwidth}
\centering
\includegraphics[width=1.2cm]{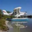}
\end{minipage}
\begin{minipage}[t]{0.08\textwidth}
\centering
\includegraphics[width=1.2cm]{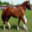}
\end{minipage}
\begin{minipage}[t]{0.08\textwidth}
\centering
\includegraphics[width=1.2cm]{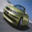}
\end{minipage}
\begin{minipage}[t]{0.08\textwidth}
\centering
\includegraphics[width=1.2cm]{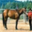}
\end{minipage}
\begin{minipage}[t]{0.08\textwidth}
\centering
\includegraphics[width=1.2cm]{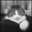}
\end{minipage}
\end{subfigure}
 
  \begin{subfigure}[t]{1\linewidth}
\centering
\begin{minipage}[c]{0.08\textwidth}
\vspace{-2em} \scriptsize Restored\\ Image 
\end{minipage}
\begin{minipage}[t]{0.08\textwidth}
\centering
\includegraphics[width=1.2cm]{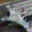}
\end{minipage}
\begin{minipage}[t]{0.08\textwidth}
\centering
\includegraphics[width=1.2cm]{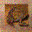}
\end{minipage}
\begin{minipage}[t]{0.08\textwidth}
\centering
\includegraphics[width=1.2cm]{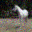}
\end{minipage}
\begin{minipage}[t]{0.08\textwidth}
\centering
\includegraphics[width=1.2cm]{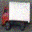}
\end{minipage}
\begin{minipage}[t]{0.08\textwidth}
\centering
\includegraphics[width=1.2cm]{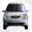}
\end{minipage}
\begin{minipage}[t]{0.08\textwidth}
\centering
\includegraphics[width=1.2cm]{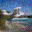}
\end{minipage}
\begin{minipage}[t]{0.08\textwidth}
\centering
\includegraphics[width=1.2cm]{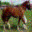}
\end{minipage}
\begin{minipage}[t]{0.08\textwidth}
\centering
\includegraphics[width=1.2cm]{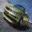}
\end{minipage}
\begin{minipage}[t]{0.08\textwidth}
\centering
\includegraphics[width=1.2cm]{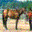}
\end{minipage}
\begin{minipage}[t]{0.08\textwidth}
\centering
\includegraphics[width=1.2cm]{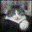}
\end{minipage}
\end{subfigure}

 \begin{subfigure}[t]{1\linewidth}
\centering
\begin{minipage}[c]{0.08\textwidth}
\vspace{-2em} \scriptsize Ground\\ Truth 
\end{minipage}
\begin{minipage}[t]{0.08\textwidth}
\centering
\includegraphics[width=1.2cm]{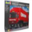}
\end{minipage}
\begin{minipage}[t]{0.08\textwidth}
\centering
\includegraphics[width=1.2cm]{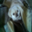}
\end{minipage}
\begin{minipage}[t]{0.08\textwidth}
\centering
\includegraphics[width=1.2cm]{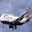}
\end{minipage}
\begin{minipage}[t]{0.08\textwidth}
\centering
\includegraphics[width=1.2cm]{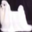}
\end{minipage}
\begin{minipage}[t]{0.08\textwidth}
\centering
\includegraphics[width=1.2cm]{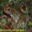}
\end{minipage}
\begin{minipage}[t]{0.08\textwidth}
\centering
\includegraphics[width=1.2cm]{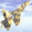}
\end{minipage}
\begin{minipage}[t]{0.08\textwidth}
\centering
\includegraphics[width=1.2cm]{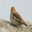}
\end{minipage}
\begin{minipage}[t]{0.08\textwidth}
\centering
\includegraphics[width=1.2cm]{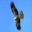}
\end{minipage}
\begin{minipage}[t]{0.08\textwidth}
\centering
\includegraphics[width=1.2cm]{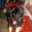}
\end{minipage}
\begin{minipage}[t]{0.08\textwidth}
\centering
\includegraphics[width=1.2cm]{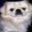}
\end{minipage}
\end{subfigure}
 
  \begin{subfigure}[t]{1\linewidth}
\centering
\begin{minipage}[c]{0.08\textwidth}
\vspace{-2em} \scriptsize Restored\\ Image 
\end{minipage}
\begin{minipage}[t]{0.08\textwidth}
\centering
\includegraphics[width=1.2cm]{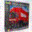}
\end{minipage}
\begin{minipage}[t]{0.08\textwidth}
\centering
\includegraphics[width=1.2cm]{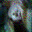}
\end{minipage}
\begin{minipage}[t]{0.08\textwidth}
\centering
\includegraphics[width=1.2cm]{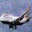}
\end{minipage}
\begin{minipage}[t]{0.08\textwidth}
\centering
\includegraphics[width=1.2cm]{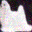}
\end{minipage}
\begin{minipage}[t]{0.08\textwidth}
\centering
\includegraphics[width=1.2cm]{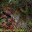}
\end{minipage}
\begin{minipage}[t]{0.08\textwidth}
\centering
\includegraphics[width=1.2cm]{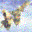}
\end{minipage}
\begin{minipage}[t]{0.08\textwidth}
\centering
\includegraphics[width=1.2cm]{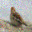}
\end{minipage}
\begin{minipage}[t]{0.08\textwidth}
\centering
\includegraphics[width=1.2cm]{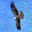}
\end{minipage}
\begin{minipage}[t]{0.08\textwidth}
\centering
\includegraphics[width=1.2cm]{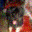}
\end{minipage}
\begin{minipage}[t]{0.08\textwidth}
\centering
\includegraphics[width=1.2cm]{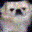}
\end{minipage}
\end{subfigure}

 \begin{subfigure}[t]{1\linewidth}
\centering
\begin{minipage}[c]{0.08\textwidth}
\vspace{-2em} \scriptsize Ground\\ Truth 
\end{minipage}
\begin{minipage}[t]{0.08\textwidth}
\centering
\includegraphics[width=1.2cm]{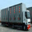}
\end{minipage}
\begin{minipage}[t]{0.08\textwidth}
\centering
\includegraphics[width=1.2cm]{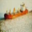}
\end{minipage}
\begin{minipage}[t]{0.08\textwidth}
\centering
\includegraphics[width=1.2cm]{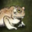}
\end{minipage}
\begin{minipage}[t]{0.08\textwidth}
\centering
\includegraphics[width=1.2cm]{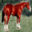}
\end{minipage}
\begin{minipage}[t]{0.08\textwidth}
\centering
\includegraphics[width=1.2cm]{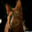}
\end{minipage}
\begin{minipage}[t]{0.08\textwidth}
\centering
\includegraphics[width=1.2cm]{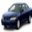}
\end{minipage}
\begin{minipage}[t]{0.08\textwidth}
\centering
\includegraphics[width=1.2cm]{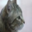}
\end{minipage}
\begin{minipage}[t]{0.08\textwidth}
\centering
\includegraphics[width=1.2cm]{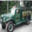}
\end{minipage}
\begin{minipage}[t]{0.08\textwidth}
\centering
\includegraphics[width=1.2cm]{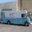}
\end{minipage}
\begin{minipage}[t]{0.08\textwidth}
\centering
\includegraphics[width=1.2cm]{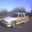}
\end{minipage}
\end{subfigure}
 
  \begin{subfigure}[t]{1\linewidth}
\centering
\begin{minipage}[c]{0.08\textwidth}
\vspace{-2em} \scriptsize Restored\\ Image 
\end{minipage}
\begin{minipage}[t]{0.08\textwidth}
\centering
\includegraphics[width=1.2cm]{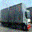}
\end{minipage}
\begin{minipage}[t]{0.08\textwidth}
\centering
\includegraphics[width=1.2cm]{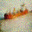}
\end{minipage}
\begin{minipage}[t]{0.08\textwidth}
\centering
\includegraphics[width=1.2cm]{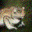}
\end{minipage}
\begin{minipage}[t]{0.08\textwidth}
\centering
\includegraphics[width=1.2cm]{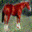}
\end{minipage}
\begin{minipage}[t]{0.08\textwidth}
\centering
\includegraphics[width=1.2cm]{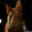}
\end{minipage}
\begin{minipage}[t]{0.08\textwidth}
\centering
\includegraphics[width=1.2cm]{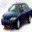}
\end{minipage}
\begin{minipage}[t]{0.08\textwidth}
\centering
\includegraphics[width=1.2cm]{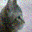}
\end{minipage}
\begin{minipage}[t]{0.08\textwidth}
\centering
\includegraphics[width=1.2cm]{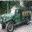}
\end{minipage}
\begin{minipage}[t]{0.08\textwidth}
\centering
\includegraphics[width=1.2cm]{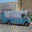}
\end{minipage}
\begin{minipage}[t]{0.08\textwidth}
\centering
\includegraphics[width=1.2cm]{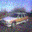}
\end{minipage}
\end{subfigure}

  \caption{ Visual examples of TGIAs-RO (our method with 10 temporal gradients) on CIFAR10  dataset (batch size = 32, image size $= 3 \times 32 \times 32$).} 
  \label{visual}
  
    \vspace{-5pt}
\end{figure*}

\begin{figure*}[htbp]
  \centering
 \vspace{-10pt}
 \begin{subfigure}[t]{1\linewidth}
\centering
\begin{minipage}[c]{0.08\textwidth}
\vspace{-2em} \scriptsize Ground\\ Truth 
\end{minipage}
\begin{minipage}[t]{0.08\textwidth}
\centering
\includegraphics[width=1.2cm]{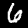}
\end{minipage}
\begin{minipage}[t]{0.08\textwidth}
\centering
\includegraphics[width=1.2cm]{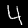}
\end{minipage}
\begin{minipage}[t]{0.08\textwidth}
\centering
\includegraphics[width=1.2cm]{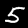}
\end{minipage}
\begin{minipage}[t]{0.08\textwidth}
\centering
\includegraphics[width=1.2cm]{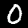}
\end{minipage}
\begin{minipage}[t]{0.08\textwidth}
\centering
\includegraphics[width=1.2cm]{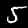}
\end{minipage}
\begin{minipage}[t]{0.08\textwidth}
\centering
\includegraphics[width=1.2cm]{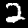}
\end{minipage}
\begin{minipage}[t]{0.08\textwidth}
\centering
\includegraphics[width=1.2cm]{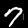}
\end{minipage}
\begin{minipage}[t]{0.08\textwidth}
\centering
\includegraphics[width=1.2cm]{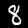}
\end{minipage}
\begin{minipage}[t]{0.08\textwidth}
\centering
\includegraphics[width=1.2cm]{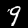}
\end{minipage}
\begin{minipage}[t]{0.08\textwidth}
\centering
\includegraphics[width=1.2cm]{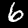}
\end{minipage}
\end{subfigure}
 
  \begin{subfigure}[t]{1\linewidth}
\centering
\begin{minipage}[c]{0.08\textwidth}
\vspace{-2em} \scriptsize Restored\\ Image 
\end{minipage}
\begin{minipage}[t]{0.08\textwidth}
\centering
\includegraphics[width=1.2cm]{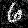}
\end{minipage}
\begin{minipage}[t]{0.08\textwidth}
\centering
\includegraphics[width=1.2cm]{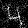}
\end{minipage}
\begin{minipage}[t]{0.08\textwidth}
\centering
\includegraphics[width=1.2cm]{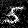}
\end{minipage}
\begin{minipage}[t]{0.08\textwidth}
\centering
\includegraphics[width=1.2cm]{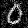}
\end{minipage}
\begin{minipage}[t]{0.08\textwidth}
\centering
\includegraphics[width=1.2cm]{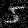}
\end{minipage}
\begin{minipage}[t]{0.08\textwidth}
\centering
\includegraphics[width=1.2cm]{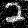}
\end{minipage}
\begin{minipage}[t]{0.08\textwidth}
\centering
\includegraphics[width=1.2cm]{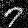}
\end{minipage}
\begin{minipage}[t]{0.08\textwidth}
\centering
\includegraphics[width=1.2cm]{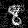}
\end{minipage}
\begin{minipage}[t]{0.08\textwidth}
\centering
\includegraphics[width=1.2cm]{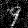}
\end{minipage}
\begin{minipage}[t]{0.08\textwidth}
\centering
\includegraphics[width=1.2cm]{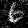}
\end{minipage}
\end{subfigure}

 \begin{subfigure}[t]{1\linewidth}
\centering
\begin{minipage}[c]{0.08\textwidth}
\vspace{-2em} \scriptsize Ground\\ Truth 
\end{minipage}
\begin{minipage}[t]{0.08\textwidth}
\centering
\includegraphics[width=1.2cm]{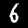}
\end{minipage}
\begin{minipage}[t]{0.08\textwidth}
\centering
\includegraphics[width=1.2cm]{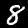}
\end{minipage}
\begin{minipage}[t]{0.08\textwidth}
\centering
\includegraphics[width=1.2cm]{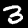}
\end{minipage}
\begin{minipage}[t]{0.08\textwidth}
\centering
\includegraphics[width=1.2cm]{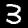}
\end{minipage}
\begin{minipage}[t]{0.08\textwidth}
\centering
\includegraphics[width=1.2cm]{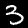}
\end{minipage}
\begin{minipage}[t]{0.08\textwidth}
\centering
\includegraphics[width=1.2cm]{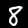}
\end{minipage}
\begin{minipage}[t]{0.08\textwidth}
\centering
\includegraphics[width=1.2cm]{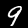}
\end{minipage}
\begin{minipage}[t]{0.08\textwidth}
\centering
\includegraphics[width=1.2cm]{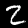}
\end{minipage}
\begin{minipage}[t]{0.08\textwidth}
\centering
\includegraphics[width=1.2cm]{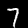}
\end{minipage}
\begin{minipage}[t]{0.08\textwidth}
\centering
\includegraphics[width=1.2cm]{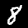}
\end{minipage}
\end{subfigure}1
 
  \begin{subfigure}[t]{1\linewidth}
\centering
\begin{minipage}[c]{0.08\textwidth}
\vspace{-2em} 
\scriptsize Restored\\ Image 
\end{minipage}
\begin{minipage}[t]{0.08\textwidth}
\centering
\includegraphics[width=1.2cm]{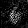}
\end{minipage}
\begin{minipage}[t]{0.08\textwidth}
\centering
\includegraphics[width=1.2cm]{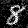}
\end{minipage}
\begin{minipage}[t]{0.08\textwidth}
\centering
\includegraphics[width=1.2cm]{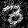}
\end{minipage}
\begin{minipage}[t]{0.08\textwidth}
\centering
\includegraphics[width=1.2cm]{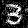}
\end{minipage}
\begin{minipage}[t]{0.08\textwidth}
\centering
\includegraphics[width=1.2cm]{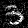}
\end{minipage}
\begin{minipage}[t]{0.08\textwidth}
\centering
\includegraphics[width=1.2cm]{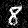}
\end{minipage}
\begin{minipage}[t]{0.08\textwidth}
\centering
\includegraphics[width=1.2cm]{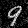}
\end{minipage}
\begin{minipage}[t]{0.08\textwidth}
\centering
\includegraphics[width=1.2cm]{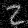}
\end{minipage}
\begin{minipage}[t]{0.08\textwidth}
\centering
\includegraphics[width=1.2cm]{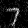}
\end{minipage}
\begin{minipage}[t]{0.08\textwidth}
\centering
\includegraphics[width=1.2cm]{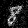}
\end{minipage}
\end{subfigure}

 \begin{subfigure}[t]{1\linewidth}
\centering
\begin{minipage}[c]{0.08\textwidth}
\vspace{-2em} \scriptsize Ground\\ Truth 
\end{minipage}
\begin{minipage}[t]{0.08\textwidth}
\centering
\includegraphics[width=1.2cm]{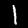}
\end{minipage}
\begin{minipage}[t]{0.08\textwidth}
\centering
\includegraphics[width=1.2cm]{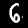}
\end{minipage}
\begin{minipage}[t]{0.08\textwidth}
\centering
\includegraphics[width=1.2cm]{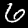}
\end{minipage}
\begin{minipage}[t]{0.08\textwidth}
\centering
\includegraphics[width=1.2cm]{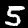}
\end{minipage}
\begin{minipage}[t]{0.08\textwidth}
\centering
\includegraphics[width=1.2cm]{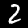}
\end{minipage}
\begin{minipage}[t]{0.08\textwidth}
\centering
\includegraphics[width=1.2cm]{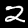}
\end{minipage}
\begin{minipage}[t]{0.08\textwidth}
\centering
\includegraphics[width=1.2cm]{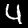}
\end{minipage}
\begin{minipage}[t]{0.08\textwidth}
\centering
\includegraphics[width=1.2cm]{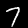}
\end{minipage}
\begin{minipage}[t]{0.08\textwidth}
\centering
\includegraphics[width=1.2cm]{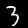}
\end{minipage}
\begin{minipage}[t]{0.08\textwidth}
\centering
\includegraphics[width=1.2cm]{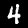}
\end{minipage}
\end{subfigure}
 
  \begin{subfigure}[t]{1\linewidth}
\centering
\begin{minipage}[c]{0.08\textwidth}
\vspace{-2em} \scriptsize Restored\\ Image 
\end{minipage}
\begin{minipage}[t]{0.08\textwidth}
\centering
\includegraphics[width=1.2cm]{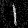}
\end{minipage}
\begin{minipage}[t]{0.08\textwidth}
\centering
\includegraphics[width=1.2cm]{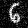}
\end{minipage}
\begin{minipage}[t]{0.08\textwidth}
\centering
\includegraphics[width=1.2cm]{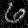}
\end{minipage}
\begin{minipage}[t]{0.08\textwidth}
\centering
\includegraphics[width=1.2cm]{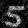}
\end{minipage}
\begin{minipage}[t]{0.08\textwidth}
\centering
\includegraphics[width=1.2cm]{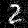}
\end{minipage}
\begin{minipage}[t]{0.08\textwidth}
\centering
\includegraphics[width=1.2cm]{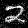}
\end{minipage}
\begin{minipage}[t]{0.08\textwidth}
\centering
\includegraphics[width=1.2cm]{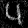}
\end{minipage}
\begin{minipage}[t]{0.08\textwidth}
\centering
\includegraphics[width=1.2cm]{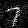}
\end{minipage}
\begin{minipage}[t]{0.08\textwidth}
\centering
\includegraphics[width=1.2cm]{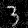}
\end{minipage}
\begin{minipage}[t]{0.08\textwidth}
\centering
\includegraphics[width=1.2cm]{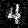}
\end{minipage}
\end{subfigure}

 \begin{subfigure}[t]{1\linewidth}
\centering
\begin{minipage}[c]{0.08\textwidth}
\vspace{-2em} \scriptsize Ground\\ Truth 
\end{minipage}
\begin{minipage}[t]{0.08\textwidth}
\centering
\includegraphics[width=1.2cm]{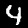}
\end{minipage}
\begin{minipage}[t]{0.08\textwidth}
\centering
\includegraphics[width=1.2cm]{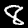}
\end{minipage}
\begin{minipage}[t]{0.08\textwidth}
\centering
\includegraphics[width=1.2cm]{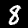}
\end{minipage}
\begin{minipage}[t]{0.08\textwidth}
\centering
\includegraphics[width=1.2cm]{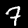}
\end{minipage}
\begin{minipage}[t]{0.08\textwidth}
\centering
\includegraphics[width=1.2cm]{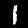}
\end{minipage}
\begin{minipage}[t]{0.08\textwidth}
\centering
\includegraphics[width=1.2cm]{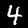}
\end{minipage}
\begin{minipage}[t]{0.08\textwidth}
\centering
\includegraphics[width=1.2cm]{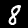}
\end{minipage}
\begin{minipage}[t]{0.08\textwidth}
\centering
\includegraphics[width=1.2cm]{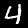}
\end{minipage}
\begin{minipage}[t]{0.08\textwidth}
\centering
\includegraphics[width=1.2cm]{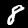}
\end{minipage}
\begin{minipage}[t]{0.08\textwidth}
\centering
\includegraphics[width=1.2cm]{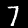}
\end{minipage}
\end{subfigure}
 
  \begin{subfigure}[t]{1\linewidth}
\centering
\begin{minipage}[c]{0.08\textwidth}
\vspace{-2em} \scriptsize Restored\\ Image 
\end{minipage}
\begin{minipage}[t]{0.08\textwidth}
\centering
\includegraphics[width=1.2cm]{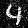}
\end{minipage}
\begin{minipage}[t]{0.08\textwidth}
\centering
\includegraphics[width=1.2cm]{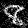}
\end{minipage}
\begin{minipage}[t]{0.08\textwidth}
\centering
\includegraphics[width=1.2cm]{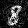}
\end{minipage}
\begin{minipage}[t]{0.08\textwidth}
\centering
\includegraphics[width=1.2cm]{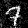}
\end{minipage}
\begin{minipage}[t]{0.08\textwidth}
\centering
\includegraphics[width=1.2cm]{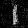}
\end{minipage}
\begin{minipage}[t]{0.08\textwidth}
\centering
\includegraphics[width=1.2cm]{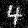}
\end{minipage}
\begin{minipage}[t]{0.08\textwidth}
\centering
\includegraphics[width=1.2cm]{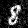}
\end{minipage}
\begin{minipage}[t]{0.08\textwidth}
\centering
\includegraphics[width=1.2cm]{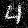}
\end{minipage}
\begin{minipage}[t]{0.08\textwidth}
\centering
\includegraphics[width=1.2cm]{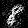}
\end{minipage}
\begin{minipage}[t]{0.08\textwidth}
\centering
\includegraphics[width=1.2cm]{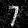}
\end{minipage}
\end{subfigure}

 \begin{subfigure}[t]{1\linewidth}
\centering
\begin{minipage}[c]{0.08\textwidth}
\vspace{-2em} \scriptsize Ground\\ Truth 
\end{minipage}
\begin{minipage}[t]{0.08\textwidth}
\centering
\includegraphics[width=1.2cm]{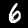}
\end{minipage}
\begin{minipage}[t]{0.08\textwidth}
\centering
\includegraphics[width=1.2cm]{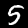}
\end{minipage}
\begin{minipage}[t]{0.08\textwidth}
\centering
\includegraphics[width=1.2cm]{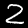}
\end{minipage}
\begin{minipage}[t]{0.08\textwidth}
\centering
\includegraphics[width=1.2cm]{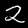}
\end{minipage}
\begin{minipage}[t]{0.08\textwidth}
\centering
\includegraphics[width=1.2cm]{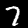}
\end{minipage}
\begin{minipage}[t]{0.08\textwidth}
\centering
\includegraphics[width=1.2cm]{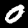}
\end{minipage}
\begin{minipage}[t]{0.08\textwidth}
\centering
\includegraphics[width=1.2cm]{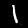}
\end{minipage}
\begin{minipage}[t]{0.08\textwidth}
\centering
\includegraphics[width=1.2cm]{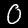}
\end{minipage}
\begin{minipage}[t]{0.08\textwidth}
\centering
\includegraphics[width=1.2cm]{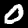}
\end{minipage}
\begin{minipage}[t]{0.08\textwidth}
\centering
\includegraphics[width=1.2cm]{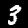}
\end{minipage}
\end{subfigure}
 
  \begin{subfigure}[t]{1\linewidth}
\centering
\begin{minipage}[c]{0.08\textwidth}
\vspace{-2em} \scriptsize Restored\\ Image 
\end{minipage}
\begin{minipage}[t]{0.08\textwidth}
\centering
\includegraphics[width=1.2cm]{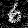}
\end{minipage}
\begin{minipage}[t]{0.08\textwidth}
\centering
\includegraphics[width=1.2cm]{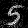}
\end{minipage}
\begin{minipage}[t]{0.08\textwidth}
\centering
\includegraphics[width=1.2cm]{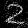}
\end{minipage}
\begin{minipage}[t]{0.08\textwidth}
\centering
\includegraphics[width=1.2cm]{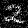}
\end{minipage}
\begin{minipage}[t]{0.08\textwidth}
\centering
\includegraphics[width=1.2cm]{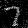}
\end{minipage}
\begin{minipage}[t]{0.08\textwidth}
\centering
\includegraphics[width=1.2cm]{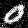}
\end{minipage}
\begin{minipage}[t]{0.08\textwidth}
\centering
\includegraphics[width=1.2cm]{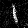}
\end{minipage}
\begin{minipage}[t]{0.08\textwidth}
\centering
\includegraphics[width=1.2cm]{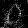}
\end{minipage}
\begin{minipage}[t]{0.08\textwidth}
\centering
\includegraphics[width=1.2cm]{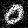}
\end{minipage}
\begin{minipage}[t]{0.08\textwidth}
\centering
\includegraphics[width=1.2cm]{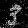}
\end{minipage}
\end{subfigure}

 \begin{subfigure}[t]{1\linewidth}
\centering
\begin{minipage}[c]{0.08\textwidth}
\vspace{-2em} \scriptsize Ground\\ Truth 
\end{minipage}
\begin{minipage}[t]{0.08\textwidth}
\centering
\includegraphics[width=1.2cm]{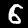}
\end{minipage}
\begin{minipage}[t]{0.08\textwidth}
\centering
\includegraphics[width=1.2cm]{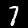}
\end{minipage}
\begin{minipage}[t]{0.08\textwidth}
\centering
\includegraphics[width=1.2cm]{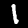}
\end{minipage}
\begin{minipage}[t]{0.08\textwidth}
\centering
\includegraphics[width=1.2cm]{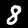}
\end{minipage}
\begin{minipage}[t]{0.08\textwidth}
\centering
\includegraphics[width=1.2cm]{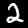}
\end{minipage}
\begin{minipage}[t]{0.08\textwidth}
\centering
\includegraphics[width=1.2cm]{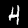}
\end{minipage}
\begin{minipage}[t]{0.08\textwidth}
\centering
\includegraphics[width=1.2cm]{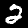}
\end{minipage}
\begin{minipage}[t]{0.08\textwidth}
\centering
\includegraphics[width=1.2cm]{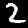}
\end{minipage}
\begin{minipage}[t]{0.08\textwidth}
\centering
\includegraphics[width=1.2cm]{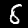}
\end{minipage}
\begin{minipage}[t]{0.08\textwidth}
\centering
\includegraphics[width=1.2cm]{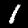}
\end{minipage}
\end{subfigure}
 
  \begin{subfigure}[t]{1\linewidth}
\centering
\begin{minipage}[c]{0.08\textwidth}
\vspace{-2em} \scriptsize Restored\\ Image 
\end{minipage}
\begin{minipage}[t]{0.08\textwidth}
\centering
\includegraphics[width=1.2cm]{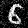}
\end{minipage}
\begin{minipage}[t]{0.08\textwidth}
\centering
\includegraphics[width=1.2cm]{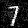}
\end{minipage}
\begin{minipage}[t]{0.08\textwidth}
\centering
\includegraphics[width=1.2cm]{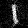}
\end{minipage}
\begin{minipage}[t]{0.08\textwidth}
\centering
\includegraphics[width=1.2cm]{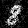}
\end{minipage}
\begin{minipage}[t]{0.08\textwidth}
\centering
\includegraphics[width=1.2cm]{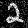}
\end{minipage}
\begin{minipage}[t]{0.08\textwidth}
\centering
\includegraphics[width=1.2cm]{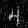}
\end{minipage}
\begin{minipage}[t]{0.08\textwidth}
\centering
\includegraphics[width=1.2cm]{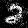}
\end{minipage}
\begin{minipage}[t]{0.08\textwidth}
\centering
\includegraphics[width=1.2cm]{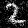}
\end{minipage}
\begin{minipage}[t]{0.08\textwidth}
\centering
\includegraphics[width=1.2cm]{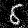}
\end{minipage}
\begin{minipage}[t]{0.08\textwidth}
\centering
\includegraphics[width=1.2cm]{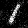}
\end{minipage}
\end{subfigure}

 \begin{subfigure}[t]{1\linewidth}
\centering
\begin{minipage}[c]{0.08\textwidth}
\vspace{-2em} \scriptsize Ground\\ Truth 
\end{minipage}
\begin{minipage}[t]{0.08\textwidth}
\centering
\includegraphics[width=1.2cm]{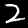}
\end{minipage}
\begin{minipage}[t]{0.08\textwidth}
\centering
\includegraphics[width=1.2cm]{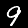}
\end{minipage}
\begin{minipage}[t]{0.08\textwidth}
\centering
\includegraphics[width=1.2cm]{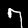}
\end{minipage}
\begin{minipage}[t]{0.08\textwidth}
\centering
\includegraphics[width=1.2cm]{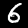}
\end{minipage}
\begin{minipage}[t]{0.08\textwidth}
\centering
\includegraphics[width=1.2cm]{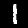}
\end{minipage}
\begin{minipage}[t]{0.08\textwidth}
\centering
\includegraphics[width=1.2cm]{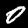}
\end{minipage}
\begin{minipage}[t]{0.08\textwidth}
\centering
\includegraphics[width=1.2cm]{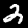}
\end{minipage}
\begin{minipage}[t]{0.08\textwidth}
\centering
\includegraphics[width=1.2cm]{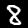}
\end{minipage}
\begin{minipage}[t]{0.08\textwidth}
\centering
\includegraphics[width=1.2cm]{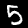}
\end{minipage}
\begin{minipage}[t]{0.08\textwidth}
\centering
\includegraphics[width=1.2cm]{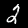}
\end{minipage}
\end{subfigure}
 
  \begin{subfigure}[t]{1\linewidth}
\centering
\begin{minipage}[c]{0.08\textwidth}
\vspace{-2em} \scriptsize Restored\\ Image 
\end{minipage}
\begin{minipage}[t]{0.08\textwidth}
\centering
\includegraphics[width=1.2cm]{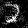}
\end{minipage}
\begin{minipage}[t]{0.08\textwidth}
\centering
\includegraphics[width=1.2cm]{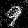}
\end{minipage}
\begin{minipage}[t]{0.08\textwidth}
\centering
\includegraphics[width=1.2cm]{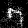}
\end{minipage}
\begin{minipage}[t]{0.08\textwidth}
\centering
\includegraphics[width=1.2cm]{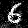}
\end{minipage}
\begin{minipage}[t]{0.08\textwidth}
\centering
\includegraphics[width=1.2cm]{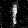}
\end{minipage}
\begin{minipage}[t]{0.08\textwidth}
\centering
\includegraphics[width=1.2cm]{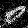}
\end{minipage}
\begin{minipage}[t]{0.08\textwidth}
\centering
\includegraphics[width=1.2cm]{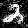}
\end{minipage}
\begin{minipage}[t]{0.08\textwidth}
\centering
\includegraphics[width=1.2cm]{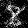}
\end{minipage}
\begin{minipage}[t]{0.08\textwidth}
\centering
\includegraphics[width=1.2cm]{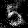}
\end{minipage}
\begin{minipage}[t]{0.08\textwidth}
\centering
\includegraphics[width=1.2cm]{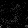}
\end{minipage}
\end{subfigure}

  \caption{ Visual examples of TGIAs-RO (our method with 10 temporal gradients) on MNIST dataset (batch size = 32, image size $28 \times 28$).} 
  \label{visual_mnist}
  
    \vspace{-5pt}
\end{figure*}

Fig. \ref{fig: bar_comparsion} illustrates the best cases and the average cases of SSIM and PSNR comparisons between proposed TGIAs-RO and the state-of-the-art methods. And Tab. \ref{tab: comparison} provides the quantitative results of methods on varying model architectures and datasets. Moreover, We provide some visual results of MNIST, CIFAR10 and ImageNet images in Fig. \ref{visual_mnist}, Fig. \ref{visual1-app} and Fig. \ref{visual}. 

\subsubsection{TGIAs-GO on Text Classification Tasks}
We also conduct experiments on text-processing tasks. Specifically, we choose a text classification task with a TextCNN with 3 convolution layers \cite{zhang2015sensitivity}. The R8 dataset (full term version) is a subset of the Reuters 21578 datasets, which has 8 categories divided into 5,485 training and 2,189 test documents.  

Different from image restoration tasks where the inputs are continuous values, language models need to preprocess discrete words into embedding (continuous values). We apply TGIAs-RO on embedding space and minimize the gradient distance between dummy embedding and ground truth ones. After optimizing the dummy embedding, we derive original words by finding the closest index entries in the embedding matrix reversely.

We reconstruct the text in Reuters R8 with two different TextCNN model architectures. One with 1 convolution layer and the other one with 3 convolution layers, each convolution layer is followed by a sigmoid function and a max pooling layer. We adopt \textit{recover rate} \cite{deng2021tag} to measure the text restoration performance, which is defined as the maximum percentage of tokens in ground truth recovered by TGIAs-RO. Results in Tab. \ref{tab: NLP} indicate that TGIAs-RO with 10 temporal gradients reconstruct text embedding better than those with fewer gradients. Moreover, texts in a larger batch size are more difficult to reconstruct, and more complex model architectures increase the failures in GIAs as Tab. \ref{tab: NLP} shows. 

\begin{table}[H]
\caption{Recover rate of TGIAs-RO on text classification tasks, the model architecture adopted is TextCNN and dataset is R8 dataset from Reuters 21587 datasets (sentence length = 30).}\label{tab: NLP}

 \renewcommand{\arraystretch}{0.93}
 \centering
\resizebox{0.48\textwidth}{!}{
\begin{tabular}{cccc}
\toprule 
\multirow{2}{*}{ Iteration  Number } & \multirow{2}{*}{ Batch Size } &\multicolumn{2}{c}{Convolution Block Number} 
\\ \cmidrule(r){3-4} 
& &  1 & 3      \\ \midrule
\multirow{3}{*}{ $T=1$ (DLG) }
&2  & 100\%  &  93.75\%  \\
&4    & 93.75\%  & 70.3\% \\ 
&8    & 78.12\%   & 31.25\%   \\ \hline
\multirow{3}{*}{ $T=5$ }&2   & 100\%    & 98.4\% \\
&4   & 97.65\% & 79.7\% \\
&8   & 87.11\% &  41.01\%  \\ \midrule
\multirow{3}{*}{ $T=10$ }&2    & 100\%    & 100\%   \\
& 4  & 100\%  & 87.5\%  \\
&8 & 91.8\%  &  60.94\%  \\
 \bottomrule
\end{tabular}
}

\end{table}

\begin{table*}[htbp]
      \caption{\scriptsize Best case examples of restored texts of TGIAs-RO with different number of temporal gradients on text classification (the model adopted is TextCNN \cite{zhang2015sensitivity} with 1 convolution layer, and the texts are from Reuters R8 dataset with sentence length 32 in batch size of 8), the bold font indicates the error texts.}\label{tab: text_egs}
  \renewcommand{\arraystretch}{0.99}
     \centering
     \resizebox{0.99\textwidth}{!}{
      \begin{tabular}[l]{cp{7.5cm}p{7.5cm}}
        \toprule
        Methods & Original Text  & Restored Text   \\ \midrule
        
 \multirow{4}{*}{ 
 \tabincell{c}{DLG (TGIAs-RO\\with 1 temporal gradients)}}
 & champion products ch approves stock split champion products inc said its board of directors approved a two for one stock split of its common shares for shareholders of record as of april  & champion products ch approves stock split champion products inc said its board of directors approved a two for one stock split of its common shares for shareholders of record as of april  \\
 \cmidrule(r){2-3}
& investment firms cut cyclops cyl stake a group of affiliated new york investment firms said they lowered their stake in cyclops corp to shares or pct of the total outstanding common stock & investment systems cut cyclops cyl stake a systems of affiliated new has bought firms \textbf{of the stock of stake in cyclops corp exchange for} or pct its the \textbf{following} outstanding \textbf{common compensated} \\\cmidrule(r){2-3}
 & circuit systems csyi buys board maker circuit systems inc said it has bought all of the stock of ionic industries inc in exchange for shares of its common following the exchange there & circuit systems cut cyclops board \textbf{stake circuit group inc affiliated it york investment all said they lowered their stake} industries inc in to shares \textbf{shares of its the total the exchange compensate}  \\
\midrule
  \multirow{4}{*}{ 
  \tabincell{c}{TGIAs-RO with \\5 temporal gradients}}  
& champion products ch approves stock split champion products inc said its board of directors approved a two for one stock split of its common shares for shareholders of record as of april  &  champion products ch approves stock split champion products inc said its board of directors approved a two for one stock split of its common shares for shareholders of record as of april \\\cmidrule(r){2-3}
& investment firms cut cyclops cyl stake a group of affiliated new york investment firms said they lowered their stake in cyclops corp to shares or pct of the total outstanding common stock  & investment firms csyi cyclops cyl stake a group of affiliated new york investment firms said they lowered their stake in cyclops corp to shares or pct of the total outstanding \textbf{exchange there} \\
\cmidrule(r){2-3}
& circuit systems csyi buys board maker circuit systems inc said it has bought all of the stock of ionic industries inc in exchange for shares of its common following the exchange there & circuit systems cut buys board maker circuit systems inc said it has bought all of the stock of ionic industries inc in exchange for shares of its common following the \textbf{common stock}  \\
\midrule

\multirow{4}{*}{\tabincell{c}{TGIAs-RO with \\10 temporal gradients}} 
& champion products ch approves stock split champion products inc said its board of directors approved a two for one stock split of its common shares for shareholders of record as of april & champion exchange ch approves stock split champion products inc said its board of directors approved a two for one stock split of its common shares for shareholders of record as of april  \\ 
\cmidrule(r){2-3}
& investment firms cut cyclops cyl stake a group of affiliated new york investment firms said they lowered their stake in cyclops corp to shares or pct of the total outstanding common stock & investment firms cut cyclops cyl stake a system of affiliated new york investment firms said they lowered their stake in cyclops corp to shares or pct of the total outstanding common \textbf{record} \\
\cmidrule(r){2-3}
&circuit systems csyi buys board maker circuit systems inc said it has bought all of the stock of ionic industries inc in exchange for shares of its common following the exchange there &china systems csyi broad buys maker circuit systems inc said it has bought all of the stock of ionic industries inc in exchange for shares of its common following the exchange there\\
      \bottomrule
      \end{tabular}}

\end{table*}

\subsection{TGIAs-RO with More Prior Knowledege}
In TGIAs-RO, each temporal gradients conduct a GIAs optimization $\calL_{grad}$, while the optimization can be enhanced with additional prior knowledge. 
\subsubsection{Batch Normalization Statistics}
\textit{BN regularizer} \cite{yin2021see} adopted  statistics from the batch normalization layers in network to design an auxiliary regularization term $\calL_{BN}$ for better image restoration. 
\subsubsection{Total Variation Regulaization Term} 
\textit{Total variation} is a measure of the complexity of an image with respect to its spatial variation. Total variation can be adopted as prior information to help image reconstruction in the form of a regularization term $\calL_{TV}$. 

The results in Tab. \ref{tab: prior} indicate that total variation regularization term $\calL_{TV}$ improves PSNR by more than 2 dB, while batch normalization regularization term $\calL_{TV}$ slightly improves PSNR within less than 1dB. 
\begin{table}[H]
\caption{Comparisons of reconstruction performance  with different prior knowledge, the model adopted is AlexNet and dataset is CIFAR10 (batch size = 16, image size = 32$\times$32).}\label{tab: prior}
 \renewcommand{\arraystretch}{1}
 \centering
\resizebox{0.48\textwidth}{!}{
\begin{tabular}{cccc}
\toprule
Regularization Terms Adopted & MSE &  PSNR/dB & SSIM     \\ \midrule
$\calL_{grad}$ & 6$e^{-3}$ & 22.32  &  0.86 \\
$\calL_{grad} + \calL_{TV}$  & 3$e^{-3}$  &  24.92  & 0.91  \\ 
$\calL_{grad} + \calL_{BN}$  &4$e^{-3}$ & 23.02 & 0.87 \\
$\calL_{grad} + \calL_{TV} + \calL_{BN}$  &3$e^{-3}$  & 25.47 & 0.92 \\
 \bottomrule
\end{tabular}
}

\end{table}

\subsection{Analysis of Failure Cases}

We repeat the experiments in Sect. \ref{sec:failure} for 20 times, and report the detailed PSNR of these failure cases to validate our empirical findings on the reason why single-temporal fails. 

\subsubsection{Failure Cases 1: Complex Model Architectures} 
We provide Tab. \ref{tab: statistics1} to show the statistics of PSNR of failure cases 1. 
Fig. \ref{fig:heatmap1} shows that the data recovery via GIAs with single-temporal gradients stuck in the bad local minima and restored data is increasingly far away from the ground truth when we add batch normalization and ReLU layers in sequence. 

The statistical results presented in Table. \ref{tab: statistics1} indicate that adding batch normalization layers reduce the PSNR of reconstructed images from 17.67dB to 13.43dB, and ReLU layers as non-convex units damage the images reconstruction quality to 11.51dB. 

\begin{figure} [h!]
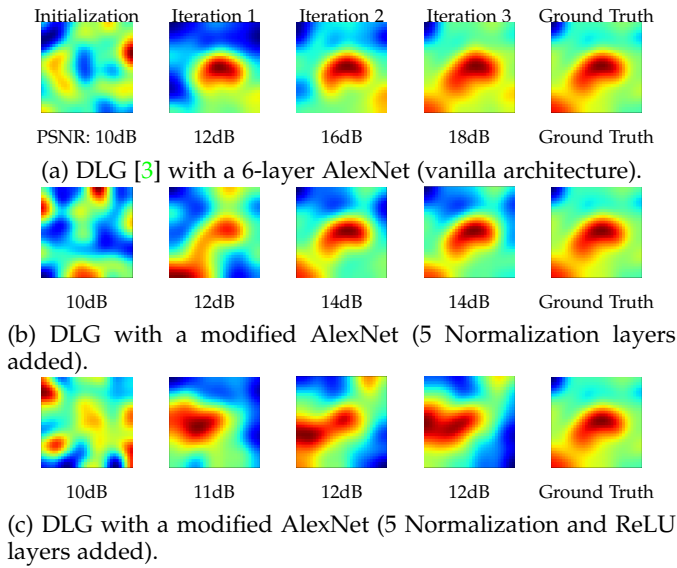

\begin{subfigure}[t]{1\linewidth}
\centering
\begin{minipage}[t]{0.18\textwidth}
\centering
\scriptsize Initialization
\end{minipage}
\begin{minipage}[t]{0.18\textwidth}
\centering
\scriptsize Iteration 1
\end{minipage}
\begin{minipage}[t]{0.18\textwidth}
\centering
\scriptsize Iteration 2
\end{minipage}
\begin{minipage}[t]{0.18\textwidth}
\centering
\scriptsize Iteration 3
\end{minipage}
\begin{minipage}[t]{0.18\textwidth}
\centering
\scriptsize Ground Truth
\end{minipage}
\end{subfigure}

\begin{subfigure}[t]{1\linewidth}
\centering
\begin{minipage}[t]{0.18\textwidth}
\centering
\includegraphics[width=1.2cm]{imgs/heatmaps/1.1.1.png}
\end{minipage}
\begin{minipage}[t]{0.18\textwidth}
\centering
\includegraphics[width=1.2cm]{imgs/heatmaps/1.1.2.png}
\end{minipage}
\begin{minipage}[t]{0.18\textwidth}
\centering
\includegraphics[width=1.2cm]{imgs/heatmaps/1.1.3.png}
\end{minipage}
\begin{minipage}[t]{0.18\textwidth}
\centering
\includegraphics[width=1.2cm]{imgs/heatmaps/1.1.4.png}
\end{minipage}
\begin{minipage}[t]{0.18\textwidth}
\centering
\includegraphics[width=1.2cm]{imgs/heatmaps/gt.png}
\end{minipage}

\centering
\begin{minipage}[t]{0.18\textwidth}
\centering
\scriptsize PSNR: 10dB
\end{minipage}
\begin{minipage}[t]{0.18\textwidth}
\centering
\scriptsize 12dB
\end{minipage}
\begin{minipage}[t]{0.18\textwidth}
\centering
\scriptsize 16dB
\end{minipage}
\begin{minipage}[t]{0.18\textwidth}
\centering
\scriptsize 18dB
\end{minipage}
\begin{minipage}[t]{0.18\textwidth}
\centering
\scriptsize Ground Truth
\end{minipage}
\subcaption{\small DLG \cite{zhu2019deep} with a 6-layer AlexNet (vanilla architecture).} 
\end{subfigure}

\begin{subfigure}[t]{1\linewidth}
\centering
\begin{minipage}[t]{0.18\textwidth}
\centering
\includegraphics[width=1.2cm]{imgs/heatmaps/1.2.1.png}
\end{minipage}
\begin{minipage}[t]{0.18\textwidth}
\centering
\includegraphics[width=1.2cm]{imgs/heatmaps/1.2.2.png}
\end{minipage}
\begin{minipage}[t]{0.18\textwidth}
\centering
\includegraphics[width=1.2cm]{imgs/heatmaps/1.2.3.png}
\end{minipage}
\begin{minipage}[t]{0.18\textwidth}
\centering
\includegraphics[width=1.2cm]{imgs/heatmaps/1.2.4.png}
\end{minipage}
\begin{minipage}[t]{0.18\textwidth}
\centering
\includegraphics[width=1.2cm]{imgs/heatmaps/gt.png}
\end{minipage}
\begin{minipage}[t]{0.18\textwidth}
\centering
\scriptsize 10dB
\end{minipage}
\begin{minipage}[t]{0.18\textwidth}
\centering
\scriptsize 12dB
\end{minipage}
\begin{minipage}[t]{0.18\textwidth}
\centering
\scriptsize 14dB
\end{minipage}
\begin{minipage}[t]{0.18\textwidth}
\centering
\scriptsize 14dB
\end{minipage}
\begin{minipage}[t]{0.18\textwidth}
\centering
\scriptsize Ground Truth 
\end{minipage}
\subcaption{\small DLG with a modified AlexNet (5  Normalization layers added).} 
\end{subfigure}

\begin{subfigure}[t]{1\linewidth}
\centering
\begin{minipage}[t]{0.18\textwidth}
\centering
\includegraphics[width=1.2cm]{imgs/heatmaps/1.3.1.png}
\end{minipage}
\begin{minipage}[t]{0.18\textwidth}
\centering
\includegraphics[width=1.2cm]{imgs/heatmaps/1.3.2.png}
\end{minipage}
\begin{minipage}[t]{0.18\textwidth}
\centering
\includegraphics[width=1.2cm]{imgs/heatmaps/1.3.3.png}
\end{minipage}
\begin{minipage}[t]{0.18\textwidth}
\centering
\includegraphics[width=1.2cm]{imgs/heatmaps/1.3.4.png}
\end{minipage}
\begin{minipage}[t]{0.18\textwidth}
\centering
\includegraphics[width=1.2cm]{imgs/heatmaps/gt.png}
\end{minipage}
\begin{minipage}[t]{0.18\textwidth}
\centering
\scriptsize 10dB
\end{minipage}
\begin{minipage}[t]{0.18\textwidth}
\centering
\scriptsize 11dB
\end{minipage}
\begin{minipage}[t]{0.18\textwidth}
\centering
\scriptsize 12dB
\end{minipage}
\begin{minipage}[t]{0.18\textwidth}
\centering
\scriptsize 12dB
\end{minipage}
\begin{minipage}[t]{0.18\textwidth}
\centering
\scriptsize Ground Truth%
\end{minipage}
\subcaption{\small DLG with a modified AlexNet (5 
 Normalization and ReLU layers added).} 
\end{subfigure}

 \caption{\small The changes of the first channel of restored image in GIAs during optimization with different model architectures. } 
  \label{fig:heatmap3}
  \vspace{-10pt}
\end{figure}

\begin{table}[H]
\caption{Statistics of the PSNR for single-temporal gradients with different model architectures adopted, the dataset is ImageNet (batch size = 8, image size = 224$\times$224).}\label{tab: statistics1}
 \renewcommand{\arraystretch}{1.2}
 \centering
\resizebox{0.48\textwidth}{!}{
\begin{tabular}{ccc}
\toprule
Model Architectures & Repeated Times   & PSNR/dB       \\ \midrule
6-layer AlexNet (Vanilla) & 20 & 17.67 $\pm$ 0.48 dB \\\hline
{\tabincell{c}{Modified AlexNet  \\ (5 batch normalization layers added)}} 
& 20   & 13.43 $\pm$ 0.92 dB   \\ \hline
{\tabincell{c}{Modified AlexNet  \\ (5 batch normalization and ReLU layers added)}} & 20 & 11.51 $\pm$ 0.43 dB   \\
 \bottomrule
\end{tabular}
}

\end{table}

\subsubsection{Failure Cases 2: Invalid Gradient Information} 
\begin{figure} [h!]
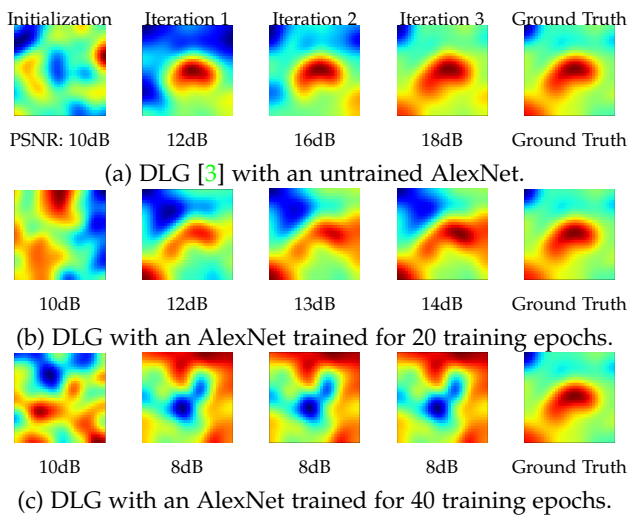

\begin{subfigure}[t]{1\linewidth}
\centering
\begin{minipage}[t]{0.18\textwidth}
\centering
\scriptsize Initialization
\end{minipage}
\begin{minipage}[t]{0.18\textwidth}
\centering
\scriptsize Iteration 1
\end{minipage}
\begin{minipage}[t]{0.18\textwidth}
\centering
\scriptsize Iteration 2
\end{minipage}
\begin{minipage}[t]{0.18\textwidth}
\centering
\scriptsize Iteration 3
\end{minipage}
\begin{minipage}[t]{0.18\textwidth}
\centering
\scriptsize Ground Truth
\end{minipage}
\end{subfigure}

\begin{subfigure}[t]{1\linewidth}
\centering
\begin{minipage}[t]{0.18\textwidth}
\centering
\includegraphics[width=1.2cm]{imgs/heatmaps/1.1.1.png}
\end{minipage}
\begin{minipage}[t]{0.18\textwidth}
\centering
\includegraphics[width=1.2cm]{imgs/heatmaps/1.1.2.png}
\end{minipage}
\begin{minipage}[t]{0.18\textwidth}
\centering
\includegraphics[width=1.2cm]{imgs/heatmaps/1.1.3.png}
\end{minipage}
\begin{minipage}[t]{0.18\textwidth}
\centering
\includegraphics[width=1.2cm]{imgs/heatmaps/1.1.4.png}
\end{minipage}
\begin{minipage}[t]{0.18\textwidth}
\centering
\includegraphics[width=1.2cm]{imgs/heatmaps/gt.png}
\end{minipage}
\begin{minipage}[t]{0.18\textwidth}
\centering
\scriptsize PSNR: 10dB
\end{minipage}
\begin{minipage}[t]{0.18\textwidth}
\centering
\scriptsize 12dB
\end{minipage}
\begin{minipage}[t]{0.18\textwidth}
\centering
\scriptsize 16dB
\end{minipage}
\begin{minipage}[t]{0.18\textwidth}
\centering
\scriptsize 18dB
\end{minipage}
\begin{minipage}[t]{0.18\textwidth}
\centering
\scriptsize Ground Truth
\end{minipage}
\subcaption{\small DLG \cite{zhu2019deep} with an untrained AlexNet.} 
\end{subfigure}

\begin{subfigure}[t]{1\linewidth}
\centering
\begin{minipage}[t]{0.18\textwidth}
\centering
\includegraphics[width=1.2cm]{imgs/heatmaps/2.2.1.png}
\end{minipage}
\begin{minipage}[t]{0.18\textwidth}
\centering
\includegraphics[width=1.2cm]{imgs/heatmaps/2.2.2.png}
\end{minipage}
\begin{minipage}[t]{0.18\textwidth}
\centering
\includegraphics[width=1.2cm]{imgs/heatmaps/2.2.3.png}
\end{minipage}
\begin{minipage}[t]{0.18\textwidth}
\centering
\includegraphics[width=1.2cm]{imgs/heatmaps/2.2.4.png}
\end{minipage}
\begin{minipage}[t]{0.18\textwidth}
\centering
\includegraphics[width=1.2cm]{imgs/heatmaps/gt.png}
\end{minipage}
\begin{minipage}[t]{0.18\textwidth}
\centering
\scriptsize 10dB
\end{minipage}
\begin{minipage}[t]{0.18\textwidth}
\centering
\scriptsize 12dB
\end{minipage}
\begin{minipage}[t]{0.18\textwidth}
\centering
\scriptsize 13dB
\end{minipage}
\begin{minipage}[t]{0.18\textwidth}
\centering
\scriptsize 14dB
\end{minipage}
\begin{minipage}[t]{0.18\textwidth}
\centering
\scriptsize Ground Truth
\end{minipage}
\subcaption{\small DLG with an AlexNet trained for 20 training epochs.} 
\end{subfigure}

\begin{subfigure}[t]{1\linewidth}
\centering
\begin{minipage}[t]{0.18\textwidth}
\centering
\includegraphics[width=1.2cm]{imgs/heatmaps/2.3.1.png}
\end{minipage}
\begin{minipage}[t]{0.18\textwidth}
\centering
\includegraphics[width=1.2cm]{imgs/heatmaps/2.3.2.png}
\end{minipage}
\begin{minipage}[t]{0.18\textwidth}
\centering
\includegraphics[width=1.2cm]{imgs/heatmaps/2.3.3.png}
\end{minipage}
\begin{minipage}[t]{0.18\textwidth}
\centering
\includegraphics[width=1.2cm]{imgs/heatmaps/2.3.4.png}
\end{minipage}
\begin{minipage}[t]{0.18\textwidth}
\centering
\includegraphics[width=1.2cm]{imgs/heatmaps/gt.png}
\end{minipage}
\begin{minipage}[t]{0.18\textwidth}
\centering
\scriptsize 10dB
\end{minipage}
\begin{minipage}[t]{0.18\textwidth}
\centering
\scriptsize 8dB
\end{minipage}
\begin{minipage}[t]{0.18\textwidth}
\centering
\scriptsize 8dB
\end{minipage}
\begin{minipage}[t]{0.18\textwidth}
\centering
\scriptsize 8dB
\end{minipage}
\begin{minipage}[t]{0.18\textwidth}
\centering
\scriptsize Ground Truth
\end{minipage}
\subcaption{\small DLG with an AlexNet trained for 40 training epochs.} 

\end{subfigure}

 \caption{\small The changes of the first channel of restored image in GIAs during optimization with models trained in different epochs. } 
  \label{fig:heatmap2-app}
\end{figure}

We provide Tab. \ref{tab: statistics2} to show the statistics of PSNR of failure cases 2.   When the complex model is trained gradually, Fig. \ref{fig:heatmap2-app} shows that the restored data is collapsed seriously (PSNR is just 8dB), which is far away from ground truth, we ascribe the failures to the invalid model gradients trained gradually.

\begin{table}[H]
 \renewcommand{\arraystretch}{1.1}
 \centering
\resizebox{0.48\textwidth}{!}{
\begin{tabular}{ccc}
\toprule
Model Architectures & Repeated Times   & PSNR/dB       \\ \midrule
6-layer untrained AlexNet & 20 & 17.67 $\pm$ 0.48 dB \\\hline
{\tabincell{c}{trained AlexNet  \\ (trained for 20 epochs)}} 
& 20   & 13.23 $\pm$ 0.42 dB   \\ \hline
{\tabincell{c}{11-layer AlexNet  \\ (trained for 40 epochs)}} & 20 &  6.54 $\pm$ 1.43 dB   \\
 \bottomrule
\end{tabular}
}
\caption{Statistics of restored image PSNR with  AlexNet trained by varying epochs, the dataset is ImageNet (batch size = 8, image size = 224$\times$224).}\label{tab: statistics2}

\end{table}

\newpage
\section{Implementation details}

\subsection{DNN Model Architectures}
The deep neural network architectures we investigated include the  well-known LeNet, AlexNet, ResNet18 and TextCNN \cite{krizhevsky2012imagenet, lecun1998gradient, he2016deep}. In particular, we modify the activation function (ReLU) in TextCNN into Sigmoid. Respectively, there are 4, 7, and 9 ReLU layers in LeNet, AlexNet and ResNet-18. When dealing with those dropout layers, we assume the model to be in evaluation mode. 
Tab. \ref{tab: archi} shows the detailed model architectures and parameter shape of LeNet, AlexNet, ResNet-18 and TextCNN. Note that we design two different AlexNet architectures for CIFAR images and ImageNet images, we report the architecture for  ImageNet.

\subsection{Datasets}
We evaluate TGIAs-RO on classification tasks with  standard MNIST, CIFAR10, ImageNet and Reuters 21578 dataset \cite{krizhevsky2009learning, deng2009imagenet}. The MNIST database of 10-class handwritten digits, has a training set of 60,000 examples, and a test set of 10,000 examples, CIFAR10 dataset consists of 60000 $32 \times 32 \times 3$ colour images in 10 classes, with 6000 images per class. The ImageNet dataset contains 14,197,122 colour images in 1000 classes, we resize the images to $224 \times 224 \times 3$ as original images. R8 (full term version) dataset is a subset of the Reuters 21578 dataset. R8 has 8 categories divided into 5,485 training and 2,189 test documents, and the text length we adopted is 32.  Respectively, we conduct stand classification tasks of MNIST,CIFAR10, ImageNet and Reuters R8 with LeNet, AlexNet, ResNet-18, and TextCNN. 


\subsection{Experiment Parameters}
\noindent\textbf{System Setup.} We consider a horizontal federated learning system in a stand-alone machine with 8 Tesla V100-SXM2 32 GB GPUs and 72 cores of Intel(R) Xeon(R) Gold 61xx CPUs. We train TGIAs-RO in total $300$ epochs via LBFGS optimization with learning rate $1$. In addition, the local iteration steps $R_l$ we set for each temporal gradient is $20$ and global iteration steps $R_g$ for robust aggregation  is $300$. For the trained model, we train the LeNet-MNIST, AlexNet-CIFAR10, ResNet-ImageNet with standard training settings.

\begin{table}[h!]
\vspace{-1pt}
    	\caption{Model architecture information for Lenet, AlexNet, ResNet-18 and TextCNN. } \label{tab: archi}
    \centering
    \renewcommand{\arraystretch}{1.03}
    \normalsize
	\resizebox{0.47\textwidth}{!}{
	
		\begin{tabular}{c|cccc}
			\toprule
			
			Modules & LeNet &AlexNet & ResNet-18 & TextCNN\\ \midrule
			Layer numbers & 5 & 8 &18 & 3 or 5\\
			Convolution layers & 2 & 5 & 16 & 1 or 3\\
			Conv. kernel size & 5 &3 & 3 & 3, (4, 5) \\
			Batch normalization & 0 & 5& 17 & 0 \\
			ReLU functions & 4 &7 & 9 & 0\\
			  Max pooling layers & 2 & 4 &2 & 3\\
			Dropout layers & 0 &2 & 1 & 1\\
			Fully connected layers & 3 & 3  &  0  & 1\\ \bottomrule
	\end{tabular}}

\end{table}

\noindent\textbf{Temporal Gradient Collection.} 
We pick model gradients for the same data batch in the first 10 iterations as our temporal information.  
For instance, if the server randomly selects 10 clients among 100 clients, they would observe each client 100 times on average across 1000 iterations. We apply the data match techniques in Sect. \ref{label and alignment} to find the first 10 temporal iterations for a data batch. It is noted that the server would get enough information for one client while training many iterations.
\subsection{Computational Cost of TGIAs-RO}

The proposed TGIAs-RO processes more temporal gradients (5, 10 or more) compared to single-temporal GIAs to achieve better image restoration performance. Thus the computation cost for TGIAs-RO is linearly higher than single-temporal GIAs. 

We evaluate the efficiency of attacking methods in the same datasets, the same model architectures, and the same computer with NVIDIA 2060TI GPU for fair comparisons, and we adopt the computation time as the metric for efficiency. Tab. \ref{tab: efficiency} presents the computational time of TGIAs-RO and other SOTA methods under varying experimental settings in a batch size of $b =8$ and $b=16$. The results indicate that multiple temporal gradients lead to almost linearly longer reconstruction time, for example, TGIAs-RO with 5 temporal gradients costs nearly half the time with 10 temporal gradients but almost 5 times that single-temporal GIA (DLG). In addition, GIAs with group consistency (GC) also cost more computation time because more GIAs optimizations are needed to explore the group consistency.

\begin{table}[htbp]
\caption{Computation time costs (minutes) of varying gradient inversion methods, we adopt TGIAs-RO with 10 temporal gradients the model architecture adopted is AlexNet and dataset is CIFAR10 (batch size = 8).}\label{tab: efficiency}
 \renewcommand{\arraystretch}{1}
 \centering
\resizebox{0.48\textwidth}{!}{
\begin{tabular}{ccccc}
\toprule
\multirow{2}{*}{ Method } &\multicolumn{2}{c}{AlexNet \& CIFAR10} &\multicolumn{2}{c}{ResNet $\&$ Imagnet}
\\ \cmidrule(r){2-3} \cmidrule(r){4-5}
&  $b =8$ & $b=16$   &  $b =8$ & $b=16$    \\ \midrule  
DLG  & 2.3  &  4.1   &  5.8 & 8.1 \\
SAPAG  & 2.4  & 4.3  &   5.5  & 8.4\\ 
GC (4 iterations in a group) & 10.8  & 16.6  & 23.2   & 30.0 \\
BN  & 4.3 & 6.3  & 10.2 &  13.4 \\
Cosine Similarity & 8.3 & 9.3 & 15.4 & 18.7 \\
TGIAs-RO (5 temporal gradients) & 11.1 & 17.1 & 24.8 & 31.6  \\
TGIAs-RO (10 temporal gradients) & 18.6 & 28.2 & 43.5 & 53.8  \\
 \bottomrule
\end{tabular}}
\end{table}

\newpage



\clearpage

\vfill

\end{document}